\documentclass[10pt,journal,compsoc]{IEEEtran}

\usepackage{xcolor}
\usepackage{cite}
\usepackage{pifont}
\usepackage{amsmath,amssymb,amsfonts,amsthm}
\usepackage{mathtools}
\usepackage{algorithmic}
\usepackage[colorlinks=true,linkcolor=blue,citecolor=blue]{hyperref}
\usepackage{graphicx}
\usepackage{textcomp}
\usepackage{subfigure}
\definecolor{azure}{rgb}{0.0, 0.5, 1.0}
\usepackage[flushleft]{threeparttable}
\usepackage{marginnote}
\usepackage{booktabs, makecell, multirow, tabularx}

\makeatletter
\newcommand\figcaption{\def\@captype{figure}\caption}
\newcommand\tabcaption{\def\@captype{table}\caption}
\makeatother
\theoremstyle{plain}
\newtheorem{theorem}{\textbf{Theorem}}[section]
\newtheorem{proposition}[theorem]{\textbf{Proposition}}

\theoremstyle{definition}

\theoremstyle{remark}

\ifCLASSOPTIONcompsoc
\else
\fi
\ifCLASSINFOpdf
\else
\fi
\begin{document}
%
\title{Scalable Graph Compressed Convolutions}
%
%
%
%
\author{Junshu~Sun,
        Shuhui~Wang,~\IEEEmembership{Member,~IEEE}, Chenxue Yang, and~Qingming~Huang,~\IEEEmembership{Fellow,~IEEE}
\IEEEcompsocitemizethanks{\IEEEcompsocthanksitem J. Sun and S. Wang are with the Key Laboratory of Intelligent Information Processing, Institute of Computing Technology, Chinese Academy
of Sciences, Beijing 100190, China. Corresponding author: Shuhui Wang.\protect\\E-mail: sunjunshu21s@ict.ac.cn; wangshuhui@ict.ac.cn.
\IEEEcompsocthanksitem C. Yang is with the Agricultural Information Institute, China Academy of Agricultural Sciences, Beijing 100081, China. \protect\\
E-mail: yangchenxue@caas.cn
\IEEEcompsocthanksitem J. Sun and Q. Huang are with the School of Computer Science and Technology, University of Chinese Academy of Sciences, Beijing 101408, China.
\IEEEcompsocthanksitem S. Wang and Q. Huang are with Peng Cheng Laboratory, Shenzhen 518066, China.}
}

%
%

\markboth{Journal of \LaTeX\ Class Files}
{Shell \MakeLowercase{\textit{et al.}}: Bare Demo of IEEEtran.cls for Computer Society Journals}
%



\IEEEtitleabstractindextext{%
\begin{abstract}
Designing effective graph neural networks~(GNNs) with message passing has two fundamental challenges, {\it i.e.}, determining optimal message-passing pathways and designing local aggregators. Previous methods of designing optimal pathways are limited with information loss on the input features. On the other hand, existing local aggregators generally fail to extract multi-scale features and approximate diverse operators under limited parameter scales. In contrast to these methods, Euclidean convolution has been proven as an expressive aggregator, making it a perfect candidate for GNN construction. However, the challenges of generalizing Euclidean convolution to graphs arise from the irregular structure of graphs. To bridge the gap between Euclidean space and graph topology, we propose a differentiable method that applies permutations to calibrate input graphs for Euclidean convolution. The permutations constrain all nodes in a row regardless of their input order and therefore enable the flexible generalization of Euclidean convolution to graphs. Based on the graph calibration, we propose the Compressed Convolution Network (CoCN) for hierarchical graph representation learning. CoCN follows local feature-learning and global parameter-sharing mechanisms of convolution neural networks. The whole model can be trained end-to-end, with compressed convolution applied to learn individual node features and their corresponding structure features. CoCN can further borrow successful practices from Euclidean convolution, including residual connection and inception mechanism. We validate CoCN on both node-level and graph-level benchmarks. CoCN achieves superior performance over competitive GNN baselines. Codes are available at \url{https://github.com/sunjss/CoCN}.
\end{abstract}

}

\maketitle

\IEEEdisplaynontitleabstractindextext

%
\IEEEpeerreviewmaketitle


\IEEEraisesectionheading{\section{Introduction}\label{sec:introduction}}

%
%
%
%

\IEEEPARstart{G}{raph} neural networks (GNNs) have become ubiquitous in the realm of graph representation learning, tackling tasks such as node classification~\cite{kipf_SemiSupervisedClassificationGraph_2017,velickovic_GraphAttentionNetworks_2018,maurya_SimplifyingApproachNode_2022}, link prediction~\cite{zhang_LinkPredictionBased_2018}, and graph classification~\cite{ying_TransformersReallyPerform_2021} in various scenarios~\cite{li_DeepGCNsMakingGCNs_2021, zhang_DynamicGraphMessage_2022, bessadok_GraphNeuralNetworks_2023}. General GNNs adopt the message passing mechanism~\cite{velickovic_MessagePassingAll_2022} to exchange information along certain pathways and update node representations iteratively. This process enables GNNs to learn the features and relationships embedded in the graph structure and therefore extract topological patterns for downstream tasks.
While message passing has been proven effective in graph learning, determining optimal message-passing pathways and designing effective local aggregators for different input graphs pose substantial challenges~\cite{zhou_GraphNeuralNetworks_2020}.

\begin{figure}[htb]
\centering
\vskip -0.1in
\hspace{0.1pt}
 \subfigure[Graph Convolution]{\label{fig:teaser-gconv}
      \begin{minipage}[t]{0.47\linewidth}
      \centering
      \includegraphics[width=\textwidth]{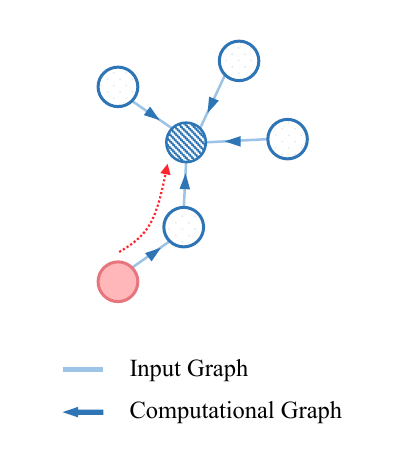}
      \end{minipage}}
\subfigure[Euclidean Convolution]{\label{fig:teaser-econv}
      \begin{minipage}[t]{0.47\linewidth}
      \centering
      \includegraphics[width=\textwidth]{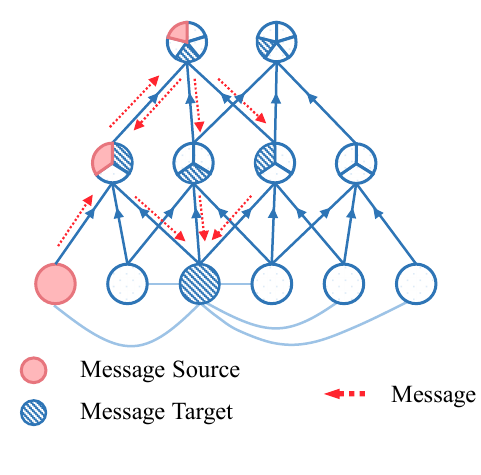}
      \end{minipage}}
\caption{\textbf{Comparison between Euclidean Convolution and Graph Convolution.} In Euclidean convolution, the message pathways are decoupled from the input topology, preventing pathway optimization from altering the input topology and causing information loss.}
\end{figure}

In determining message-passing pathways, traditional GNNs~\cite{kipf_SemiSupervisedClassificationGraph_2017,velickovic_GraphAttentionNetworks_2018} encompass information aggregation within adjacent nodes to encode graph structures, coupling the computational graphs with the input graph topology (Fig.~\ref{fig:teaser-gconv}). 
The coupled message-passing pathways are determined once given the input graphs and thus precluded from adaptive optimization for downstream tasks. As a result, the coupled pathways exhibit limited flexibility and degenerate the ability of message passing in graph learning~\cite{alon_BottleneckGraphNeural_2021,topping_UnderstandingOversquashingBottlenecks_2021,digiovanni_OversquashingMessagePassing_2023a}.
Although efforts have been devoted to decoupling the message-passing process and constructing appropriate message pathways~\cite{abu-el-haija_MixHopHigherOrderGraph_2019,klicpera_DiffusionImprovesGraph_2019,rong_DropEdgeDeepGraph_2020,black_UnderstandingOversquashingGNNs_2023}, these methods can lead to information loss regarding the input structures. Therefore, determining optimal message-passing pathways while preserving graph structures remains an open challenge.

In terms of local aggregators, graph convolutions are widely used in various GNN models~\cite{zhang_GraphConvolutionalNetworks_2019}. Nevertheless, graph convolutions have inherent limitations. First, the customized graph convolutions based on polynomials exhibit limited expressiveness within constrained parameter scales. Second, graph convolutions lack direct capabilities for hierarchical representation learning. Existing graph convolution models require additional techniques to extract features across multiple scales, such as node clustering~\cite{ying_HierarchicalGraphRepresentation_2018} and node drop~\cite{gao_GraphUNets_2019} methods. In contrast to graph convolutions, convolution neural networks (CNNs)~\cite{he_DeepResidualLearning_2016} based on Euclidean convolution are free of the above problems. Owing to the local feature learning ability and global parameter sharing mechanism, CNNs can learn diverse convolution operators and capture multi-scale local patterns on regular grid data. The expressive capacity of CNNs has propelled their success across diverse domains, including image understanding~\cite{krizhevsky_ImageNetClassificationDeep_2017, he_DeepResidualLearning_2016} and video understanding~\cite{ji_3DConvolutionalNeural_2013, tran_LearningSpatiotemporalFeatures_2015}. 

The expressiveness and flexibility of Euclidean convolution make it a perfect substitution for graph convolution. Meanwhile, applying Euclidean convolution on the graph feature matrix enables decoupled message passing from the input graph. As presented in Fig.~\ref{fig:teaser-econv}, the computational graph of Euclidean convolution serves as the message-passing pathways for input nodes while being decoupled from input graph structures. Modifications to the computational graph of Euclidean convolutions do not necessitate alterations to the original input graph structures. Consequently, Euclidean convolution can achieve structure-preserving message passing among input nodes. In light of the effectiveness of Euclidean convolution as both an effective aggregator and decoupled information conductor, a straightforward message-passing design is to employ Euclidean convolution for GNNs on non-grid graph data.

Nevertheless, directly generalizing Euclidean convolution to graphs faces challenges from various aspects.
From the data perspective, graphs generally have irregular local structures that are different from the grid data in Euclidean space. Therefore, a calibration module is required to calibrate graphs for Euclidean convolution. To achieve this, methods have been proposed based on node sequence selection~\cite{niepert_LearningConvolutionalNeural_2016, eliasof_PathGCNLearningGeneral_2022}. However, these methods may generate less informative calibrated graphs for the subsequent convolution operations, since their calibration methods are independent of the convolution process and hence cannot be optimized for specific tasks. 
From the perspective of operators, Euclidean convolutions are sensitive to the local spatial order, while GNNs should preserve permutation invariance, \emph{i.e.}, produce the same node representations regardless of the order of the nodes. For permutation invariant transformation, recent studies~\cite{murphy_JanossyPoolingLearning_2019, murphy_RelationalPoolingGraph_2019, huang_GoingDeeperPermutationSensitive_2022} propose to enumerate or sample permutations to achieve permutation invariance with permutation-sensitive operators in GNNs, but they are computationally intractable on large-scale graphs.

In this paper, we propose the Compressed Convolution Network (CoCN), a hierarchical GNN model for end-to-end graph representation learning. Compared to existing models, we highlight our CoCN as follows: \textbf{(1)} it optimizes message-passing pathways for convolution in an end-to-end fashion while preserving permutation invariance; \textbf{(2)} it can learn diverse local operators; and \textbf{(3)} it directly applies hierarchical feature learning on graphs.
Technically, CoCN generalizes Euclidean convolution to graphs with two main components, \emph{i.e.}, Permutation Generation and Diagonal Convolution. For Permutation Generation, we consider graph calibration as a permutation problem that arranges input nodes under a proper order. 
To achieve this, CoCN approximates 
the permutation matrix through a differentiable transformation with node position regression and cyclic shift. The convergence of our proposed permutation generation method and the permutation invariance of the permuted features are theoretically demonstrated in this paper. By employing permutation on the input graphs, Permutation Generation assigns nodes with the learned order. The learned order determines the receptive field for the subsequent convolution and constructs the computational graphs as decoupled message-passing pathways.

\begin{figure*}[htb]
\centering
\includegraphics[width=0.96\textwidth]{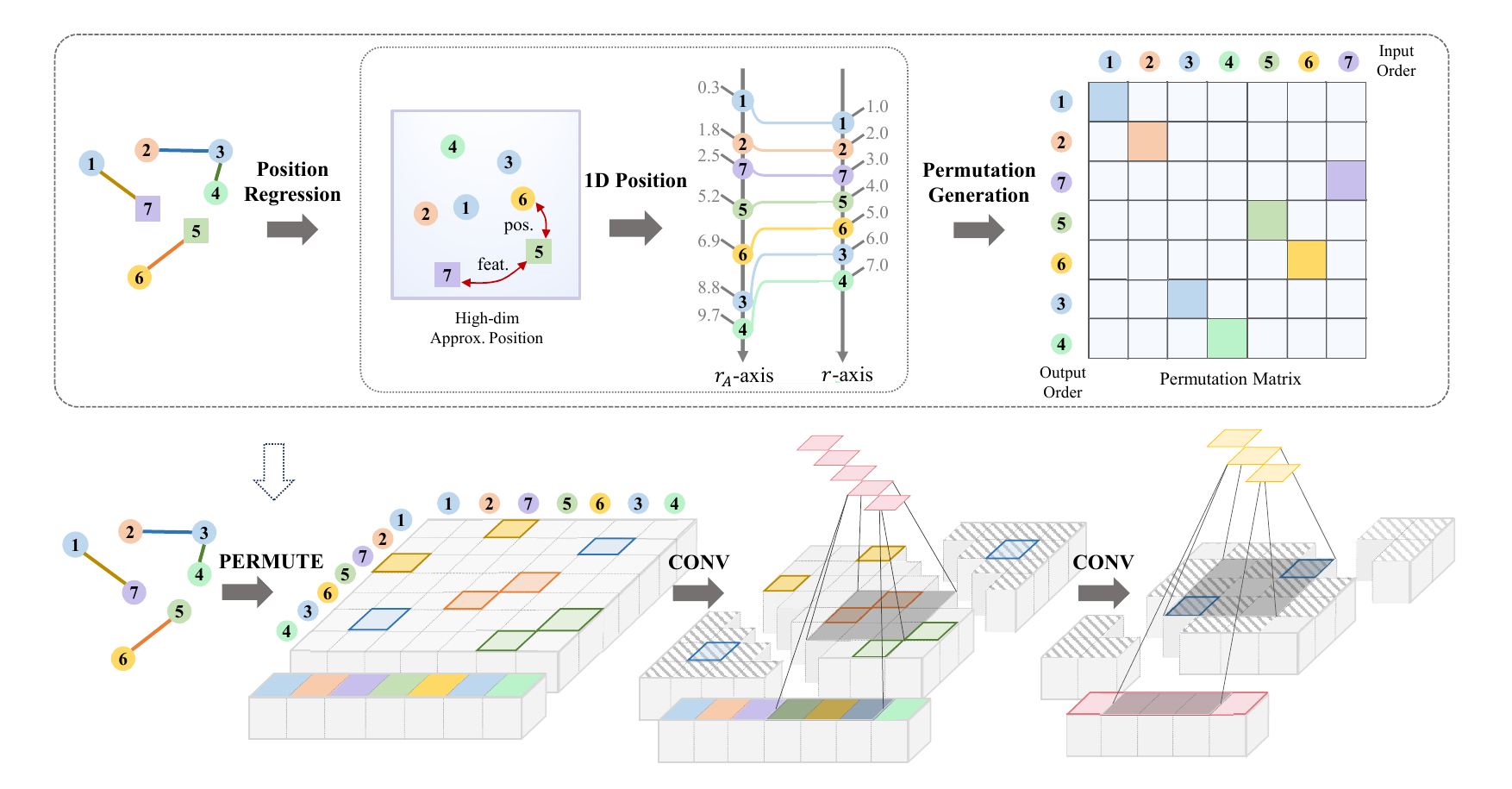}
\vskip -0.1in
\caption{\textbf{Compressed Convolution Network Pipeline.} CoCN first permutes node feature and edge feature matrices based on the learnable permutation matrix. Then diagonal convolution is applied to both feature matrices following the diagonal sliding fashion on the edge feature matrix. The off-diagonal features, displayed with the twill pattern, are not involved in the convolution but are learned in the subsequent layers.}
\label{fig:pipeline}
\end{figure*}

Based on the permuted graph representation, Diagonal Convolution is proposed to aggregate both individual node features and the corresponding structure features. It inherits the local feature-learning and global parameter-sharing mechanisms from Euclidean convolution, and follows the diagonal sliding fashion for edge feature learning. Moreover, to directly achieve hierarchical graph feature learning, anti-diagonal compression (Fig.~\ref{fig:pipeline}) is proposed for edge features update. By applying Diagonal Convolution with anti-diagonal compression iteratively, the input nodes are compressed into node sets, and the edge features matrix is compressed from the anti-diagonal direction. Hence, we name Diagonal Convolution with anti-diagonal compression as a \textit{compressed convolution layer}.
Building upon the Diagonal Convolution, CoCN can learn diverse local operators to extract both node features and edge features explicitly.

\textbf{Contributions.} Our contribution can be summarized as follows: \textbf{(1)} We propose a novel method to calibrate graphs that enables the generalization of the permutation-sensitive Euclidean convolution to graphs. \textbf{(2)} We propose a hierarchical GNN model, CoCN, which can learn both individual node features and the corresponding structure features from coarse to fine. \textbf{(3)} We demonstrate the advantages of CoCN on both node-level tasks and graph-level tasks.

A preliminary version of this work was published in~\cite{sun_AllinARow_2023}. This journal manuscript extends the initial version in several aspects:~
\textbf{(1)} More types of graph data. We improve CoCN to extend its effectiveness to more types of graph data. First, based on the existence of node features, graph data can be categorized into graphs with explicit or implicit node features. Our vanilla CoCN model~\cite{sun_AllinARow_2023} can only be applied to the former types of benchmarks. In this manuscript, we propose a position regression method tailored for the implicit types of benchmarks. Second, regarding the graph scales, the CoCN in this paper can be applied to large-scale graphs with sparse implementation and node sequence segmentation. Third, we evaluate CoCN on both homophilic and heterophilic graph benchmarks. Comprehensive empirical analysis demonstrates the effectiveness and superior performance of our CoCN family on these types of graph data.
\textbf{(2)} Generalization of the common practices in CNNs. We investigate the generalizability of some successful practices on CNNs from Euclidean space to graph space, including residual connection and inception mechanism. We find that with special modification, the residual connection can constantly benefit CoCN on various benchmarks while the inception mechanism also improves the performance of CoCN on parts of benchmarks.
\textbf{(3)} More graph tasks. We examine the performance of CoCN on more graph tasks, including the graph isomorphism test, link prediction, and the downstream brain connectomics classification. The evaluation results show the effectiveness of CoCN as a general graph representation learning backbone.

\begin{figure}[htb]
\centering
\includegraphics[width=0.94\linewidth]{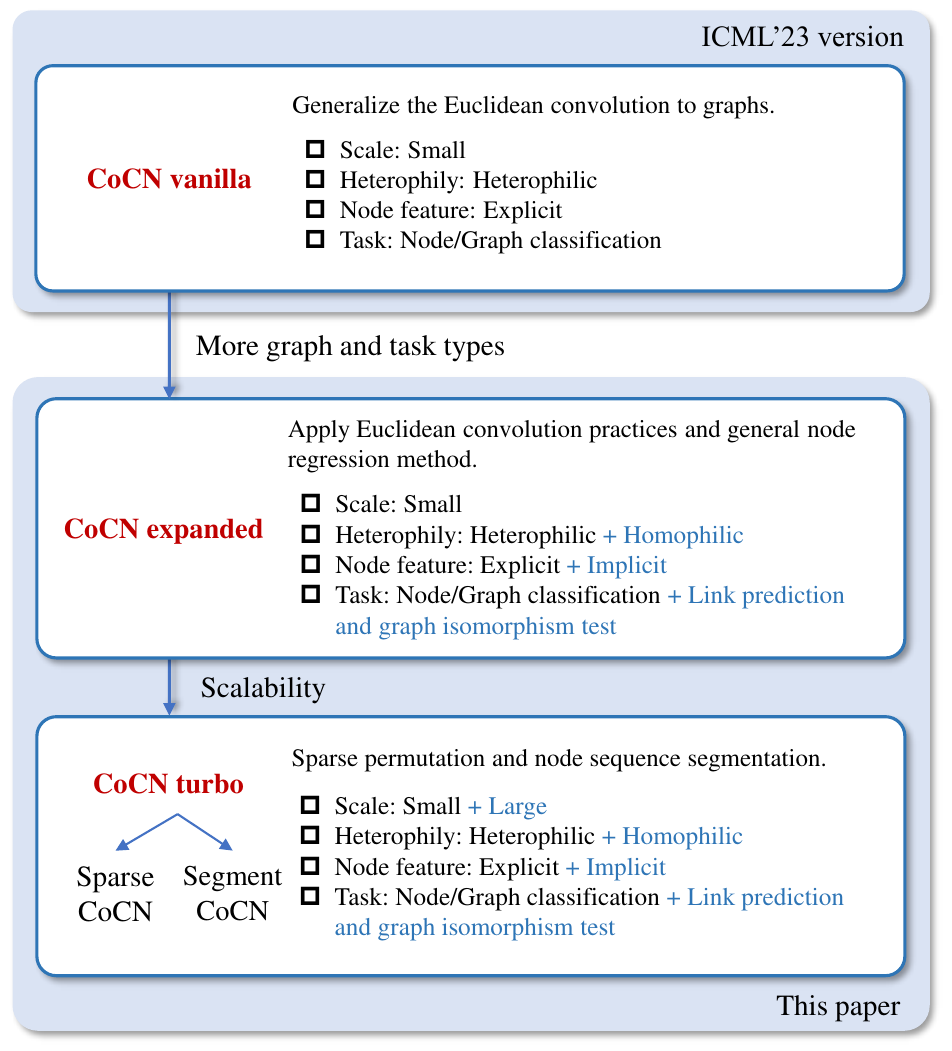}
\caption{\textbf{The CoCN model family.}}\label{fig:family}
\end{figure}

We delve into the generalization of Euclidean convolution to graphs with a series of CoCN models. The initial version of CoCN in \cite{sun_AllinARow_2023} is named \texttt{CoCN vanilla}. To extend the effectiveness of CoCN, we boost CoCN vanilla with position regression tailored for implicit graph features, residual connection, and inception mechanism. The extended version is named \texttt{CoCN expanded}. Both \texttt{CoCN vanilla} and \texttt{CoCN expanded} show effectiveness on small-scale benchmarks. Finally, to deal with graphs of various scales, we develop more scalable CoCNs named \texttt{CoCN turbo}, \textit{i.e.}, Sparse CoCN and Segment CoCN, through sparsification and node sequence segmentation. The comparison among different versions of CoCN is depicted in Fig.~\ref{fig:family}. In the subsequent context, we refer to the CoCN models as CoCN when not specifying any particular version. The rest of this paper is organized as follows. Section~\ref{sec:related-work} presents related works with their major differences to our method. Section~\ref{sec:perm-gen} and~\ref{sec:diag-conv} describe our differentiable Permutation Generation and Diagonal Convolution methods, respectively. Section~\ref{sec:cocn} presents the implementation of \texttt{CoCN vanilla} and \texttt{CoCN expanded}. In Section~\ref{sec:scalable-cocn}, we describe how to extend CoCN to large-scale benchmarks. In Section~\ref{sec:experiment}, comprehensive evaluations are conducted on CoCN with various types of graph data and tasks. Finally, in Section~\ref{sec:conclusion} we conclude our study.

\section{Related Work}\label{sec:related-work}
\subsection{Convolutional Message-passing Aggregators}
To generalize convolution operators to graph-structured data, Bruna et al.,~\cite{bruna_SpectralNetworksLocally_2013} first propose graph convolution based on graph signal processing theory. ChebNet~\cite{defferrard_ConvolutionalNeuralNetworks_2016} further reduces the filter parameter scale by learning convolution filters with parameterized graph Laplacian polynomials. Following up on this work, models with simpler filter structures have been studied, {\it e.g.}, first-order low-pass filter model GCN~\cite{kipf_SemiSupervisedClassificationGraph_2017} and SGC~\cite{wu_SimplifyingGraphConvolutional_2019}. Except for graph Laplacian, methods have been proposed to explore the effectiveness of other polynomials for graph learning. Klicpera et al.,~\cite{klicpera_DiffusionImprovesGraph_2019} extend graph Laplacian to a more general transition matrix. Inspired by Feynman path integral theory~\cite{dick_PathIntegralsQuantum_2020}, PAN~\cite{ma_PathIntegralBased_2020} formulates graph convolution as the polynomial of adjacency matrix with coefficients depending on the corresponding path. JacobiConv~\cite{wang_HowPowerfulAre_2022} adopt Jacobi polynomials. Guo and Wei~\cite{guo_GraphNeuralNetworks_2023} explore learnable and optimal bases for the polynomials in GNN. Although polynomials can approximate any sophisticated filter theoretically~\cite{wang_HowPowerfulAre_2022}, its expressive power is bounded by the order, while high-order filters are known to be computationally expensive.

The other vital type of graph convolution is combining the calibrated graphs with shared filters. These methods can learn complex filters without the constraint of polynomial order. PATCHY-SAN~\cite{niepert_LearningConvolutionalNeural_2016} extracts the calibrated neighborhoods to serve as the input data to CNNs. However, its graph calibration cannot be integrated into the learning process and optimized for specific tasks, resulting in task-agnostic calibrated graphs with less discriminating features for downstream tasks.
To incorporate more expressive calibrated graphs, enumerating methods have been proposed~\cite{murphy_JanossyPoolingLearning_2019, murphy_RelationalPoolingGraph_2019, huang_GoingDeeperPermutationSensitive_2022, michel_PathNeuralNetworks_2023}, albeit with intractable computational costs. Consequently, sampling strategies are developed to improve computational efficiency. PathConv~\cite{eliasof_PathGCNLearningGeneral_2022} calibrates the graph structures with random walk and applies convolution to the generated node sequence. To guide the random walk process, PathNet~\cite{sun_HomophilyStructureawarePath_2022} further introduces entropy maximization and sampling paths with increasing entropy. PG-GNN~\cite{huang_GoingDeeperPermutationSensitive_2022} also utilizes sampling methods to convert the neighborhood of nodes into sequences and models the pairwise correlations by RNN. Despite using expressive spatial filters, the above-mentioned sampling strategies still involve increasing computational costs and may introduce noisy signals. Compared to existing calibration-based models, our CoCN calibrates the input graphs with learnable permutations that can be optimized for specific tasks, saving time from enumeration or sampling.

\subsection{Construction of Message-passing Pathways}
The efforts devoted to determining optimal message pathways aim to modify the structure of a graph, facilitating more effective message passing for graph learning. To obtain optimal pathways, assessments of communication within the input graph structure have been studied, such as balanced Forman curvature~\cite{topping_UnderstandingOversquashingBottlenecks_2021}, Ollivier-Ricci curvature~\cite{nguyen_RevisitingOversmoothingOversquashing_2023}, and effective resistance~\cite{black_UnderstandingOversquashingGNNs_2023, digiovanni_OversquashingMessagePassing_2023a}. Based on the communication assessment, the addition and removal of edges between nodes are employed to the input structures~\cite{suresh_BreakingLimitGraph_2021, chen_MeasuringRelievingOverSmoothing_2020a, arnaiz-rodriguez_DiffWireInductiveGraph_2022, dong_UnderstandingReducingGraph_2023}. However, directly altering the structure of the input graph may lead to information loss~\cite{gutteridge_DRewDynamicallyRewired_2023} and degrade the expressive capacity of GNNs. In contrast, Euclidean convolutions serving as message-passing pathways are decoupled from the input graphs, and do not necessitate alterations to the original graph structures.

\subsection{Hierarchical Message Passing}
Except for node-level and graph-level features, the intermediate-scale features are also crucial for graph representation learning~\cite{ying_HierarchicalGraphRepresentation_2018, boguna_NetworkGeometry_2021}. To this end, GNNs with hierarchical representation learning have been studied, generally called graph pooling or graph coarsening. Defferrard et al.~\cite{defferrard_ConvolutionalNeuralNetworks_2016} use the Graclus greedy algorithm to select and combine node pairs at every coarsening level. To further preserve structural information, more topology-based clustering methods are adopted for iterative graph pooling, such as edge collapsing~\cite{Hu2006EfficientHF, chen_HARPHierarchicalRepresentation_2018}, structural similarity~\cite{hu_HierarchicalGraphConvolutional_2019} and spectral similarity~\cite{deng_GraphZoomMultilevelSpectral_2020}.
Other node clustering methods~\cite{ying_HierarchicalGraphRepresentation_2018, wang_SecondOrderPoolingGraph_2020, baek_AccurateLearningGraph_2022} learn soft assignment matrix rather than deterministic clustering to perform graph pooling. Except for node clustering, node drop methods~\cite{cangea_SparseHierarchicalGraph_2018, gao_GraphUNets_2019, lee_SelfAttentionGraphPooling_2019, gao_TopologyAwareGraphPooling_2021} use top-$K$ selection method to drop irrelevant nodes. Despite their effectiveness, these methods suffer from information loss~\cite{wu_StructuralEntropyGuided_2022}. To address this problem, Wu et al.~\cite{wu_StructuralEntropyGuided_2022} use structural entropy to get a learning-free hierarchical structure.  
Different from graph pooling methods, our CoCN avoids node clustering or node drop layer-by-layer and performs hierarchical representation learning directly through diagonal convolution with anti-diagonal compression.

\section{Permutation Generation}\label{sec:perm-gen}
In this section, we describe how to approximate the permutation matrix through differentiable procedures. The learnable nature of this method enables us to optimize task-specific calibration on the general graph-structured data and further facilitates the whole model to be trained end-to-end.

\subsection{Preliminaries}
Let $\mathcal{G}=(\mathcal{V}, \mathcal{E})$ be a graph with node set $\mathcal{V}=\{v_1, v_2, \cdots, v_n\}$ of $n$ nodes and edge set $\mathcal{E}$. Each node $v\in\mathcal{V}$ has a feature vector $\mathbf{x}_v\in\mathbb{R}^d$ where $d$ denotes the number of features. We use $\mathbf{X}\in\mathbb{R}^{n\times d}$ to denote the graph node feature matrix. Node feature vector $\mathbf{x}_{i}^\top$ corresponds to the $i$-th row of $\mathbf{X}$. Let $\mathbf{A}\in\mathbb{R}^{n\times n}$ be the adjacency matrix of $\mathcal{G}$ and $\mathbf{D}\in\mathbb{R}^{n\times n}$ be the diagonal degree matrix. $\mathbf{A}_{i,j}=1$ if there exists an edge $e_{i,j}=(i,j)\in\mathcal{E}$ and $\mathbf{D}_{i,i} = \sum_{j}\mathbf{A}_{i,j}$. We use $\mathbf{1} \in \mathbb{R}^n$ to denote the all-ones vector.  

Let $\mathcal{P}$ be the permutation set over $n$ indices, $|\mathcal{P}|=n!$. For any $\mathbf{P}\in\mathcal{P}$, $\mathbf{P}\in\mathbb{R}^{n\times n}$, $\mathbf{P1}=\mathbf{1}^\top \mathbf{P}=\mathbf{1}$, $\mathbf{P}_{i,j} \in \{0, 1\}$. Given function $f(\cdot, \cdot)$ and input features $\mathbf{X}$ and $\mathbf{A}$, $f$ is permutation equivariant such that for any permutation matrix $\mathbf{P}\in\mathcal{P}$, $f(\mathbf{PX},\mathbf{PAP}^\top)=\mathbf{P}f(\mathbf{X},\mathbf{A})$. As the intrinsic features of graphs stay invariant under permutations of the node order, GNNs are required to exhibit permutation equivariance concerning the input node sequence and output invariant features for each node.

\subsection{Position Regression}
Permutation on graphs can be regarded as node assignment. Take row permutation as an example, $\mathbf{P}_{i,j}=1$ indicates moving the node corresponding to the $j$-th row to the $i$-th row. The objective of permutation generation is to learn the target position for each node. Therefore, permutation generation can be first formulated as a position regression problem. To capture the global correlations between input nodes, both individual node features and their corresponding structures should be taken into consideration. To this end, the position regression solution is derived under the assumption that nodes with similar features or short paths in between will get closer position predictions. The approximate position $\mathbf{r}_A\in\mathbb{R}^n$ is given by:
\begin{equation}\label{eq:approximate-pos}
\mathbf{r}_A=f\left(\mathbf{X}, \mathbf{A}\right),
\end{equation}
where $f$ can be any permutation equivariant function. Since the approximate position for each node is a scalar, we can apply a global pairwise value comparison to rank the nodes. The resulted absolute position $\mathbf{r}\in\mathbb{R}^n$ can be formulated as:
\begin{equation}\label{eq:absolute-pos}
\mathbf{r}=\texttt{sgn}\left(\mathbf{r}_A \mathbf{1}^\top-\mathbf{1}\mathbf{r}_A^\top\right)\mathbf{1},
\end{equation}
where $\texttt{sgn}(\cdot)$ denotes the sign function. If $x>0$, $\texttt{sgn}(x)=1$ otherwise $\texttt{sgn}(x)=0$. Note that the sign function is not differentiable. To address this problem, we use the sigmoid function with ReLU to approximate the sign function in backpropagation.

\subsection{Permutation Matrix Generation}
With Eq.~\ref{eq:approximate-pos} and Eq.~\ref{eq:absolute-pos}, the generated node positions are assigned to all input nodes. A natural follow-up question is how to convert the absolute position $\mathbf{r}$ into a permutation matrix. Given absolute position $\mathbf{r}_i$ for the $i$-th node, the corresponding permutation matrix $\mathbf{P}$ should have an entry equal to $1$ at $(\mathbf{r}_i, i)$. To convert absolute positions into a permutation matrix, the model needs to assign $1$ to the given positions. We resort to the cyclic shift~\cite{Carter2009VisualGT} of matrix entries to achieve entry value assignment on matrices.

We consider the column vector $\mathbf{P}_{\cdot, j}$ corresponding to the $j$-th node assignment. Let $\mathbf{m}\in\mathbb{R}^n$ denote the indicator of $\mathbf{P}_{\cdot, j}$, initialized as $\mathbf{m}_i=i$. The zero entry of $\mathbf{m}$ indicates that the corresponding entry of $\mathbf{P}_{\cdot, j}$ equals $1$, while non-zero entries indicate the number of cyclic shift steps to $1$. $\mathbf{P}_{i,j}$ is $\mathbf{m}_i$ steps away from $1$. The cyclic shift on $\mathbf{m}$ is equivalent to value assignment on $\mathbf{P}_{\cdot, j}$. 
We use element-wise addition and modulus operator to cyclically shift the indicator entries. 
Given absolute position $k$, the initial indicator $\mathbf{m}$ takes $k$ steps cyclic shift by $(\mathbf{m} - k + n) \pmod{n}$.

We now extend to determine the whole permutation matrix for the given absolute position $\mathbf{r}$. Each column of the permutation matrix corresponds to a single node assignment. Let $\mathbf{m1}^\top$ denote the initial indicator of the permutation matrix. The absolute position $\mathbf{r}$ can be converted into a permutation matrix as:
\begin{equation}\label{eq:permutation}
\hat{\mathbf{P}}=\exp\left\{-\tau \left[\left(\mathbf{m1}^\top- \mathbf{1r}^\top + n\right)\pmod{n}\right]\right\},
\end{equation}
where $\tau$ denotes the relaxation factor and $\exp(\cdot)$ denotes the exponential function for mapping indicator entries to permutation entries. Note that the Eq.~\ref{eq:permutation} gives the relaxed permutation matrix to overcome the gradient vanishing problem, where $\hat{\mathbf{P}}_{i,j}\in\left(0,1\right]$. One can instead use Proposition~\ref{prop:permutation} or replace $\exp(\cdot)$ with other functions to get the standard permutation matrix. Proofs for propositions in this section are provided in the Appendix.

\begin{proposition}[Permutation Convergence]\label{prop:permutation}
$\hat{\mathbf{P}}$ converges to standard permutation matrix as relaxation factor $\tau$ approaching positive infinity:~$\underset{\tau\rightarrow +\infty}{lim}\hat{\mathbf{P}}=\mathbf{P}, \mathbf{P}\in\mathcal{P}$.
\end{proposition}

For simplicity, we compile Eq.~\ref{eq:approximate-pos}-\ref{eq:permutation} as operation $\texttt{PERM}(\cdot)$. The output of $\texttt{PERM}(\cdot)$ can be used to generate permutation invariant input for subsequent convolution. 

\begin{proposition}[Permutation Invariant Input]\label{prop:invariance}
Let $f$ in $\texttt{PERM}(\cdot)$ be a permutation equivariant function. Then given $\hat{\mathbf{P}}=\texttt{PERM}(\mathbf{X}, \mathbf{A})$, $\hat{\mathbf{X}}=\hat{\mathbf{P}}\mathbf{X}$, and $\hat{\mathbf{A}}= \hat{\mathbf{P}}\mathbf{A}\hat{\mathbf{P}}^\top$, for any $\mathbf{P}\in\mathcal{P}$ on $\mathbf{X}$ and $\mathbf{A}$, $\hat{\mathbf{X}}$ and $\hat{\mathbf{A}}$ are invariant.
\end{proposition}

The calibration on graphs with position regression and order permutation transfers the irregular graph to a regular vector with proper order. Proposition~\ref{prop:permutation} and~\ref{prop:invariance} demonstrate that such permutation is differentiable, convergent, and input-order invariant, which facilitates flexible calculation of spatial convolutions.

\section{Diagonal Convolution}\label{sec:diag-conv}
The permutations serve as a spatial position mapping to Euclidean space, arranging input nodes in a row. This graph calibration ensures a uniform local geometry structure shared by all nodes, facilitating the generalization of convolution from Euclidean space to graphs. Built upon this, we now proceed to introduce the diagonal convolution operator, specifically tailored for the calibrated node sequence and its corresponding topological structure.

Let $\hat{\mathbf{X}}=\hat{\mathbf{P}}\mathbf{X}$ be the sequential node feature matrix and $\hat{\mathbf{A}}=\hat{\mathbf{P}}\mathbf{A}\hat{\mathbf{P}}^\top$ be the sequential topological feature matrix. Note that the adjacency matrix can be seen as the edge feature matrix with a single channel. We can easily generalize our model to multi-channel edge features. For simplicity, we only consider the adjacency matrix here. Similar to 2D convolution, the diagonal structural convolution with a single kernel can be formulated as:
\begin{equation}\label{eq:edge-conv}
\mathcal{K}_{i}^{str}\left(\hat{\mathbf{A}},k\right)=\sum_{p=0}^{k-1}\sum_{q=0}^{k-1}\mathbf{w}_{p,q}\hat{\mathbf{A}}_{i+p,i+q},
\end{equation}
where $\mathbf{w}\in\mathbb{R}^{k\times k}$ denotes the convolution kernel parameters for $\hat{\mathbf{A}}$, $k$ denotes the kernel size and $i$ denotes the index of $\hat{\mathbf{A}}$ that corresponds to the top-left entry of convolution kernel. Eq.~\ref{eq:edge-conv} gives the single-step diagonal structural convolution. Typical 2D convolution follows the progressive sliding fashion on images where the kernels first slide in a column-wise manner and then row-wise. Instead, we perform diagonal sliding on $\hat{\mathbf{A}}$ to adapt to the scale of graphs. Specifically, for $k\times k$ convolution with step length $s$, the top-left entry index $i$ starts at $(0,0)$, moves to $(s,s)$ and stops at $(n-k+1,n-k+1)$. Further including sequential node features, the diagonal structural convolution can be generalized to a unified diagonal convolution as:
\begin{equation}\label{eq:clique-conv}
\mathcal{K}_{i}\left(\hat{\mathbf{A}}, \hat{\mathbf{X}}, k\right)=\sum_{p=0}^{k-1}\left(\sum_{q=0}^{k-1}\mathbf{w}_{p,q}\hat{\mathbf{A}}_{i+p,i+q}+\sum_{t=0}^{d-1}\mathbf{v}_{p,t}\hat{\mathbf{X}}_{i+p,t}\right),
\end{equation}
where $\mathbf{v}\in\mathbb{R}^{k\times d}$ denotes the convolution kernel parameters for $\hat{\mathbf{X}}$. For the input graph with $n$ nodes, applying $k\times k$ diagonal convolution with diagonal sliding step length $s$ gives rise to the node feature vector of length $\lfloor\frac{n-k}{s}\rfloor+1$.  Let $\mathbf{S}=(\mathbf{S}_j)\in\mathbb{R}^{\lfloor\frac{n-k}{s}\rfloor+1}$ denote the output single channel node feature vector. We can formulate the multi-step diagonal convolution with diagonal sliding as:
\begin{equation}
\begin{aligned}
&\mathbf{S} = \mathcal{K}\left(\hat{\mathbf{A}}, \hat{\mathbf{X}},k,s\right),\\
&\mathbf{S}_j = \mathcal{K}_{i}\left(\hat{\mathbf{A}}, \hat{\mathbf{X}},k\right),\ i=sj.
\end{aligned}
\end{equation}

\section{Compressed Convolution Networks}\label{sec:cocn}
We present the general framework of the Compressed Convolution Network (CoCN), a GNN model that employs diagonal convolution for hierarchical graph representation learning. We commence with a detailed explanation on the implementation of the fundamental building blocks of CoCN and then describe the network architecture for graph-related tasks at both the node level and graph level. Under the general framework, \texttt{CoCN vanilla} is developed with explicit node features for position regression and plain compressed convolution. On the other hand, \texttt{CoCN expanded} extends to graphs with implicit node features and adopts common practices from Euclidean convolution, including residual connection~\cite{he_DeepResidualLearning_2016} and inception mechanism~\cite{szegedy_GoingDeeperConvolutions_2015}.

\subsection{Position Regression Module}\label{ssec:permutation-module}
The permutation module is described in Section~\ref{sec:perm-gen}. There are many possible implementations of the permutation equivariant function in Eq.~\ref{eq:approximate-pos}. In this paper, we propose solutions tailored to deal with graphs based on the explicitness of node features. For graphs with explicit node features, the node features are explicitly available which can be denoted as $\mathbf{X}$. For graphs with implicit node features, characterized by the lack of explicitly defined node features, the information regarding node features is implicitly encoded within the inherent structures of the graph.

\textbf{Graphs with explicit node features.}~
We use MLP to learn node features and the Laplacian operator to smooth the approximate position of connected nodes. Nodes with similar features or short paths in between will have similar positions. Eq.~\ref{eq:approximate-pos} can be written as:
\begin{equation}\label{eq:approximate-ex-pos-implementation}
\mathbf{r}_A=\tilde{\mathbf{A}}^t \texttt{MLP}(\mathbf{X}),
\end{equation}
where $\tilde{\mathbf{A}}=\mathbf{D}^{-\frac{1}{2}}\mathbf{A}\mathbf{D}^{-\frac{1}{2}}$ denotes the Laplacian operator, and $t$ denotes the power of $\tilde{\mathbf{A}}$ to adjust the smoothness. $\mathtt{MLP}(\cdot)$ uses ReLU as the activation function at each layer.

\textbf{Graphs with implicit node features.}~
For graphs only characterized by topological descriptions, explicit node features are not available for position regression. Consequently, we base node position regression solely on the distances between nodes, leading to proximate positions for nodes with short paths in between. This intuition can be formulated as:
\begin{equation}\label{eq:implicit-intuition}
\begin{aligned}
    &\mathbf{M}^U\mathbf{11}^{\top}-\mathbf{11}^{\top}{\mathbf{M}^U}^\top = \mathbf{M}^D\odot d(\mathbf{A}),\\
    &\mathbf{M}^D = 2\mathbf{M}^U-\mathbf{11}^\top,\\
    &\mathbf{M}^U = \texttt{sgn}\left(\mathbf{r}_A \mathbf{1}^\top-\mathbf{1}\mathbf{r}_A^\top\right),
\end{aligned}
\end{equation}
where $\mathbf{M}^U$ and $\mathbf{M}^D$ denote the sign matrices for the position comparison between nodes. $\mathbf{M}^{U}_{ij}, \mathbf{M}^{D}_{ij} = 1$ if $\mathbf{r}_{Ai}>\mathbf{r}_{Aj}$, otherwise $\mathbf{M}^{U}_{ij}=0$ and $\mathbf{M}^{D}_{ij}=-1$. $\odot$ denotes element-wise multiplication. $d(\cdot)$ denotes distance function such that $d(\mathbf{A})\in\mathbb{R}^{n\times n}$. By substituting Eq.~\ref{eq:absolute-pos} into Eq.~\ref{eq:implicit-intuition}, the distance-based constraint on the node positions is reformulated as:
\begin{equation}
    \mathbf{r1}^\top-\mathbf{1r}^\top=(2\mathbf{r}-n\mathbf{1})\mathbf{1}^\top\odot \frac{1}{n}d(\mathbf{A}).
\end{equation}
To solve for the position $\mathbf{r}$, we adopt the equivalent vectorization with the Kronecker product and have the approximate position $\mathbf{r}_A$ as:
\begin{equation}\label{eq:approximate-im-pos-implementation}
\begin{aligned}
    \mathbf{r}_A \approx & \left[ \texttt{diag}\left(\left(\frac{2}{n}d(\mathbf{A})-1\right)^{\circ 2}\mathbf{1}\right) \right.\\    
    &\ + \left.\frac{4}{n}d(\mathbf{A}) - 1\right]^{-1} \frac{2}{n}d(\mathbf{A})^{\circ 2}\mathbf{1},
\end{aligned}
\end{equation}
where $(\cdot)^{\circ 2}$ denotes element-wise square and $d(\cdot)$ is implemented as the scaled shortest path distance. For detailed derivation, please refer to the Appendix.
Note that in Eq.~\ref{eq:approximate-im-pos-implementation}, we avoid introducing any learnable parameter. Therefore, the approximate position for graphs with implicit node features can be computed before the training procedure and then transformed with $\mathtt{MLP}(\cdot)$.

In practice, we produce multiple approximate positions $\mathbf{r}_A$ to allow node arrangements under different similarities. The permutation generation module then follows Eq.~\ref{eq:absolute-pos} and Eq.~\ref{eq:permutation} to generate multiple permutations to the subsequent convolution layers.

\subsection{Compressed Convolution Module}
\subsubsection{Plain Compressed Convolution} 
To construct a multi-layer network with diagonal convolution, we need to update the input features at each layer. As described in Section~\ref{sec:diag-conv}, the diagonal convolution follows the diagonal sliding fashion on the input edge feature matrix ($\hat{\mathbf{A}}$ for example). At each step, the diagonal convolution extracts node set features including individual node features from $\hat{\mathbf{X}}$ and the structure features among nodes from the main diagonal blocks of $\hat{\mathbf{A}}$. 

For individual node feature update, we take input nodes as $n$ node sets where each node set only contains a single node. Let $\mathbf{H}$ be the node set feature matrix and $\mathbf{H}^{(0)}=\hat{\mathbf{X}}$ be the initial node set feature matrix. $\mathbf{H}$ can be updated with the diagonal convolution output.

For structure feature update, let $\mathbf{E}$ be the structure feature matrix initialized with $\mathbf{E}^{(0)}=\hat{\mathbf{A}}$\footnote{ $\hat{\mathbf{A}}$ can be replaced with any multi-channel edge feature matrix.}. The off-diagonal features in $\mathbf{E}$ that are not involved in the convolution represent the topological structure features among node sets. If the diagonal convolution takes unit sliding steps, we can update $\mathbf{E}$ by removing the main diagonal blocks. We use PyTorch-style pseudo-code to formulate the update equation as $\mathbf{E}^{(l)}=\texttt{Tri}(\mathbf{E}^{(l-1)},k)=\texttt{triu}(\mathbf{E}^{(l-1)},k)+\texttt{tril}(\mathbf{E}^{(l-1)},-k)$, where $k$ denotes the kernel size, $\texttt{triu}(\cdot,i)$ denotes the upper triangular matrix with $i$ diagonals above the main diagonal and $\texttt{tril}(\cdot,-i)$ denotes the lower triangular matrix with $i$ diagonals below the main diagonal. If the diagonal convolution takes non-unit steps, we use standard 2D max pooling with the same kernel size and step size as diagonal convolution to update the structure features.

With the proposed update method, the structure feature matrix $\mathbf{E}$ is compressed from the anti-diagonal direction. We name this network layer with diagonal convolution and anti-diagonal compression as \emph{compressed convolution layer}. The $l$-th compressed convolution layer with $k^{(l)}\times k^{(l)}$ diagonal convolution and sliding step $s^{(l)}$ is formulated as:
\begin{equation}\label{eq:coco-layer-node}
\mathbf{H}^{(l)} = \sigma\left(\mathcal{K}\left(\mathbf{E}^{(l-1)}, \mathbf{H}^{(l-1)},k^{(l)},s^{(l)}\right)\right),
\end{equation}
\begin{equation}\label{eq:coco-layer-edge}
\mathbf{E}^{(l)} = \begin{cases}
\texttt{Tri}\left(\mathbf{E}^{(l-1)},k^{(l)}\right), &\text{if $s^{(l)}=1$;}\\
\texttt{MaxPool}\left(\mathbf{E}^{(l-1)},k^{(l)},s^{(l)}\right), &\text{otherwise},
\end{cases}
\end{equation}
where $\sigma$ denotes the ReLU function, $\mathbf{E}^{(l)}\in\mathbb{R}^{n^{(l)}\times n^{(l)}}$ and $n^{(l)}$ denotes the number of the output node sets. Following the common practice in CNN models, compressed convolution layers will contain multiple convolution kernels which give rise to $\mathbf{H}^{(l)}\in\mathbb{R}^{n^{(l)}\times c^{(l)}}$, where $c^{(l)}$ denotes the number of kernels. Compressed convolution layer with non-unit step is referred to as \textit{compressed pooling layer} in the following context.

\begin{figure}[htb]
\centering
\includegraphics[width=0.45\textwidth]{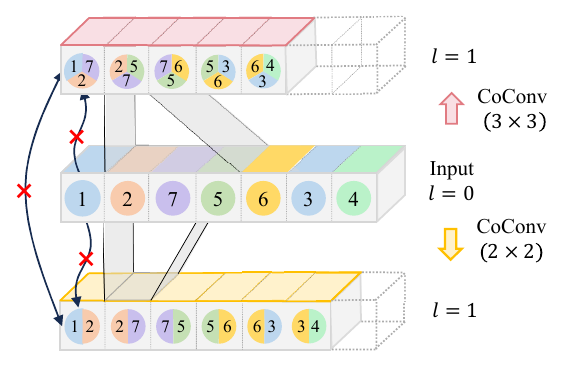}
\caption{\textbf{Incosistency in the Sizes of Node Sets.}}\label{fig:align}
\end{figure}

\subsubsection{Compositional Compressed Convolution}
Inspired by the practice of traditional CNN models~\cite{he_DeepResidualLearning_2016}, we seek to generalize the concept of residual connection~\cite{he_DeepResidualLearning_2016} and inception mechanism~\cite{szegedy_GoingDeeperConvolutions_2015} for CoCN. Let $\mathbf{H}^{(l, k)}\in\mathbb{R}^{n^{(l, k)}\times c^{(l, k)}}$ be the convolution output of the $l$-th layer with $k\times k$ kernels. Residual connection requires feature shape alignment in the first dimension given different values of $l$ while inception aligns with different $k$. In Euclidean convolution, this can be easily implemented through feature padding. However, in compressed convolution, different $l$ and $k$ not only give rise to different values of $n^{(l, k)}$, but also result in different sizes of the output node sets. As presented in Fig.~\ref{fig:align}, groups of nodes or node sets are compressed into higher-level node sets, expanding the size of the output node sets. The output node sets at the $1$-st layer with $2\times 2$ kernel contain $2$ nodes while $3\times 3$ kernel gives rise to $3$ nodes.
Although padding could be utilized to align the number of node sets across layers and kernels, the sizes of individual node sets would still be inconsistent. This inconsistency hinders the output fusion of $\mathbf{H}^{(l, k)}$ with different $l$ and $k$, thus obstructing the generalization of residual connection and inception mechanism. To address this problem, we develop a residual compressed convolution module and an inception compressed convolution module for CoCN.

\begin{figure}[htb]
\centering
\subfigure[Euclidean Conv.]{\label{fig:res-ori}
      \begin{minipage}[t]{0.3\linewidth}
      \centering
      \includegraphics[width=\textwidth]{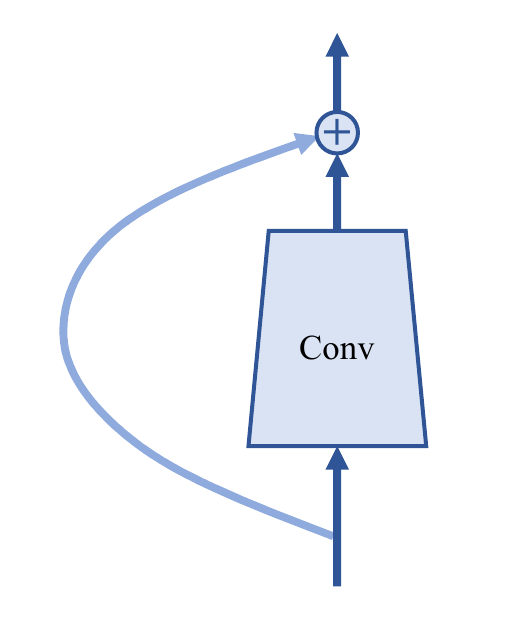}
      \end{minipage}}
 \subfigure[Compressed Conv.]{\label{fig:res-conv}
      \begin{minipage}[t]{0.32\linewidth}
      \centering
      \includegraphics[width=\textwidth]{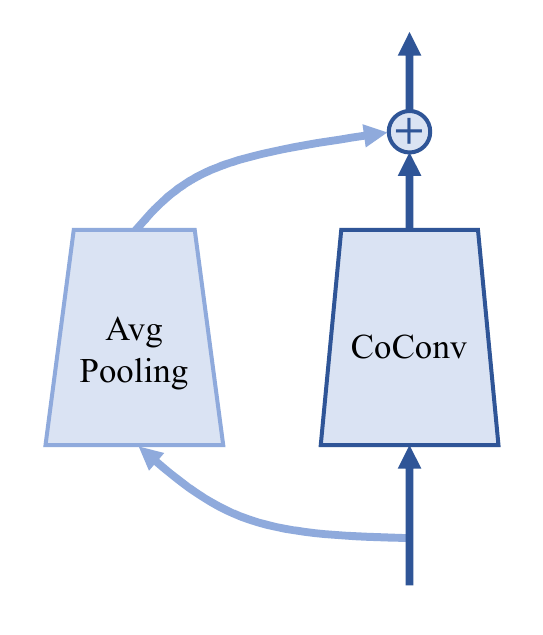}
      \end{minipage}}
 \subfigure[Transposed Conv.]{\label{fig:res-tconv}
      \begin{minipage}[t]{0.3\linewidth}
      \centering
      \includegraphics[width=\textwidth]{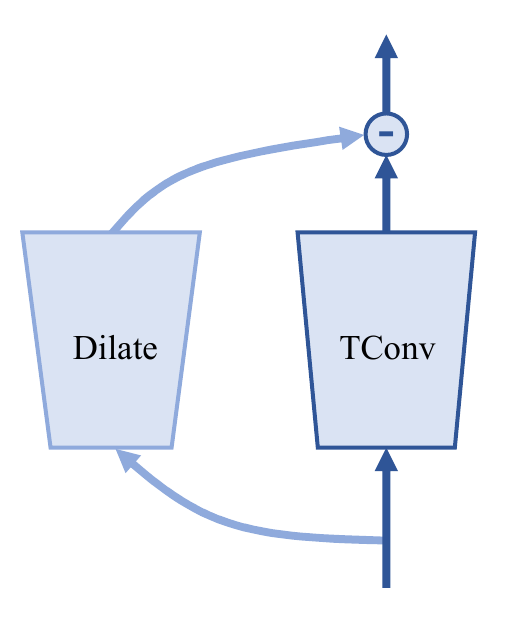}
      \end{minipage}}
\caption{\textbf{Residual Convolution Comparison.} \textit{Conv} denotes Euclidean convolution. \textit{CoConv} denotes compressed convolution. \textit{TConv} denotes transposed convolution.}
\end{figure}

\textbf{Residual Compressed Convolution.}~
The insight behind residual learning~\cite{he_DeepResidualLearning_2016} in Euclidean convolution is that by explicitly decoupling the forward process into identity mapping and learnable mapping (Fig.~\ref{fig:res-ori}), the layer should, at a minimum, preserve the original input features. Specifically, the identity mapping passes on the original inputs as intrinsic features while the learnable mapping extracts the additional features. As a result, a deeper model performs at least as well as its shallower counterpart. Motivated by this key insight, we explore the definition of residual connection in CoCN by answering ``What are the intrinsic features the layer should preserve in compressed convolution?"

Through the $l$-th compressed convolution layer, CoCN extracts hierarchical features where the lower-level collection of node sets $\mathcal{V}^{(l-1)}$ are condensed into higher-level collection $\mathcal{V}^{(l)}$ in a bottom-up manner. The initial collection $\mathcal{V}{(0)}$ is the input node set $\mathcal{V}$, where each node set is a single node. The extracted features indicate what and how the lower-level components contribute to the construction of higher-level node sets. Given individual features $\mathbf{H}^{(l-1)}$ and the corresponding relation features $\mathbf{E}^{(l-1)}$ of the lower-level node sets $s^{(l-1)}$, the extracted output should at least preserve the features $\mathbf{H}^{(l-1)}$ while incrementally learns the topological structures among these constitutive lower-level node sets. Therefore, identity mapping can be substituted with average pooling to preserve individual construction features $\mathbf{H}^{(l-1)}$, as presented in Fig.~\ref{fig:res-conv}. Eq.~\ref{eq:coco-layer-node} can be written as:
\begin{equation}
\begin{aligned}
\mathbf{H}^{(l)} =& \sigma\left(\mathcal{K}\left(\mathbf{E}^{(l-1)}, \mathbf{H}^{(l-1)},k^{(l)},s^{(l)}\right)\right)\\
			     &+\texttt{AvgPool}\left( \mathbf{H}^{(l-1)},k^{(l)},s^{(l)}\right).
\end{aligned}
\end{equation}

In addition to the compressed features, fine-grained features are further required for node-level tasks. In consequence, after producing features of different levels, we then iteratively recover the lower-level features through transposed convolution in a top-down manner. Given the input higher-level features $\hat{\mathbf{H}}^{(l)}$, the target of transposed convolution $\texttt{TConv}(\cdot, k^{(l)},s^{(l)})$ is to decouple lower-level features from higher-level supersets. Different from the incremental learning in compressed convolution layers, transposed convolution learns to remove the redundant features from the intrinsic superset features. Therefore, we substitute the addition-based residual connection with the subtraction-based residual connection, as presented in Fig.~\ref{fig:res-tconv}. The residual transposed convolution can be formulated as:
\begin{equation}
\begin{aligned}
\hat{\mathbf{H}}^{(l-1)} =& \sigma\left(
\texttt{AvgPool}\left(\texttt{Dilat}(\hat{\mathbf{H}}^{(l)}),k^{(l)},s^{(l)}\right)\right)\\
&-\texttt{TConv}\left(\hat{\mathbf{H}}^{(l)},k^{(l)},s^{(l)}\right),
\end{aligned}
\end{equation}
where $\texttt{Dilat}(\cdot)$ denotes the intra-element padding, $\hat{\mathbf{H}}^{(l-1)}$ denotes the recovered fine-grained features. Different from the input features in the bottom-up compression, the recovered features not only contain the individual features of lower-level node sets but also reveal the connection to other node sets at a higher level. 

\begin{figure}[htb]
\centering
\subfigure[Euclidean Conv.]{\label{fig:incep-ori}
      \begin{minipage}[t]{0.48\linewidth}
      \centering
      \includegraphics[width=\textwidth]{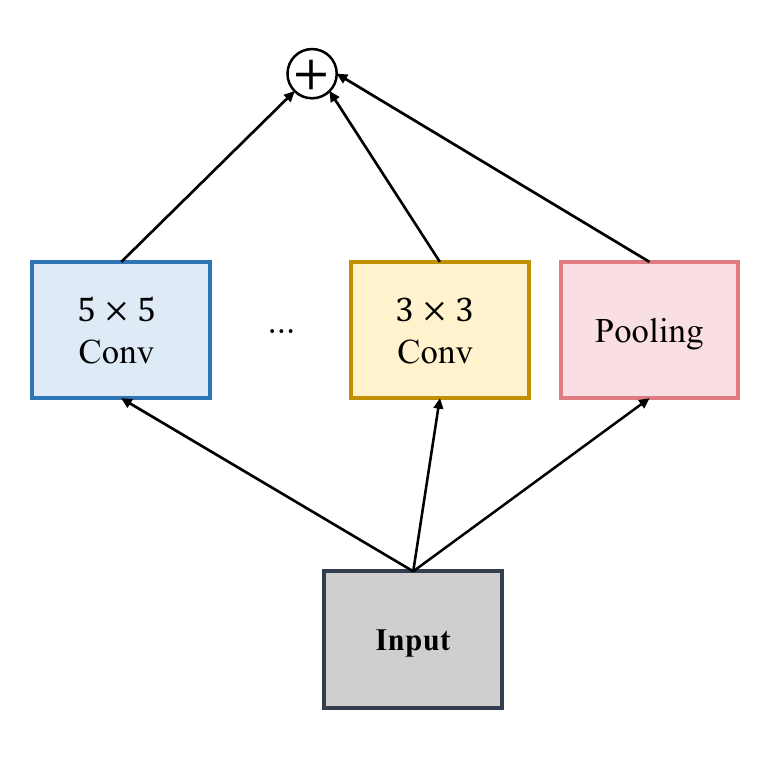}
      \end{minipage}}
 \subfigure[Compressed Conv.]{\label{fig:incep-cocn}
      \begin{minipage}[t]{0.48\linewidth}
      \centering
      \includegraphics[width=\textwidth]{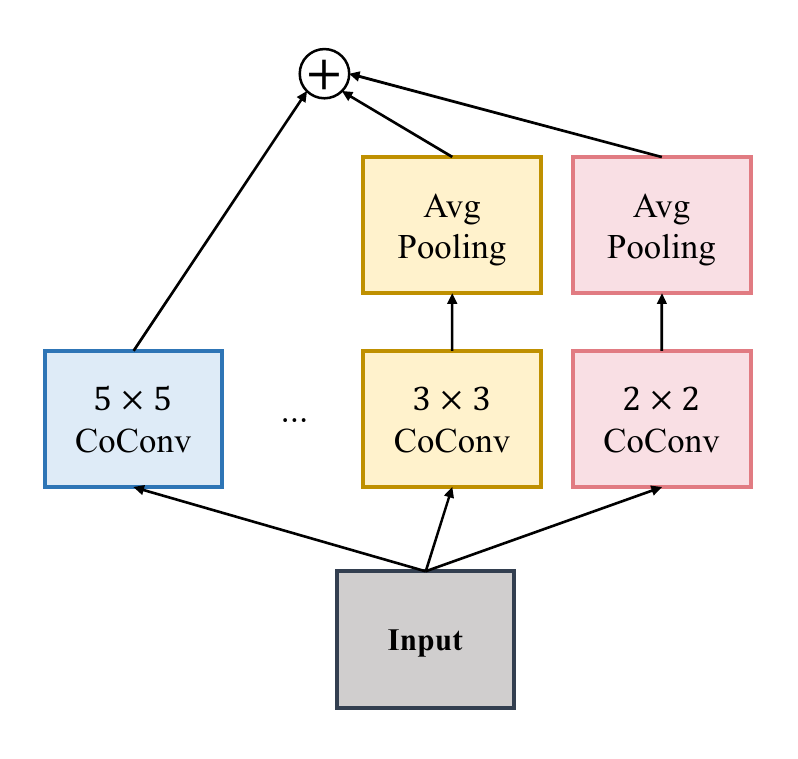}
      \end{minipage}}
\caption{\textbf{Inception Convolution Comparison.} \textit{Conv} denotes Euclidean convolution. \textit{CoConv} denotes compressed convolution.}
\end{figure}

\textbf{Inception Compressed Convolution.}~
Except for the misalignment among the node sets across layers, different sizes of convolution kernels lead to distinct node sets. This misalignment hinders the generalization of inception module~\cite{szegedy_GoingDeeperConvolutions_2015} to CoCN. To bridge the gap among different kernel outputs, we append a post-pooling layer to individual convolution kernels and sum up the pooling result as the layer output. An example of inception compressed convolution module is presented in Fig.~\ref{fig:incep-cocn}. By paralyzing multiple convolution kernels, CoCN can gain a larger receptive field with large kernels while capturing fine-grained local features with small kernels.

\subsection{Model Implementations}
We now present the network structure of CoCN. CoCN mainly contains four modules, \emph{i.e.}, input module, permutation generation module, convolution module, and output module. The detailed network structures of CoCN for graph-level tasks and node-level tasks are presented in the Appendix.
The input module is implemented with a single-layer linear transformation followed by layer normalization and ReLU. The permutation generation module has been described in Subsection~\ref{ssec:permutation-module}. In the convolution module, features permuted under different permutation matrices are learned in parallel, where different permutation heads share the same module parameters. 

For both tasks, the down-sampling convolution block consists of $L_1$ compressed convolution layers with unit step, $L_2$ compressed pooling layers with the same step length as filter size, and an optional single compressed convolution layer with unit step. To recover node representations from multi-scale features, we stack $L_1+L_2+1$ transposed convolution layers~\cite{noh_LearningDeconvolutionNetwork_2015} as an up-sampling convolution block for node-level tasks. 

In the output module of graph-level tasks, CoCN uses max pooling to extract graph-level representation. While for node-level tasks, CoCN re-permutes the extracted features from the $L$-th layer with $\hat{\mathbf{P}}^{\top} \mathbf{H}^{(L)}$ and concatenates the re-permuted features under different permutations to perform the final predictions.

\section{Scalable Compressed Convolution}\label{sec:scalable-cocn}
CoCN achieves permutation invariance through learnable permutations. Compared to enumeration methods~\cite{murphy_JanossyPoolingLearning_2019,murphy_RelationalPoolingGraph_2019,huang_GoingDeeperPermutationSensitive_2022}, the permutation module of CoCN depends on specific tasks to generate the required permutations. As a result, \texttt{CoCN vanilla} and \texttt{CoCN expanded} reduce the time 
complexity of permutation modeling on $n$ nodes from $O(n!)$ to $O(n^2)$, thus enabling tractable node permutation on graphs. However, as the scale of real-world graphs continuously grows, the scalability of GNNs becomes vital for graph learning~\cite{lim_LargeScaleLearning_2022}. To extend CoCN on large-scale graphs, we first remove the relaxation in Eq.~\ref{eq:permutation} and propose the scalable sparse CoCN with sparse permutation matrix and edge features. Furthermore, to ensure the tractability of CoCNs on graphs of various scales, segment CoCN is developed, which learns the order of input nodes globally and performs compressed convolution in a constrained window on the learned node sequence. Both sparse and segment CoCNs are extended from \texttt{CoCN expanded}, and termed unitedly as \texttt{CoCN turbo} in this paper.

\begin{table*}[t]
\caption{\textbf{Complexity Comparison.}}
\label{tab:complexity}
\begin{center}
\begin{threeparttable}
\begin{small}
\begin{tabular}{llccc} 
\hline
&                         & Expanded       & Sparse        & Segment                                  \\ 
\hline
\multirow{4}{*}{\begin{tabular}[c]{@{}l@{}}Time\\Comlexity\end{tabular}} & App. Position Reg.           & \multicolumn{3}{c}{$O(m)$~[\ref{eq:approximate-ex-pos-implementation}] or  $O(n)$~[\ref{eq:approximate-im-pos-implementation}]}                                              \\
 & Abs. Position Reg.        & $O(n^2)$~[\ref{eq:absolute-pos}]      & $O(nlogn)$~[$\dagger$] & $O(nlogn)+O(b^2n)\rightarrow O(nlogn)+O(b^2n_b)$~[$\dagger$,~\ref{eq:absolute-pos}] \\
 & Permutation Matrix Gen. & $O(n^2)$~[\ref{eq:permutation}]      & $O(n)$~[\ref{eq:sparse-permutation}]        & $O(b^2n)\rightarrow O(b^2n_b)$~[\ref{eq:permutation}]           \\
 & Convolution             & $O(n^3)$      &  $\mathtt{max}(O(m),O(n))$     & $O(b^3n)\rightarrow O(b^3n_b)$           \\ 
\hline
\multicolumn{2}{l}{Space Complexity}                       & $O(n^2)+O(m)$ & $O(n)+O(m)$   & $O(b^2n)+O(m)\rightarrow O(b^2n_b)+O(m)$  \\
\hline
\end{tabular}
\end{small}
\begin{tablenotes}
    \item  $n$ and $m$ denote the number of input nodes and edges, respectively. $b$ denotes the length of segments. $n_b$ denotes the number of nodes tackled in a single epoch for segment CoCN. $\dagger$ represents that the corresponding time complexity depends on the ranking algorithm.
\end{tablenotes}
\end{threeparttable}
\end{center}
\end{table*}

\subsection{Sparse Compressed Convolution Networks}
The computational bottleneck of CoCN originates from the permutation generation module. To enable differentiable graph calibration, this module performs relaxation on the permutation matrix $\hat{\mathbf{P}}$. The computation of $\hat{\mathbf{P}}$ involves dense pairwise calculations in Eq.~\ref{eq:absolute-pos} and Eq.~\ref{eq:permutation}. As a result, the output dense permutation matrix further produces dense edge features, imposing heavy computation demands on subsequent modules. To overcome the computational bottleneck, improvements on the permutation module are required.

For the absolute position $\mathbf{r}$ in Eq.~\ref{eq:absolute-pos}, we derive the refined forward and backward computation separately. In forward propagation, a discrete ranking algorithm $\texttt{Rank}(\cdot)$ is employed to replace the pairwise computation. For backpropagation, the original Eq.~\ref{eq:absolute-pos} can be reformulated as
\begin{equation}
\begin{matrix}
    \texttt{(original)} &
    \begin{aligned}
        \frac{\partial\mathbf{r}}{\partial\mathbf{r}_A} &= 
        \frac{\partial}{\partial\mathbf{r}_A}
        \texttt{sigmoid}\left(\mathbf{C}\right)\mathbf{1},\\
        \mathbf{C} &= \texttt{ReLU}\left(\mathbf{r}_A \mathbf{1}^\top-\mathbf{1}\mathbf{r}_A^\top\right),
    \end{aligned}
\end{matrix}
\end{equation}
where $\mathbf{C}$ denotes the dense pairwise position value comparison result. The sigmoid function with ReLU is adopted to approximate the sign function. For each node, only nodes with lower ranking values contribute to the original result. To avoid the dense pairwise computation in $\mathbf{C}$, first-order Taylor expansion is employed to approximate the sigmoid function. Therefore, the backpropagation can be reformulated as
\begin{equation}\label{eq:approximate-back}
\begin{matrix}
    \texttt{(approx.)} &
    \begin{aligned}
        \frac{\partial\mathbf{r}}{\partial\mathbf{r}_A} &= 
        \frac{\partial}{\partial\mathbf{r}_A} 
        \ \frac{1}{4}\hat{\mathbf{r}},\\
        \hat{\mathbf{r}}_i &= \mathbf{r}_i\mathbf{r}_{Ai} -
        \quad\mathclap{\ \sum_{j \in \!\{c|\mathbf{r}_c <\mathbf{r}_i\!\} }}
        \quad\mathbf{r}_{Aj}.
    \end{aligned}
\end{matrix}
\end{equation}
For the detailed derivation of Eq.~\ref{eq:approximate-back}, please refer to the Appendix.
The space complexity of Eq.~\ref{eq:absolute-pos} is now reduced to $O(n)$ using Eq.~\ref{eq:approximate-back}. The time complexity depends on the discrete ranking algorithms that are computationally efficient with GPU implementations ~\cite{paszke_PyTorchImperativeStyle_2019}.

For Eq.~\ref{eq:permutation}, the relaxation factor $\tau$ controls the approximation level to the standard permutation matrix. According to the empirical practice in Tab.~\ref{tab:ablation-tau} on small-scale benchmarks, the optimal value of $\tau$ is around $1$ to $10$. In consequence, most entries in $\hat{\mathbf{P}}$ are close to 0, empowering the usage of standard permutation matrices in CoCN:
\begin{equation}\label{eq:sparse-permutation}
    \mathbf{P}_{ij} = 
    \begin{cases}
        \texttt{exp}\left\{\texttt{Rank}(\mathbf{r}_A)_j - \mathbf{r}_j\right\}, &\text{if $\texttt{Rank}(\mathbf{r}_A)_j = i$;}\\
        0, &\text{otherwise,}
    \end{cases}
\end{equation}
where $\texttt{Rank}(\mathbf{r}_A)$ denotes the discrete ranking result, and $\mathbf{r}$ denotes the ranking result with approximate gradient in Eq.~\ref{eq:approximate-back}. $\mathbf{P}$ is a sparse matrix, demanding $O(n)$ space complexity.

\begin{figure}[htb]
\centering
\includegraphics[width=0.5\textwidth]{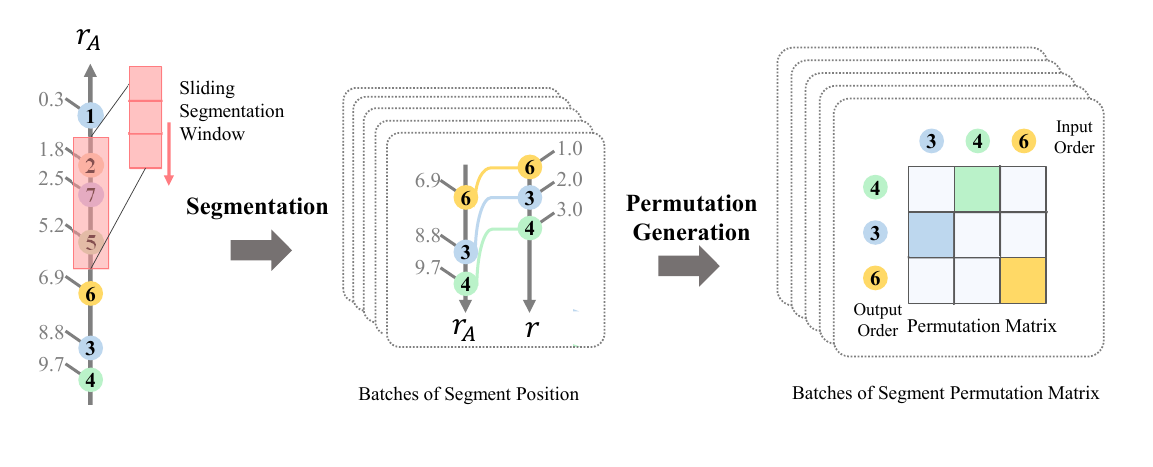}
\caption{\textbf{Segmentation on the Node Sequence.}}\label{fig:segment}
\end{figure}

\subsection{Segment Compressed Convolution Networks}
As the scale of real-world graphs continuously increases, general GNNs may face limited capacity of computational equipment. To ensure the tractability of CoCN on various scales of input graphs, we further slice the permuted node sequence into segments of length $b$. More specifically, the input nodes can be first transformed into a permuted node sequence with a discrete sorting algorithm. A sliding window of size $b$ is applied to the sequence with a unit sliding step, as presented in Fig.~\ref{fig:segment}. To ensure the number of generated segments equaling the number of input nodes, circular padding is employed on the node sequence. Each segment constitutes a segment graph with features $\mathbf{X}_b\in\mathbb{R}^{b\times d}$ and $\mathbf{A}_b\in\mathbb{R}^{b\times b}$ extracted from the input graphs. Given a segment with approximate position $\mathbf{r}_A^{\texttt{seg}}\in\mathbb{R}^{b}$, CoCN then performs permutation and convolution on the segment level. It should be noted that as the position regression process is computationally efficient, the approximate positions can be computed on the full-graph level before the segmentation. 

For both node-level and graph-level tasks, segment CoCN follows the graph-level network structure.
In node-level tasks, the generated segment graph can be regarded as the property graph of the first node on the segment. Thus, the pooled output representation for a given segment graph is utilized to make predictions for the corresponding node.

\begin{table*}[t]
\caption{\textbf{Graph Classification Results (measured by accuracy: \%).}}
\label{tab:graph}
\begin{center}
\begin{small}
\begin{sc}
\begin{tabular}{lcccccc} 
  \toprule
                       & MUTAG                   & PROTEINS                & NCI1                    & IMDB-B                  & IMDB-M                  & COLLAB                   \\ 
  \hline
  graphs               & 188                     & 1,113                   & 4,110                   & 1,000                   & 1,500                   & 5,000                    \\
  avg nodes            & 17.93                   & 39.06                   & 29.87                   & 19.77                   & 13                      & 74.5                     \\
  avg edges            & 39.6                    & 145.6                   & 64.6                    & 193.1                   & 131.87                  & 4,914.4                  \\
  node features        & 7                       & 3                       & 37                      & 0                       & 0                       & 0                        \\ 
  \hline
  PATCHY-SAN~\cite{niepert_LearningConvolutionalNeural_2016}      & 88.95$_{\color{gray}\pm4.21}$          & 75.00$_{\color{gray}\pm2.51}$          & 78.60$_{\color{gray}\pm1.90}$          & 71.00$_{\color{gray}\pm2.29}$          & 45.23$_{\color{gray}\pm2.84}$          & 72.60$_{\color{gray}\pm2.15}$           \\
  GCN~\cite{kipf_SemiSupervisedClassificationGraph_2017}          & 69.50$_{\color{gray}\pm1.78}$          & 73.24$_{\color{gray}\pm0.73}$          & 76.29$_{\color{gray}\pm1.79}$          & 73.26$_{\color{gray}\pm0.46}$          & 50.39$_{\color{gray}\pm0.41}$          & 80.59$_{\color{gray}\pm0.27}$           \\
  GIN~\cite{xu*_HowPowerfulAre_2019}                              & 81.39$_{\color{gray}\pm1.53}$          & 71.46$_{\color{gray}\pm1.66}$          & 80.00$_{\color{gray}\pm1.40}$          & 72.78$_{\color{gray}\pm0.86}$          & 48.13$_{\color{gray}\pm1.36}$          & 78.19$_{\color{gray}\pm0.63}$           \\
  GraphSAGE~\cite{hamilton_InductiveRepresentationLearning_2017}  & 83.60$_{\color{gray}\pm9.60}$          & 73.00$_{\color{gray}\pm4.50}$          & 76.00$_{\color{gray}\pm1.80}$          & 68.80$_{\color{gray}\pm4.50}$          & 47.60$_{\color{gray}\pm3.50}$          & 73.90$_{\color{gray}\pm1.70}$           \\
  PG-GNN~\cite{huang_GoingDeeperPermutationSensitive_2022}        & -                                      & 76.80$_{\color{gray}\pm3.80}$          & 82.80$_{\color{gray}\pm1.30}$          & 76.80$_{\color{gray}\pm2.60}$          & 53.20$_{\color{gray}\pm3.60}$          & 80.90$_{\color{gray}\pm0.80}$           \\
  PathNet~\cite{michel_PathNeuralNetworks_2023}                   & -                                      & 70.50$_{\color{gray}\pm3.90}$          & 64.10$_{\color{gray}\pm2.30}$          & 70.40$_{\color{gray}\pm3.80}$          & 49.10$_{\color{gray}\pm3.60}$          & -                                       \\
  PathNN~\cite{sun_HomophilyStructureawarePath_2022}              & -                                      & 75.20$_{\color{gray}\pm3.90}$          & 82.30$_{\color{gray}\pm1.90}$          & 72.60$_{\color{gray}\pm3.30}$          & 50.80$_{\color{gray}\pm4.50}$          & -                                       \\
  TopKPool~\cite{gao_GraphUNets_2019}                             & 67.61$_{\color{gray}\pm3.36}$          & 70.48$_{\color{gray}\pm1.01}$          & 67.02$_{\color{gray}\pm2.25}$          & 71.58$_{\color{gray}\pm0.95}$          & 48.59$_{\color{gray}\pm0.72}$          & 77.58$_{\color{gray}\pm0.85}$           \\
  SAGPool~\cite{lee_SelfAttentionGraphPooling_2019}               & 73.67$_{\color{gray}\pm4.28}$          & 71.56$_{\color{gray}\pm1.49}$          & 67.45$_{\color{gray}\pm1.11}$          & 72.55$_{\color{gray}\pm1.28}$          & 50.23$_{\color{gray}\pm0.44}$          & 78.03$_{\color{gray}\pm0.31}$           \\
  ASAP~\cite{ranjan_ASAPAdaptiveStructure_2020}                   & 77.83$_{\color{gray}\pm1.49}$          & 73.92$_{\color{gray}\pm0.63}$          & 71.48$_{\color{gray}\pm0.42}$          & 72.81$_{\color{gray}\pm0.50}$          & 50.78$_{\color{gray}\pm0.75}$          & 78.64$_{\color{gray}\pm0.50}$           \\
  DiffPool~\cite{ying_HierarchicalGraphRepresentation_2018}       & 79.22$_{\color{gray}\pm1.02}$          & 76.25$_{\color{gray}\pm1.00}$          & 62.32$_{\color{gray}\pm1.90}$          & 73.14$_{\color{gray}\pm0.70}$          & 51.31$_{\color{gray}\pm0.72}$          & 82.13$_{\color{gray}\pm0.43}$           \\
  SEP~\cite{wu_StructuralEntropyGuided_2022}                      & 85.56$_{\color{gray}\pm1.09}$          & 76.42$_{\color{gray}\pm0.39}$          & 79.35$_{\color{gray}\pm0.33}$          & 74.12$_{\color{gray}\pm0.56}$          & 51.53$_{\color{gray}\pm0.65}$          & 81.28$_{\color{gray}\pm0.15}$           \\
  GMT~\cite{baek_AccurateLearningGraph_2022}                      & 83.44$_{\color{gray}\pm1.33}$          & 75.09$_{\color{gray}\pm0.59}$          & 76.35$_{\color{gray}\pm2.62}$          & 73.48$_{\color{gray}\pm0.76}$          & 50.66$_{\color{gray}\pm0.82}$          & 80.74$_{\color{gray}\pm0.54}$           \\ 
  \hline
  \textbf{CoCN vanilla~(Ours)} & 87.08$_{\color{gray}\pm0.17}$ & 76.86$_{\color{gray}\pm0.13}$ & 82.89$_{\color{gray}\pm0.19}$ & 77.26$_{\color{gray}\pm0.27}$ & 56.32$_{\color{gray}\pm0.18}$ & 86.15$_{\color{gray}\pm0.10}$  \\
  \textbf{CoCN expanded~(Ours)} & \textbf{89.89$_{\color{gray}\pm0.39}$} & \textbf{79.48$_{\color{gray}\pm0.51}$} & \textbf{85.83$_{\color{gray}\pm0.86}$} & \textbf{77.94$_{\color{gray}\pm0.33}$} & \textbf{56.73$_{\color{gray}\pm0.36}$} & \textbf{87.22$_{\color{gray}\pm0.13}$}  \\
  \bottomrule
  \end{tabular}
\end{sc}
\end{small}
\end{center}
\end{table*}

\subsection{Complexity Analysis}
The theoretical complexity comparison among \texttt{CoCN expanded}, sparse CoCN, and segment CoCN is presented in Tab.~\ref{tab:complexity}. We note that the computational and space complexity of \texttt{CoCN expanded} stay in line with \texttt{CoCN vanilla}. Therefore, only \texttt{CoCN expanded} is analyzed in this section. We further conduct empirical complexity analysis in Section~\ref{ssec:ablation}. The space complexity is determined by the edge feature matrix and permutation matrix. Let $m$ denote the number of input edges. As the real-world graphs are sparsely connected~\cite{boguna_NetworkGeometry_2021}, the edge features of input graphs can be constructed as sparse matrices with space complexity of $O(m)$. In consequence, the approximate position regression for nodes with explicit features in Eq.~\ref{eq:approximate-ex-pos-implementation} can be performed with time complexity of $O(m)$. In the regression with implicit node features, Eq.~\ref{eq:approximate-im-pos-implementation} can be computed before the training procedure, resulting in a time complexity of $O(n)$ during training.

For the absolute position regression, the required time complexity of Eq.~\ref{eq:absolute-pos} in the \texttt{CoCN expanded} is $O(n^2)$. In sparse CoCN, this can be reduced through a discrete ranking algorithm with an average time complexity of $O(nlogn)$. In segment CoCN, the sliding segmentation process on the sorted node sequence gives rise to $n$ segments with a single segment of length $b$. However, as the number of input nodes $n$ can be extremely large, the resulting space complexity of $O(b^2n)$ in Eq.~\ref{eq:absolute-pos} becomes intractable. To address this problem, we adopt mini-batching on the input nodes after segmentation and pass on $n_b$ nodes with their corresponding segments to the subsequent modules.
For permutation matrix generation, \texttt{CoCN expanded} demands a time complexity of $O(n^2)$ in Eq.~\ref{eq:permutation}. In contrast, sparse CoCN removes the relaxation on the permutation matrix and achieves $O(n)$ time complexity in Eq.~\ref{eq:sparse-permutation}.

The computational bottleneck in the convolution module originates from the permutation process, where matrix multiplications of $\hat{\mathbf{X}}=\hat{\mathbf{P}}\mathbf{X}$ and $\hat{\mathbf{A}}=\hat{\mathbf{P}}\mathbf{A}\hat{\mathbf{P}}^\top$ have time complexity of $O(n^3)$. This can be reduced to $\mathtt{max}(O(m),O(n))$ with the sparse permutation matrix and $O(b^3n_b)$ with mini-batching segments.

\section{Experiments}\label{sec:experiment}
We empirically evaluate CoCN on real-world benchmarks. The code of CoCN is implemented with PyTorch~\cite{paszke_PyTorchImperativeStyle_2019} and PyTorch-Geometric~\cite{fey_FastGraphRepresentation_2019}. All the experiments are conducted on a single Nvidia Geforce RTX 4090. The detailed model structure, baseline, and hyper-parameter settings are delineated in the Appendix.
CoCNs with plain compressed convolution are denoted as \texttt{CoCN vanilla}, while CoCNs with compositional compressed convolution and the general node regression are denoted as \texttt{CoCN expanded}. In the ablation, experiments are based on the basic CoCN framework and augmented with specific modules.

\subsection{Graph-level Tasks}\label{ssec:graph-class}
\subsubsection{Experimental Setups}
For graph-level tasks, we evaluate CoCN on graph classification benchmarks and graph isomorphism benchmarks. In graph classification, we use nine benchmarks including three biochemical benchmarks~\cite{Morris+2020} (MUTAG, PROTEINS, and NCI1), three social network benchmarks~\cite{Morris+2020} (COLLAB, IMDB-BINARY, and IMDB-MULTI), and three brain connectomics benchmarks~\cite{said_NeuroGraphBenchmarksGraph_2023} (HCP-Task, HCP-Gender, and HCP-Age). In the graph isomorphism test, we use three benchmarks including Graph8c, sr25~\cite{balcilar_BreakingLimitsMessage_2021}, and EXP~\cite{abboud_SurprisingPowerGraph_2021}. For biochemical and social network benchmarks, we perform 10-fold cross-validation following~\cite{xu*_HowPowerfulAre_2019} and report average performance. Since COLLAB, IMDB-BINARY, and IMDB-MULTI have no input node features, we follow the common practice~\cite{xu*_HowPowerfulAre_2019} and use the one-hot encoding of node degrees as node features. For graph isomorphism test benchmarks with implicit node features, we employ Eq.~\ref{eq:approximate-im-pos-implementation} to obtain node positions. The statistics of these benchmarks are summarized in Tab.~\ref{tab:graph}-~\ref{tab:graph-iso}.


\subsubsection{Performance Analysis}
\textbf{Performance on Graph Classification.}~
The results of CoCN and baselines for graph classification are presented in Tab.~\ref{tab:graph} and Tab.~\ref{tab:graph-brain}. CoCN surpasses convolutional GNNs and graph pooling methods on all benchmarks. Especially on social network benchmarks like COLLAB that do not provide any node features, \texttt{CoCN vanilla} and \texttt{CoCN expanded} outperform baseline models by 1.71\% and 3.75\% on average, respectively. Compared to information-preserving graph pooling models SEP and GMT, CoCN benefits from its explicit structure learning ability and extracts more discriminative features solely from the input structures. The structure learning ability of CoCN can be further validated in the graph isomorphism test and the ablation study~(Section \ref{ssec:ablation}). For MUTAG, only PATCHY-SAN outperforms \texttt{CoCN vanilla} but it becomes less promising compared to \texttt{CoCN expanded}. PATCHY-SAN is very unstable across repeat tests and degrades a lot on COLLAB and IMDB-M due to the lack of edge feature learning. In contrast, CoCN achieves better and more stable performance on the rest of the benchmarks. In Section~\ref{ssec:ablation}, we further demonstrate that CoCN can achieve comparable performance to PATCHY-SAN with only structure features on MUTAG.

Between \texttt{CoCN vanilla} and \texttt{CoCN expanded}, \texttt{CoCN expanded} consistently achieves better performance, indicating more effectiveness of the compositional compressed convolution against its plain counterpart. We further validate our compositional compressed convolution on brain connectomics classification tasks against the Euclidean convolution model. As presented in Tab.~\ref{tab:graph-brain}, \texttt{CoCN expanded} surpasses CNN by 4.77\%, indicating its ability to effectively tackle graph classification tasks.

\begin{table}[t]
\caption{\textbf{Graph Classification Results on Brain Connectomics Benchmarks (measured by accuracy: \%).}}
\label{tab:graph-brain}
\vspace{-0.1in}
\begin{center}
\begin{threeparttable}
\begin{small}
\begin{tabular}{lccc} 
\toprule
              & HCP-Task       & HCP-Gender     & HCP-Age         \\ 
\midrule
Graphs        & 7,443           & 1,078           & 1,065            \\
Avg Nodes     & 400            & 1,000           & 1,000            \\
Avg Edges     & 7,029.18        & 45,578.61       & 45,588.40        \\
Node Features & 400            & 1,000           & 1,000            \\ 
\midrule
CNN~\cite{said_NeuroGraphBenchmarksGraph_2023}           & 95.88          & 76.39          & 44.23           \\
RF~\cite{said_NeuroGraphBenchmarksGraph_2023}            & 88.98          & 69.90          & 43.38           \\
k-GNN~\cite{morris_WeisfeilerLemanGo_2019}         & 93.23          & 82.13          & 40.84           \\
GCN~\cite{kipf_SemiSupervisedClassificationGraph_2017}           & 94.21          & 75.46          & 42.72           \\
ResGCN~\cite{li_DeepGCNsCanGCNs_2019}        & 94.61          & 78.33          & 43.85           \\
GIN~\cite{xu*_HowPowerfulAre_2019}           & 89.79          & 75.56          & 40.00           \\
ChebNet~\cite{defferrard_ConvolutionalNeuralNetworks_2016}       & 94.45          & 59.07          & 44.98           \\
GAT~\cite{velickovic_GraphAttentionNetworks_2018}           & 95.02          & 76.20          & 41.97           \\
\midrule
\textbf{CoCN exp. (Ours)}   & \textbf{96.44} & \textbf{83.80} & \textbf{46.01}  \\
\bottomrule
\end{tabular}
\end{small}
\begin{tablenotes}
    \item CoCN exp. denotes CoCN expanded.
\end{tablenotes}
\end{threeparttable}
\vspace{-0.2in}
\end{center}
\end{table}
\begin{table}[t]
\caption{\textbf{Graph Isomorphism Test (measured by number of undistinguished non-isomorphic graph pairs).}}
\label{tab:graph-iso}
\vskip -0.1in
\setlength{\tabcolsep}{13pt}
\begin{center}
\begin{threeparttable}
\begin{small}
\begin{tabular}{lccc} 
\hline
                        & Graph8c    & sr25       & EXP         \\
\# non-iso. pairs & 61,000,000 & 105        & 600         \\ 
\toprule
MLP~\cite{balcilar_BreakingLimitsMessage_2021}                     & 293,000    & 105        & 600         \\
GCN~\cite{kipf_SemiSupervisedClassificationGraph_2017}                     & 4,775      & 105        & 600         \\
GAT~\cite{velickovic_GraphAttentionNetworks_2018}                     & 1,828      & 105        & 600         \\
GIN~\cite{xu*_HowPowerfulAre_2019}                     & 386        & 105        & 600         \\
ChebNet~\cite{defferrard_ConvolutionalNeuralNetworks_2016}                 & 44         & 105        & 71          \\
PPGN~\cite{maron_ProvablyPowerfulGraph_2019}                    & 0          & 105        & 0           \\
GNNML~\cite{balcilar_BreakingLimitsMessage_2021}                   & 0          & 105        & 0           \\ 
\midrule
\textbf{CoCN exp.~(Ours)}             & \textbf{0} & \textbf{0} & \textbf{0}  \\
\bottomrule
\end{tabular}
\end{small}
\begin{tablenotes}
    \item CoCN exp. denotes CoCN expanded.
\end{tablenotes}
\end{threeparttable}
\end{center}
\end{table}

\begin{table*}[]
\caption{\textbf{Node Classification Results (measured by accuracy: \%).}}
\vspace{-0.1in}
\label{tab:node}
\begin{center}
\begin{small}
\begin{sc}
\begin{tabular}{lccccccc} 
  \toprule
                                         & Chameleon               & Squirrel                & Cornell                 & Texas                   & Wisconsin               & Actor                   & \multirow{4}{*}{Avg Rank}  \\ 
  \cline{1-7}
  Nodes                                  & 2,277                   & 5,201                   & 198                     & 183                     & 251                     & 7,600                   &                            \\
  Edges                                  & 36,101                  & 198,493                 & 295                     & 309                     & 499                     & 33544                   &                            \\
  Features                               & 2,325                   & 2,089                   & 1,703                   & 1,703                   & 1,703                   & 932                     &                            \\ 
  \hline
  GCN~\cite{kipf_SemiSupervisedClassificationGraph_2017}          & 59.82$_{\color{gray}\pm2.58}$          & 36.89$_{\color{gray}\pm1.34}$          & 57.03$_{\color{gray}\pm4.67}$          & 59.46$_{\color{gray}\pm5.25}$          & 59.80$_{\color{gray}\pm6.99}$          & 30.26$_{\color{gray}\pm0.79}$          & 12.17  \\
  MixHop~\cite{abu-el-haija_MixHopHigherOrderGraph_2019}          & 60.50$_{\color{gray}\pm2.53}$          & 43.80$_{\color{gray}\pm1.48}$          & 73.51$_{\color{gray}\pm6.34}$          & 77.84$_{\color{gray}\pm7.73}$          & 75.88$_{\color{gray}\pm4.90}$          & 32.22$_{\color{gray}\pm2.34}$          & 9.33                       \\
  ChebNet~\cite{defferrard_ConvolutionalNeuralNetworks_2016}      & 55.24$_{\color{gray}\pm2.76}$          & 43.86$_{\color{gray}\pm1.64}$          & 74.30$_{\color{gray}\pm7.46}$          & 77.30$_{\color{gray}\pm4.07}$          & 79.41$_{\color{gray}\pm4.46}$          & 34.11$_{\color{gray}\pm1.09}$          & 9.17                       \\
  GraphSAGE~\cite{hamilton_InductiveRepresentationLearning_2017}  & 49.06$_{\color{gray}\pm1.88}$          & 36.73$_{\color{gray}\pm1.21}$          & 80.08$_{\color{gray}\pm2.96}$          & 82.03$_{\color{gray}\pm2.77}$          & 81.36$_{\color{gray}\pm3.91}$          & 35.07$_{\color{gray}\pm0.15}$          & 8.00                       \\
  Geom-GCN~\cite{pei_GeomGCNGeometricGraph_2020}                  & 60.00$_{\color{gray}\pm2.81}$          & 38.15$_{\color{gray}\pm0.92}$          & 60.54$_{\color{gray}\pm3.67}$          & 66.76$_{\color{gray}\pm2.72}$          & 64.51$_{\color{gray}\pm3.66}$          & 31.59$_{\color{gray}\pm1.15}$          & 11.17                      \\
  FAGCN~\cite{bo_LowfrequencyInformationGraph_2021}               & 46.07$_{\color{gray}\pm2.11}$          & 30.83$_{\color{gray}\pm0.69}$          & 76.76$_{\color{gray}\pm5.87}$          & 76.49$_{\color{gray}\pm2.87}$          & 79.61$_{\color{gray}\pm1.58}$          & 34.82$_{\color{gray}\pm1.35}$          & 10.17                       \\
  APPNP~\cite{gasteiger_PredictThenPropagate_2018}                & 40.44$_{\color{gray}\pm2.02}$          & 29.20$_{\color{gray}\pm1.45}$          & 56.76$_{\color{gray}\pm4.58}$          & 55.10$_{\color{gray}\pm6.23}$          & 54.59$_{\color{gray}\pm6.13}$          & 30.02$_{\color{gray}\pm0.89}$          & 14.00                      \\
  LINKX~\cite{lim_LargeScaleLearning_2021}                        & 68.42$_{\color{gray}\pm1.38}$          & 61.81$_{\color{gray}\pm1.80}$          & 77.84$_{\color{gray}\pm5.81}$          & 74.60$_{\color{gray}\pm8.37}$          & 75.49$_{\color{gray}\pm5.72}$          & 36.10$_{\color{gray}\pm1.55}$          & 7.17                       \\
  GGCN~\cite{yan_TwoSidesSame_2022}                               & 71.14$_{\color{gray}\pm1.84}$          & 55.17$_{\color{gray}\pm1.58}$          & 85.68$_{\color{gray}\pm6.63}$          & 84.86$_{\color{gray}\pm4.55}$          & 86.86$_{\color{gray}\pm3.29}$          & \textbf{37.54$_{\color{gray}\pm1.56}$}                & 2.50                       \\
  ACMII-GCN~\cite{luan_RevisitingHeterophilyGraph_2022}           & 68.46$_{\color{gray}\pm1.70}$          & 51.80$_{\color{gray}\pm1.50}$          & 85.95$_{\color{gray}\pm5.64}$          & \textbf{86.76$_{\color{gray}\pm4.75}$}  & \textbf{87.45$_{\color{gray}\pm3.74}$}  & 36.43$_{\color{gray}\pm1.20}$          & 2.33                       \\
  GPRGCN~\cite{chien_AdaptiveUniversalGeneralized_2022}           & 62.59$_{\color{gray}\pm2.04}$          & 46.31$_{\color{gray}\pm2.46}$          & 78.11$_{\color{gray}\pm6.55}$          & 81.35$_{\color{gray}\pm5.32}$          & 82.55$_{\color{gray}\pm6.23}$          & 35.16$_{\color{gray}\pm0.90}$          & 6.17                       \\
  GCNII~\cite{chen_SimpleDeepGraph_2020}                          & 63.86$_{\color{gray}\pm3.04}$          & 38.47$_{\color{gray}\pm1.58}$          & 77.86$_{\color{gray}\pm3.79}$          & 80.39$_{\color{gray}\pm3.83}$          & 77.57$_{\color{gray}\pm3.40}$          & 37.44$_{\color{gray}\pm1.30}$          & 6.50                       \\
  AERO-GNN\cite{lee_DeepAttentionGraph_2023}                      & 71.58$_{\color{gray}\pm2.40}$          & 61.76$_{\color{gray}\pm2.40}$          & 81.24$_{\color{gray}\pm6.80}$          & 84.35$_{\color{gray}\pm5.20}$          & 84.80$_{\color{gray}\pm3.30}$          & 36.57$_{\color{gray}\pm1.10}$          & 3.33                       \\ 
  \hline
  \textbf{CoCN vanilla (Ours)}                                    & 79.17$_{\color{gray}\pm0.17}$         & 72.95$_{\color{gray}\pm0.23}$         & \textbf{86.22$_{\color{gray}\pm0.49}$} & 85.21$_{\color{gray}\pm0.49}$          & 86.88$_{\color{gray}\pm0.45}$          & 36.35$_{\color{gray}\pm0.12}$          & 1.67                 \\
  \textbf{CoCN expanded (Ours)}                                            & \textbf{79.35$_{\color{gray}\pm1.42}$} & \textbf{72.98$_{\color{gray}\pm1.78}$} & 83.39$_{\color{gray}\pm1.65}$          & 84.90$_{\color{gray}\pm1.05}$          & 86.92$_{\color{gray}\pm1.41}$          & 36.62$_{\color{gray}\pm0.85}$          & \textbf{1.33}                 \\
  \bottomrule
  \end{tabular}
\end{sc}
\end{small}
\end{center}
\end{table*}

\textbf{Performance on Graph Isomorphism Test.}~
To evaluate CoCN on graph structure learning, we conduct experiments on graph isomorphism tests. The numbers of undistinguished graph pairs of CoCN and baselines are presented in Tab.~\ref{tab:graph-iso}. Among the baseline models, PPGN and GNNML that are expressively equivalent to 3-WL (Weisfeiler-Lehman) gain the best results while failing to distinguish any non-isomorphic graph pairs in sr25. In contrast, CoCN successfully discriminates all the non-isomorphic graph pairs, owing to its explicit structure learning on the input graphs. This indicates the effectiveness of CoCN in graph structure learning.

\subsection{Node-level Tasks}\label{ssec:node-class}
\subsubsection{Experimental Setups}
For node classification, we conduct experiments on nine benchmarks including small-scale benchmarks (Chameleon, Squirrel~\cite{rozemberczki_MultiScaleAttributedNode_2021}, Cornell, Texas, Wisconsin~\cite{pei_GeomGCNGeometricGraph_2020}, and Actor~\cite{tang_SocialInfluenceAnalysis_2009}), large-scale benchmarks (questions, amazon-ratings, tolokers, and minesweeper)~\cite{platonov_CriticalLookEvaluation_2023}, (CoauthorCS, CoauthorPhysics, AmazonComputers, and AmazonPhoto)~\cite{shchur_PitfallsGraphNeural_2019} and genius~\cite{lim_LargeScaleLearning_2022}. For Chameleon, Squirrel, Cornell, Texas, and Wisconsin, we use the same random train/validation/test splits of 48\%/32\%/20\% as~\cite{pei_GeomGCNGeometricGraph_2020}\footnote{The original random splits in~\cite{pei_GeomGCNGeometricGraph_2020} is 60\%/20\%/20\%, which is different from their open source data splits.} and report average performance over ten splits. For the rest benchmarks, we follow the original experimental settings~\cite{lim_LargeScaleLearning_2022, shchur_PitfallsGraphNeural_2019, platonov_CriticalLookEvaluation_2023}. 
We also conduct experiments on PCQM-Contact~\cite{dwivedi_LongRangeGraph_2022} to explore the potential of CoCN on the link prediction task. 
The statistics of node-level benchmarks are summarized in Tab.~\ref{tab:node}-\ref{tab:node-homo}.

\subsubsection{Performance Analysis}
\textbf{Performance on small-scale benchmarks.}~
The results of CoCN and baselines are presented in Tab.~\ref{tab:node}.  CoCN gets competitive results on all six benchmarks and outperforms other baseline models on Chameleon, Squirrel, and Cornell. For Texas, Wisconsin, and Actor, \texttt{CoCN vanilla} and \texttt{CoCN expanded} slightly underperform GGCN and ACMII-GCN. This is because the node connectivity is weak on these benchmarks, making the structure features less informative for CoCN. However, CoCN still constantly achieves superior performance compared to other models, which proves the efficacy of our model for node classification. \texttt{CoCN expanded} gains better performance than \texttt{CoCN vanilla} on average. The effectiveness of \texttt{CoCN expanded} can be further verified on the filtered benchmarks in Tab.~\ref{tab:filter}, where it is less sensitive to node duplicates compared to \texttt{CoCN vanilla}.

\begin{table*}
    \begin{minipage}{0.51\linewidth}
        \tabcaption{\textbf{Node Classification Results on Large-scale Heterophilic Benchmarks (measured by accuracy for amazon-ratings, ROC AUC for the rest: \%).} OOM denotes hyperparameter settings run out of memory.}
        \label{tab:node-hetero}
        \centering
        \resizebox{\textwidth}{!}{
\begin{small}
\begin{tabular}{lccccc} 
\toprule
               & \begin{tabular}[c]{@{}c@{}}amazon\\-ratings\end{tabular} & \begin{tabular}[c]{@{}c@{}}mine\\sweeper\end{tabular} & tolokers   & questions  & genius      \\ 
\midrule
Nodes            & 24,492                                                   & 10,000      & 11,758     & 48,921     & 421,961     \\
Egdes            & 93,050                                                   & 39,402      & 519,000    & 153,540    & 984,979     \\
Features         & 300                                                      & 7           & 10         & 301        & 12          \\ 
\midrule
GCN~\cite{kipf_SemiSupervisedClassificationGraph_2017}       & 48.70$_{\color{gray}\pm0.63}$              & 89.75$_{\color{gray}\pm0.52}$           & 83.64$_{\color{gray}\pm0.67}$          & 76.09$_{\color{gray}\pm1.27}$          & 87.42$_{\color{gray}\pm0.37}$  \\
GAT~\cite{velickovic_GraphAttentionNetworks_2018}            & 49.09$_{\color{gray}\pm0.63}$              & \textbf{92.01$_{\color{gray}\pm0.68}$}  & \textbf{83.70$_{\color{gray}\pm0.47}$} & 77.43$_{\color{gray}\pm1.20}$          & 55.80$_{\color{gray}\pm0.87}$  \\
H$_2$GCN~\cite{zhu_HomophilyGraphNeural_2020}                & 36.47$_{\color{gray}\pm0.23}$              & 89.71$_{\color{gray}\pm0.31}$           & 73.35$_{\color{gray}\pm1.01}$          & 63.59$_{\color{gray}\pm1.46}$          & OOM         \\
GPRGCN~\cite{chien_AdaptiveUniversalGeneralized_2022}        & 44.88$_{\color{gray}\pm0.34}$              & 86.24$_{\color{gray}\pm0.61}$           & 72.94$_{\color{gray}\pm0.97}$          & 55.48$_{\color{gray}\pm0.91}$          & 90.05$_{\color{gray}\pm0.31}$  \\
SGC~\cite{wu_SimplifyingGraphConvolutional_2019}             & 50.66$_{\color{gray}\pm0.48}$              & 70.88$_{\color{gray}\pm0.90}$           & 80.70$_{\color{gray}\pm0.97}$          & 75.91$_{\color{gray}\pm0.96}$          & 82.36$_{\color{gray}\pm0.37}$  \\
FAGCN~\cite{bo_LowfrequencyInformationGraph_2021}            & 44.12$_{\color{gray}\pm0.30}$              & 88.17$_{\color{gray}\pm0.73}$           & 77.75$_{\color{gray}\pm1.05}$          & 77.24$_{\color{gray}\pm1.26}$          & -           \\
GloGNN~\cite{li_FindingGlobalHomophily_2022}                 & 36.89$_{\color{gray}\pm0.14}$              & 51.08$_{\color{gray}\pm1.23}$           & 73.39$_{\color{gray}\pm1.17}$          & 65.74$_{\color{gray}\pm1.19}$          & 90.66$_{\color{gray}\pm0.11}$  \\ 
\midrule         
\textbf{Sp. CoCN (Ours)}                                     & \textbf{50.74$_{\color{gray}\pm0.15}$}     & 90.72$_{\color{gray}\pm0.17}$           & 82.03$_{\color{gray}\pm0.16}$          & \textbf{77.55$_{\color{gray}\pm0.31}$} & 89.25$_{\color{gray}\pm0.43}$       \\
\textbf{Seg. CoCN (Ours)}                                    & 43.06$_{\color{gray}\pm0.21}$              & 72.98$_{\color{gray}\pm1.69}$           & 77.55$_{\color{gray}\pm2.40}$          & 73.14$_{\color{gray}\pm0.85}$          & \textbf{90.80$_{\color{gray}\pm0.89}$}  \\
\bottomrule
\end{tabular}
\end{small}
}
    \end{minipage}%
    \hfill
    \begin{minipage}{0.48\linewidth}
        \tabcaption{\textbf{Node Classification Results on Large-scale Homophilic Benchmarks (measured by accuracy: \%).}}
        \label{tab:node-homo}
        \vspace{1em}
        \centering
        \resizebox{\textwidth}{!}{
\begin{small}
\begin{tabular}{lcccc} 
\hline
               & \begin{tabular}[c]{@{}c@{}}Coauthor\\CS\end{tabular} & \begin{tabular}[c]{@{}c@{}}Coauthor\\Physics\end{tabular} & \begin{tabular}[c]{@{}c@{}}Amazon\\Photo\end{tabular} & \begin{tabular}[c]{@{}c@{}}Amazon\\Computers\end{tabular}  \\ 
\hline
Nodes            & 18,333                                                & 34,493                                                     & 7,650                                                  & 238,162                                                     \\
Egdes            & 81,894                                                & 247,962                                                    & 119,043                                                & 245,778                                                     \\
Features         & 6,805                                                 & 8,415                                                      & 745                                                   & 767                                                        \\ 
\hline
GCN~\cite{kipf_SemiSupervisedClassificationGraph_2017}        & 92.92$_{\color{gray}\pm0.12}$                  & 96.18$_{\color{gray}\pm0.07}$                  & 92.70$_{\color{gray}\pm0.20}$                   & 89.65$_{\color{gray}\pm0.52}$                                                 \\
GAT~\cite{velickovic_GraphAttentionNetworks_2018}             & 93.61$_{\color{gray}\pm0.14}$                  & 96.17$_{\color{gray}\pm0.08}$                  & 93.87$_{\color{gray}\pm0.11}$                   & 90.78$_{\color{gray}\pm0.17}$                                                 \\
GraphSAGE~\cite{hamilton_InductiveRepresentationLearning_2017}& 89.93$_{\color{gray}\pm0.79}$                  & 92.47$_{\color{gray}\pm0.94}$                  & 79.37$_{\color{gray}\pm1.38}$                   & 88.04$_{\color{gray}\pm0.85}$                                                 \\
GPRGCN~\cite{chien_AdaptiveUniversalGeneralized_2022}         & 95.13$_{\color{gray}\pm0.09}$                  & 96.85$_{\color{gray}\pm0.08}$                  & 94.49$_{\color{gray}\pm0.14}$                   & 89.32$_{\color{gray}\pm0.29}$                                                 \\
GRAND~\cite{chamberlain_GRANDGraphNeural_2021}                & 89.20$_{\color{gray}\pm0.62}$                  & 90.72$_{\color{gray}\pm0.87}$                  & 89.05$_{\color{gray}\pm0.73}$                   & 81.09$_{\color{gray}\pm0.70}$                                                 \\
GCNII~\cite{chen_SimpleDeepGraph_2020}                        & 91.16$_{\color{gray}\pm0.28}$                  & 92.97$_{\color{gray}\pm0.60}$                  & 89.98$_{\color{gray}\pm0.86}$                   & 82.72$_{\color{gray}\pm0.98}$                                                 \\
A-DGN~\cite{gravina_AntiSymmetricDGNStable_2023}              & 91.71$_{\color{gray}\pm0.43}$                  & 93.27$_{\color{gray}\pm0.62}$                  & 90.52$_{\color{gray}\pm0.40}$                   & 82.35$_{\color{gray}\pm0.89}$                                                 \\ 
\hline
\textbf{Sp. CoCN (Ours)}                                      & \textbf{95.25$_{\color{gray}\pm0.19}$}         & \textbf{96.91$_{\color{gray}\pm0.12}$}         & \textbf{95.03$_{\color{gray}\pm0.03}$}          & \textbf{91.27$_{\color{gray}\pm0.16}$}                                          \\
\textbf{Seg.~CoCN (Ours)}                                     & 92.19$_{\color{gray}\pm0.89}$                  & 96.27$_{\color{gray}\pm0.47}$                  & 92.00$_{\color{gray}\pm0.18}$                   & 87.51$_{\color{gray}\pm0.68}$                                                   \\
\hline
\end{tabular}
\end{small}
}
    \end{minipage}
\end{table*}

\textbf{Performance on large-scale benchmarks.}~
We perform experiments on large-scale benchmarks to evaluate the effectiveness of \texttt{CoCN turbo}, \emph{i.e.}, sparse CoCN and segment CoCN. The comparison results between CoCN and baseline models are presented in Tab.~\ref{tab:node-hetero} and \ref{tab:node-homo}. We can see that both sparse and segment CoCNs successfully scale up to large-scale graphs. Sparse CoCN achieves superior or comparable performance against baseline models while being consistent between homophilic and heterophilic benchmarks. For segment CoCN, it gains competitive performance to the baselines. Specifically, segment CoCN outperforms other compared models on the largest benchmark genius. This indicates that increasing the training data size benefits the optimization of segment CoCN.

\begin{table}[t]
\caption{\textbf{Link Prediction Results on PCQM-Contact (measured by mean reciprocal rank and hits at topK: \%).}}\label{tab:link}
\begin{center}
\begin{threeparttable}
\begin{small}
\setlength{\tabcolsep}{3pt}
\begin{tabular}{lccc} 
\toprule
                & MRR        & Hits@1     & Hits@3      \\ 
\midrule
Cache-GNN+RWSE~\cite{ma_AugmentingRecurrentGraph_2023}                    & 34.88$_{\color{gray}\pm0.08}$           & 14.63$_{\color{gray}\pm0.11}$           & 41.02$_{\color{gray}\pm0.08}$  \\
Drew-GCN~\cite{gutteridge_DRewDynamicallyRewired_2023}                    & 34.44$_{\color{gray}\pm0.17}$           & -                                       & -                              \\
Graph Diffuser~\cite{glickman_DiffusingGraphAttention_2023}               & 33.88$_{\color{gray}\pm0.11}$           & 13.69$_{\color{gray}\pm0.12}$           & 40.53$_{\color{gray}\pm0.11}$  \\
GCN~\cite{kipf_SemiSupervisedClassificationGraph_2017}                    & 32.34$_{\color{gray}\pm0.06}$           & 13.21$_{\color{gray}\pm0.07}$           & 37.91$_{\color{gray}\pm0.04}$  \\
GatedGCN~\cite{bresson_ResidualGatedGraph_2018}                           & 32.18$_{\color{gray}\pm0.11}$           & 12.79$_{\color{gray}\pm0.18}$           & 37.83$_{\color{gray}\pm0.04}$  \\
GINE~\cite{hu*_StrategiesPretrainingGraph_2019, xu*_HowPowerfulAre_2019}  & 31.80$_{\color{gray}\pm0.27}$           & 13.37$_{\color{gray}\pm0.13}$           & 36.42$_{\color{gray}\pm0.43}$  \\
GCNII~\cite{chen_SimpleDeepGraph_2020}                                    & 31.61$_{\color{gray}\pm0.04}$           & 13.25$_{\color{gray}\pm0.09}$           & 36.07$_{\color{gray}\pm0.03}$  \\ 
\midrule
\textbf{CoCN exp. (Ours)}                                                      & \textbf{37.13$_{\color{gray}\pm0.37}$}  & \textbf{17.52$_{\color{gray}\pm0.28}$}  & \textbf{41.25$_{\color{gray}\pm0.52}$}       \\
\bottomrule
\end{tabular}
\end{small}
\begin{tablenotes}
    \item CoCN exp. denotes CoCN expanded.
\end{tablenotes}
\end{threeparttable}
\end{center}
\vskip -0.1in
\end{table}

\textbf{Performance on the link prediction benchmark.}~
To explore the potential of CoCN as a general graph representation learning backbone, we evaluate CoCN on the link prediction task on PCQM-Contact dataset. By employing compressed convolution on the input graphs, CoCN compares pair-wise similarities between nodes and classifies their relationships. As presented in Tab.~\ref{tab:link}, CoCN surpasses the compared models on all three metrics, which verifies the effectiveness of CoCN on general graph tasks.

\textbf{Filtered Benchmarks.}~
Platonov et al.,~\cite{platonov_CriticalLookEvaluation_2023} discover substantial overlap between duplicate nodes in the training, validation, and testing subsets of Chameleon and Squirrel. This data leakage across splits leads GNNs to overfit the test data during training, making the performance of GNNs on Chameleon and Squirrel less reliable. To further validate the performance of CoCN on Chameleon and Squirrel, we utilize filtered versions of these benchmarks that exclude duplicate nodes across splits. The performance comparison between the original and filtered benchmarks is presented in Tab.~\ref{tab:filter}. The results of baselines are from~\cite{platonov_CriticalLookEvaluation_2023}. The ranks denote the accuracy ranking out of 15 models on the original and filtered benchmarks, respectively. Part of the results are presented in Tab.~\ref{tab:filter}. For full results, please refer to the Appendix.
CoCN consistently outperforms baseline models on both original and filtered benchmarks. Compared to other models, CoCN is less sensitive to node duplicates and has stronger generalizing ability. These results provide further validation of the reliability of CoCN on Chameleon and Squirrel, demonstrating its potential for node classification.

\subsection{Model Analysis}\label{ssec:ablation}

\subsubsection{Permutation Analysis}
\begin{figure}[htb]
\centering
\subfigure{
  \rotatebox{90}{\scriptsize{~~~~Adj. Matrix}}
  \begin{minipage}[t]{0.22\linewidth}
  \centering
  \includegraphics[width=\textwidth,height=0.98\textwidth]{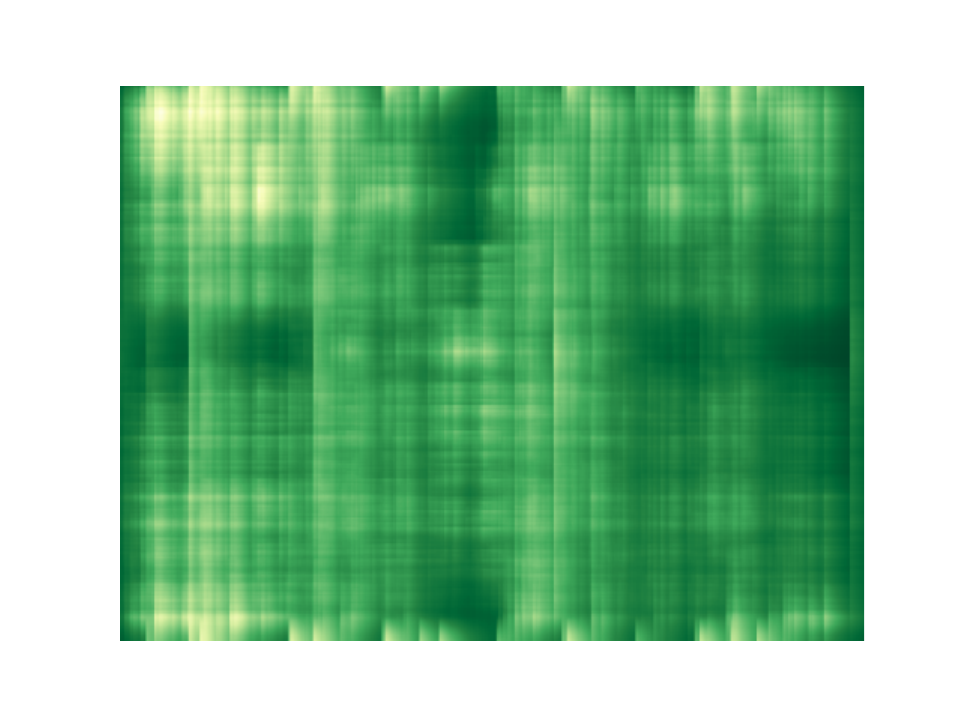}
  \end{minipage}}
\subfigure{
  \begin{minipage}[t]{0.22\linewidth}
  \centering
  \includegraphics[width=\textwidth,height=0.98\textwidth]{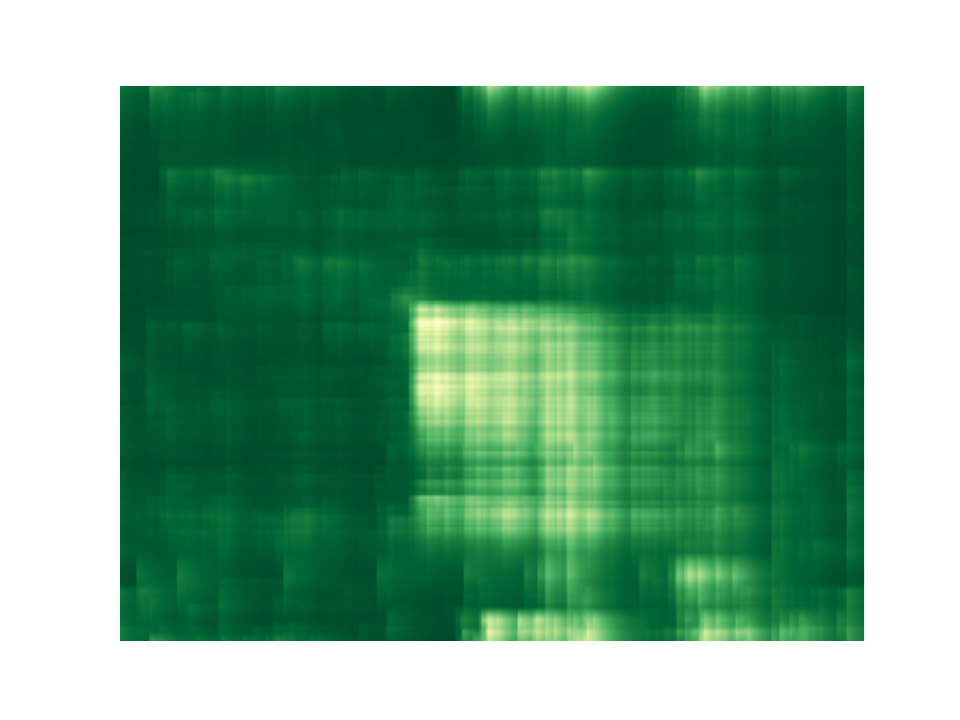}
  \end{minipage}}
\subfigure{
  \begin{minipage}[t]{0.22\linewidth}
  \centering
  \includegraphics[width=\textwidth,height=0.98\textwidth]{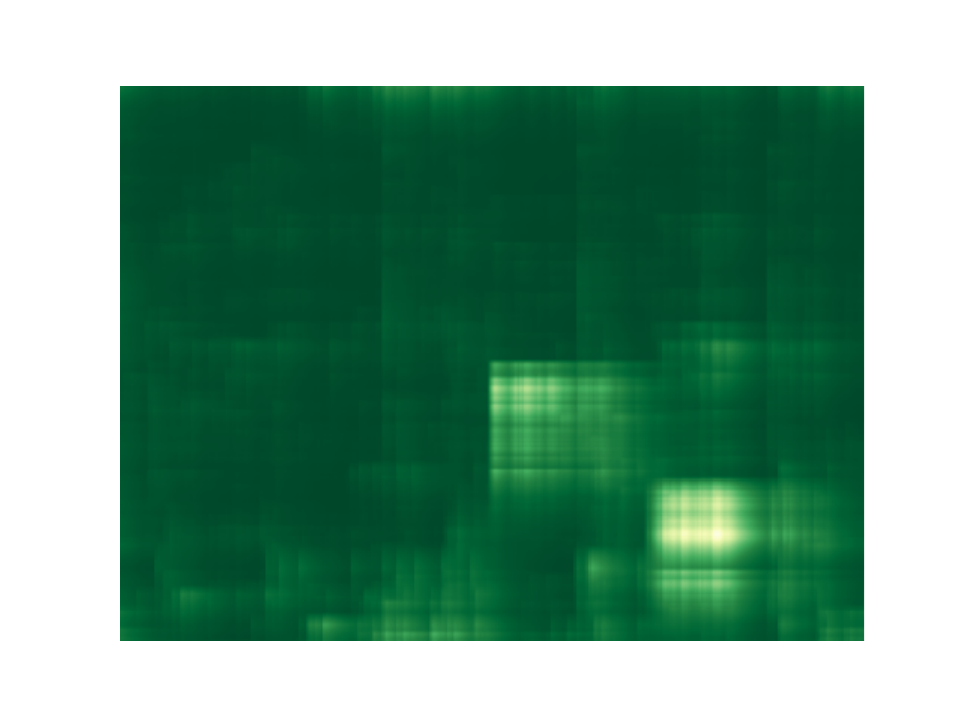}
  \end{minipage}}
\subfigure{
  \begin{minipage}[t]{0.22\linewidth}
  \centering
  \includegraphics[width=\textwidth,height=0.98\textwidth]{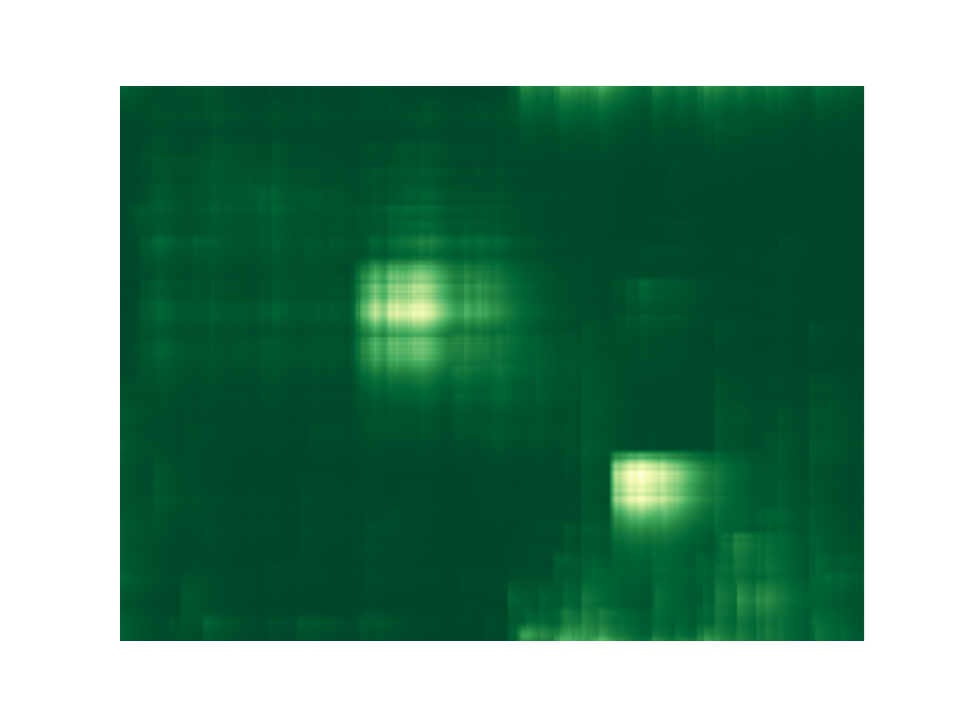}
  \end{minipage}}
\setcounter{subfigure}{0}

\vspace{-0.2in}
\subfigure[0]{
  \rotatebox{90}{\scriptsize{~Node Feat. Sim.}}
  \begin{minipage}[t]{0.22\linewidth}
  \centering
  \includegraphics[width=\textwidth]{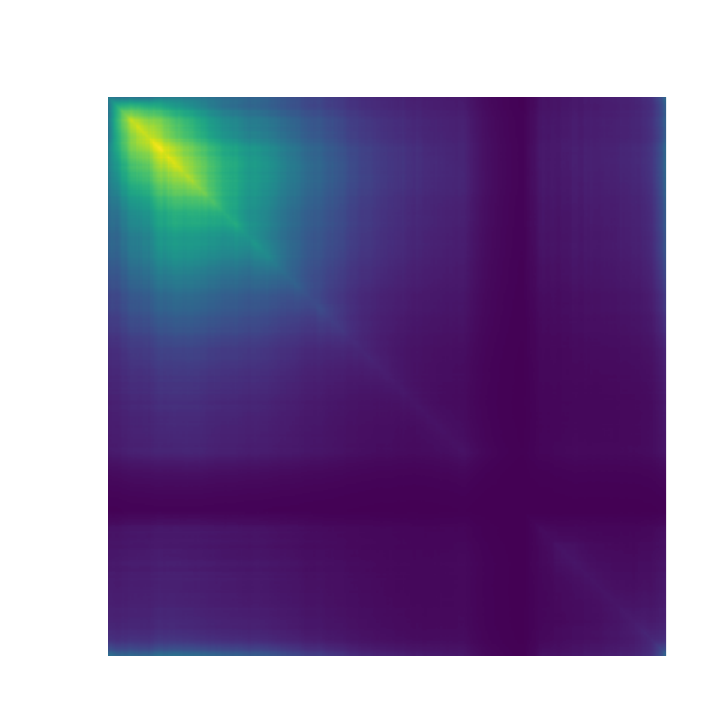}
  \end{minipage}}
\subfigure[1]{
  \begin{minipage}[t]{0.22\linewidth}
  \centering
  \includegraphics[width=\textwidth]{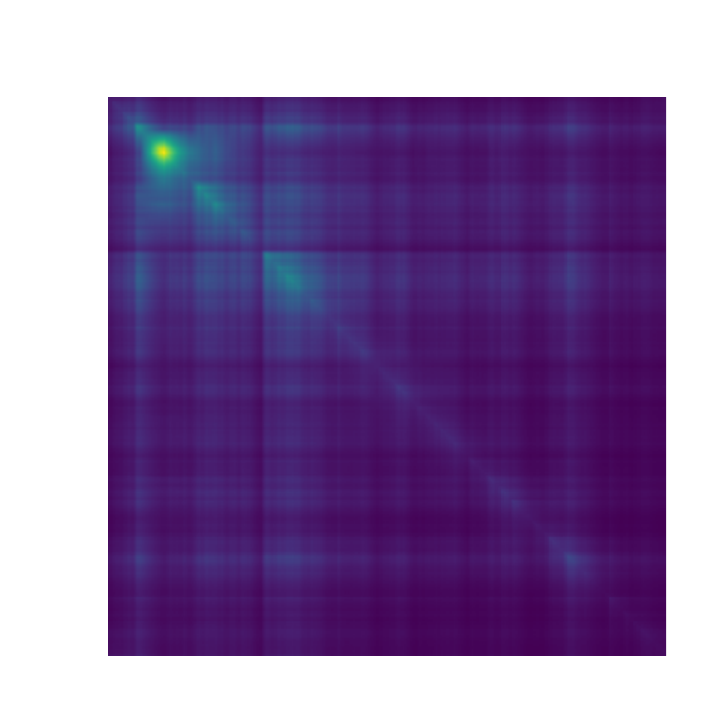}
  \end{minipage}}
\subfigure[2]{
  \begin{minipage}[t]{0.22\linewidth}
  \centering
  \includegraphics[width=\textwidth]{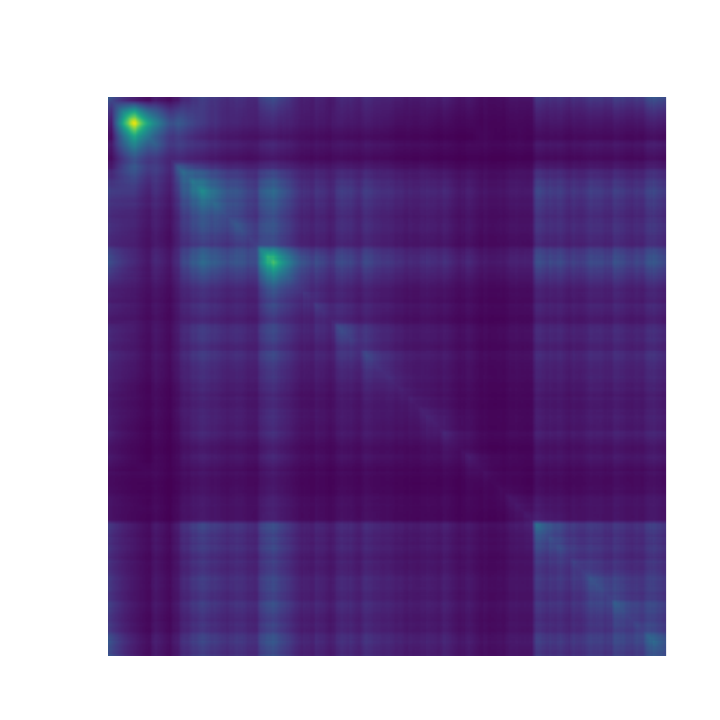}
  \end{minipage}}
\subfigure[4]{
  \begin{minipage}[t]{0.22\linewidth}
  \centering
  \includegraphics[width=\textwidth]{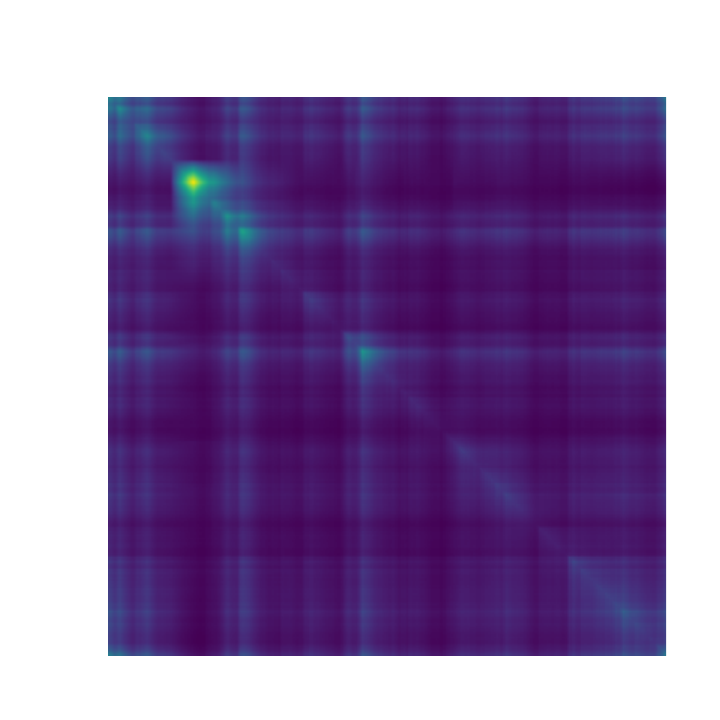}
  \end{minipage}}
\caption{\textbf{Permutation Results with Explicit Node Features.} The results are compared under different values of the smoothness parameter $t$. The first row displays the permuted adjacency matrix and the brighter pixels indicate that the nodes are connected in the adjacency matrix. The second row displays the permuted node feature similarity and the brightness of the pixels indicates the similarity.}\label{fig:permutation-ex-result}
\end{figure}

\textbf{Position Regression Comparison.}~
We conduct module analysis on the position regression module in Section~\ref{ssec:permutation-module}, where nodes with similar features or short paths in between will get closer position prediction. For graphs with explicit node features, both the node feature matrix and the adjacency matrix are employed. To adjust the contribution of graph structures in determining node position, a smoothness parameter $t$ is adopted. We compare the permutation matrix $\hat{\mathbf{P}}_t$ with different values of $t$ on Chameleon. The relaxation factor $\tau$ is set to $0.01$. The permuted adjacency matrix $\hat{\mathbf{P}}_t \mathbf{A} \hat{\mathbf{P}}^\top_t$ and the permuted node feature similarity $\hat{\mathbf{P}}_t \mathbf{X} \mathbf{X}^\top \hat{\mathbf{P}}^\top_t$ are visualized in the first row and second row of Fig.~\ref{fig:permutation-ex-result}. For the adjacency matrix, the brighter pixels indicate that nodes are connected in the input graphs. For the feature similarity results, the brightness of the pixels indicates the similarity values. When $t$ equals $0$, the absolute positions only depend on node features. Nodes with similar features are assigned together. As $t$ increases, the adjacency of graph nodes has more influence on the permutation. Edges in the adjacency matrix, \emph{i.e.}, the brighter pixels in Fig.~\ref{fig:permutation-ex-result}, are permuted close to the main diagonal. This indicates that densely connected nodes are assigned closer positions.

\begin{table}[t]
\caption{\textbf{Node Classification Results (measured by accuracy: \%) on Original / Filtered Chameleon and Squirrel.}}\label{tab:filter}
\begin{center}
\begin{threeparttable}
\begin{small}
\setlength{\tabcolsep}{3pt}
\begin{tabular}{lcccccc} 
\toprule
            & \multicolumn{3}{c}{Chameleon} & \multicolumn{3}{c}{Squirrel}  \\ 
\cline{2-7}
            & Ori.  & Filt. & Ranks         & Ori.  & Filt. & Ranks         \\ 
\midrule
ResNet+SGC~\cite{wu_SimplifyingGraphConvolutional_2019}  & 49.93 & 41.01 & 11/4          & 34.36 & 38.36 & 11/7          \\
ResNet+adj~\cite{platonov_CriticalLookEvaluation_2023}  & 71.07 & 38.67 & 4/11          & 65.46 & 38.37 & 4/6           \\
GCN~\cite{kipf_SemiSupervisedClassificationGraph_2017}         & 50.18 & 40.89 & 9.5/5         & 39.06 & 39.47 & 8/4           \\
FSGNN~\cite{maurya_SimplifyingApproachNode_2022}       & 77.85 & 40.61 & 3/6           & 68.93 & 35.92 & 3/10           \\
FAGCN~\cite{bo_LowfrequencyInformationGraph_2021}       & 64.23 & 41.90 & 7/3           & 47.63 & 41.08 & 6/2           \\
\midrule
\textbf{CoCN va.~(Ours)} & 79.17 & 41.95 & 2/2           & 72.95 & 39.69 & 2/3           \\
\textbf{CoCN exp.~(Ours)} & \textbf{79.35} & \textbf{43.15} & \textbf{1}/\textbf{1}           & \textbf{72.98} & \textbf{41.57} & \textbf{1}/\textbf{1}           \\
\bottomrule
\end{tabular}
\end{small}
\begin{tablenotes}
    \item CoCN va. denotes CoCN vanilla and CoCN exp. denotes CoCN expanded.
\end{tablenotes}
\end{threeparttable}
\end{center}
\end{table}

\begin{figure}[htb]
\centering
\subfigure{
  \rotatebox{90}{\scriptsize{~~~~~~~CHAMELEON}}
  \begin{minipage}[t]{0.28\linewidth}
  \centering
  \includegraphics[width=\textwidth,height=0.98\textwidth]{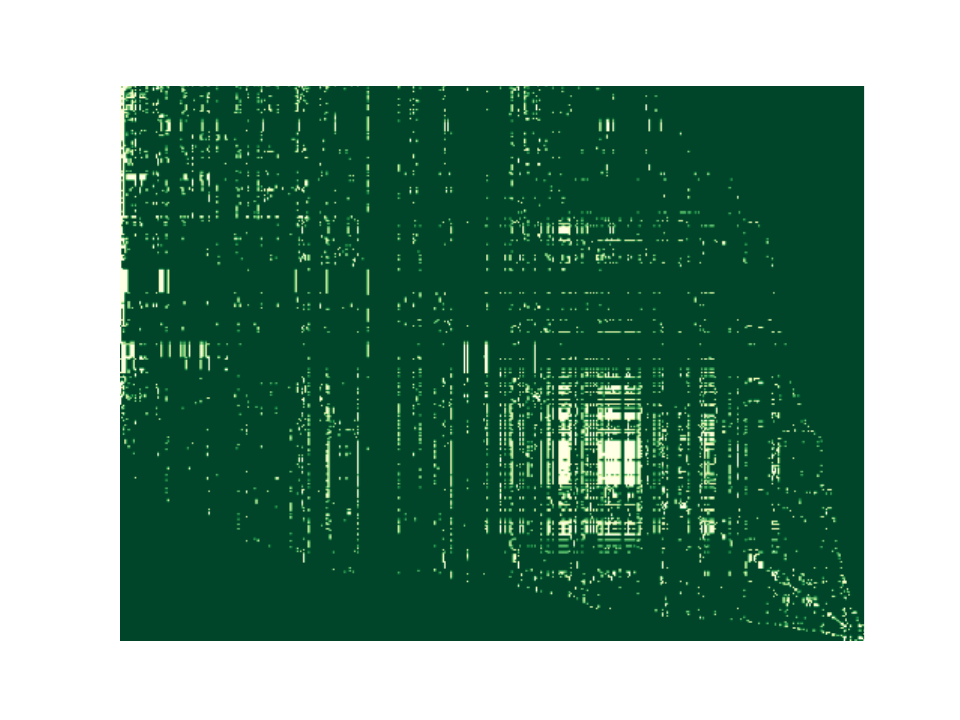}
  \end{minipage}}
\hspace{-0.1in}
\subfigure{
  \begin{minipage}[t]{0.28\linewidth}
  \centering
  \includegraphics[width=\textwidth,height=0.98\textwidth]{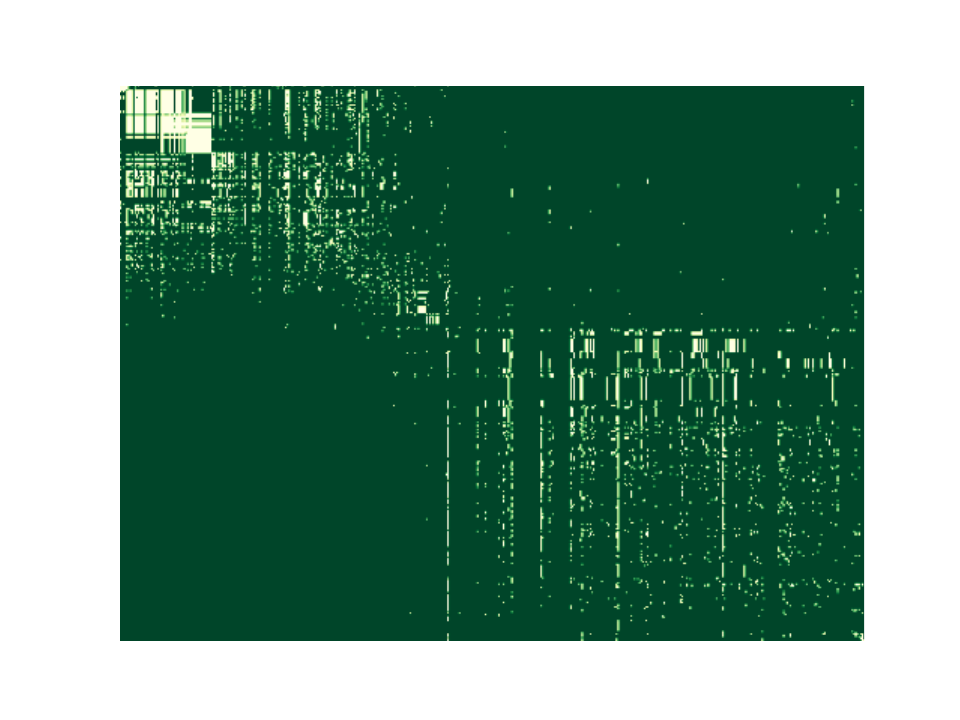}
  \end{minipage}}
\hspace{-0.1in}
\subfigure{
  \begin{minipage}[t]{0.28\linewidth}
  \centering
  \includegraphics[width=\textwidth,height=0.98\textwidth]{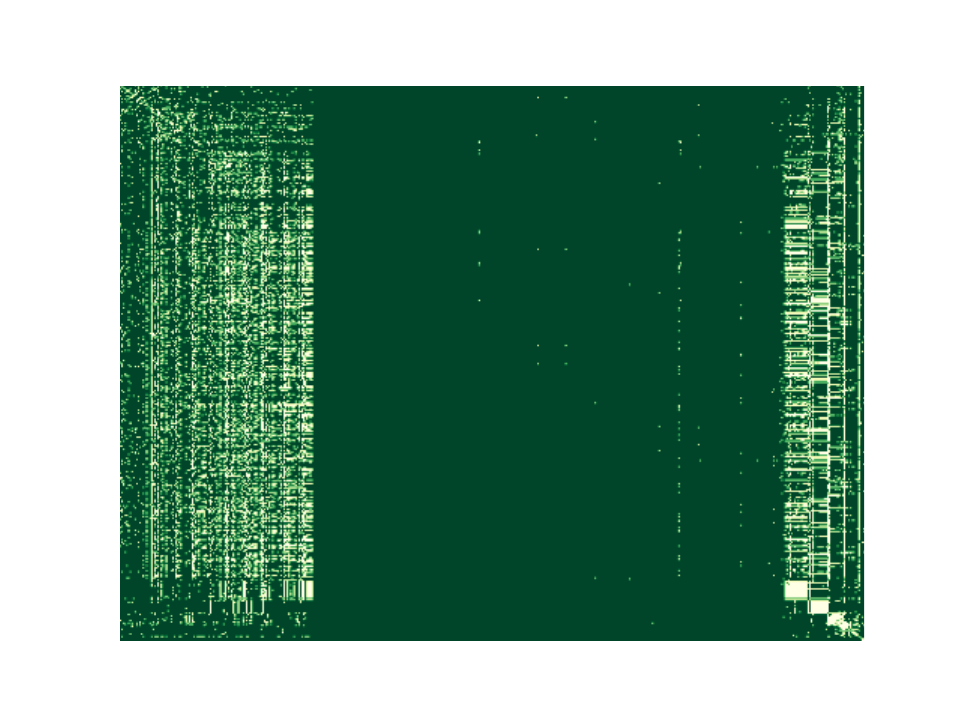}
  \end{minipage}}
\setcounter{subfigure}{0}

\vspace{-0.3in}
\subfigure{
  \rotatebox{90}{\scriptsize{~~~~~~~~~SQUIRREL}}
  \begin{minipage}[t]{0.28\linewidth}
  \centering
  \includegraphics[width=\textwidth,height=0.98\textwidth]{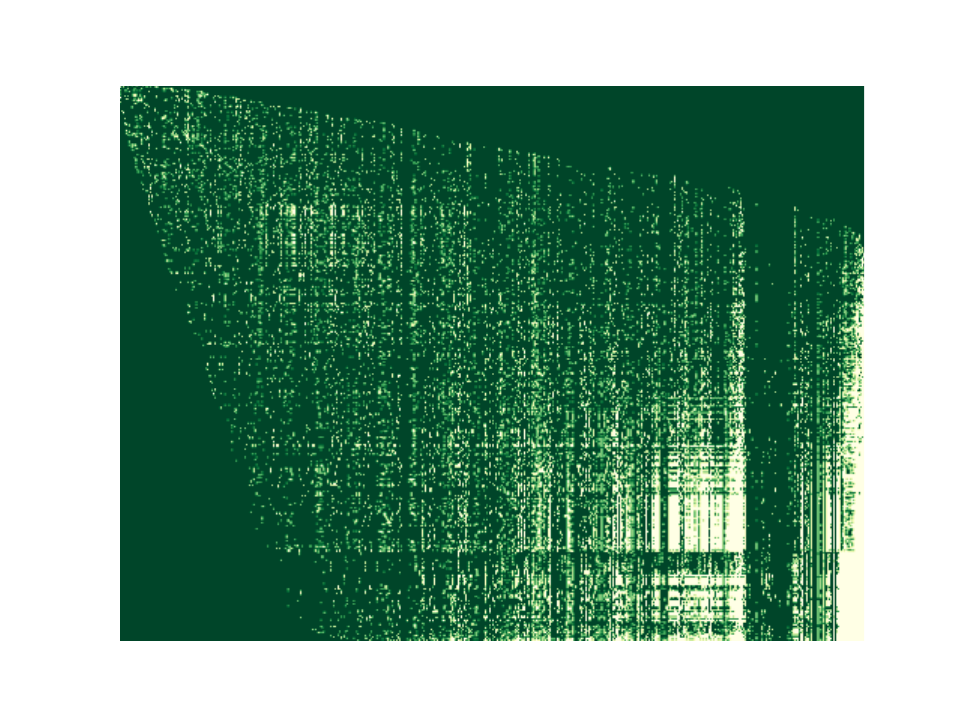}
  \end{minipage}}
\hspace{-0.1in}
\subfigure{
  \begin{minipage}[t]{0.28\linewidth}
  \centering
  \includegraphics[width=\textwidth,height=0.98\textwidth]{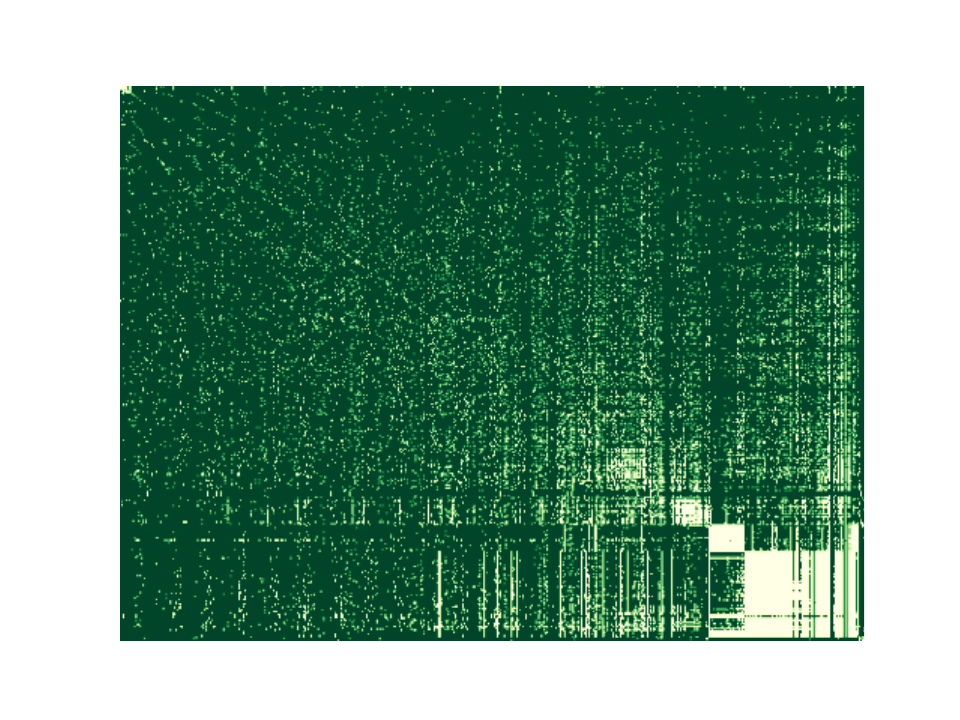}
  \end{minipage}}
\hspace{-0.1in}
\subfigure{
  \begin{minipage}[t]{0.28\linewidth}
  \centering
  \includegraphics[width=\textwidth,height=0.98\textwidth]{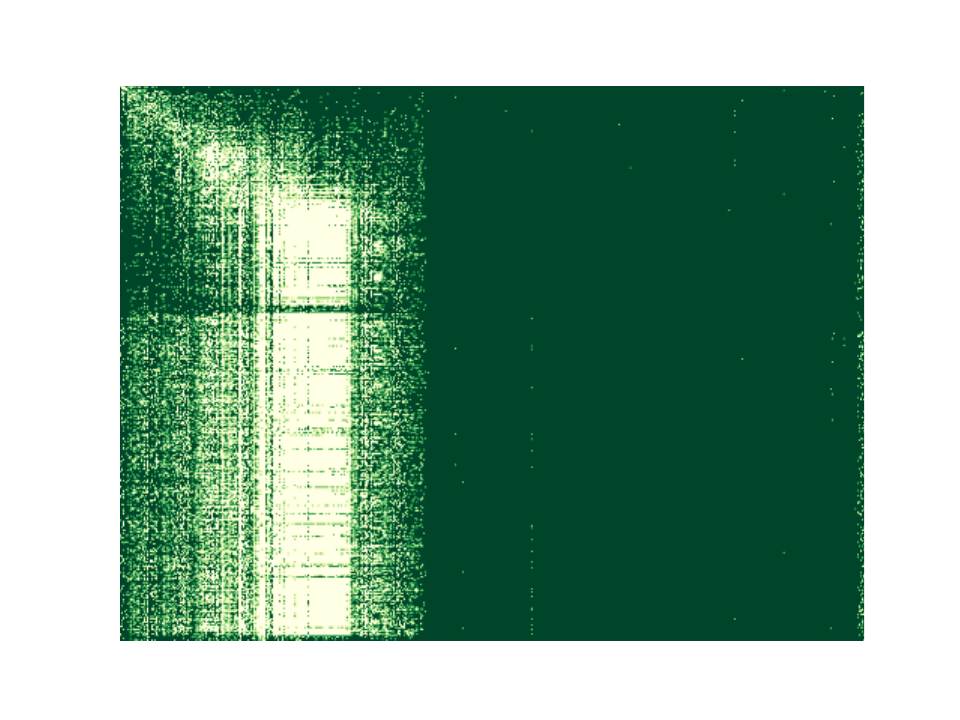}
  \end{minipage}}

\vspace{-0.3in}
\setcounter{subfigure}{0}
\subfigure[Ours]{
  \rotatebox{90}{\scriptsize{~~~~~~~~~~~~~Actor}}
  \begin{minipage}[t]{0.28\linewidth}
  \centering
  \includegraphics[width=\textwidth,height=0.98\textwidth]{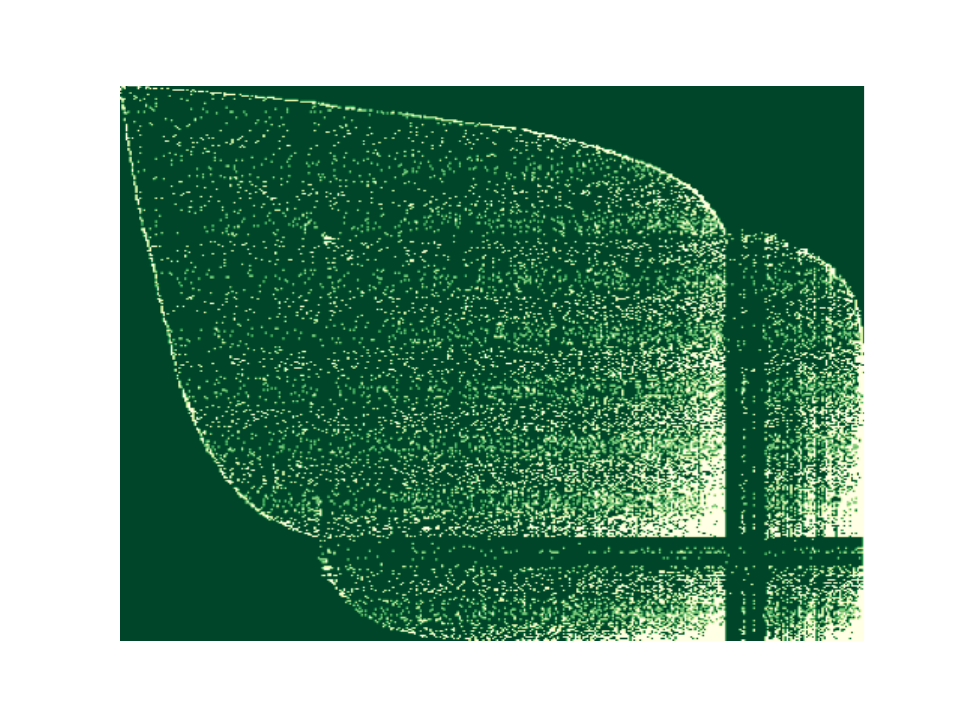}
  \end{minipage}}
\hspace{-0.1in}
\subfigure[LapPE]{
  \begin{minipage}[t]{0.28\linewidth}
  \centering
  \includegraphics[width=\textwidth,height=0.98\textwidth]{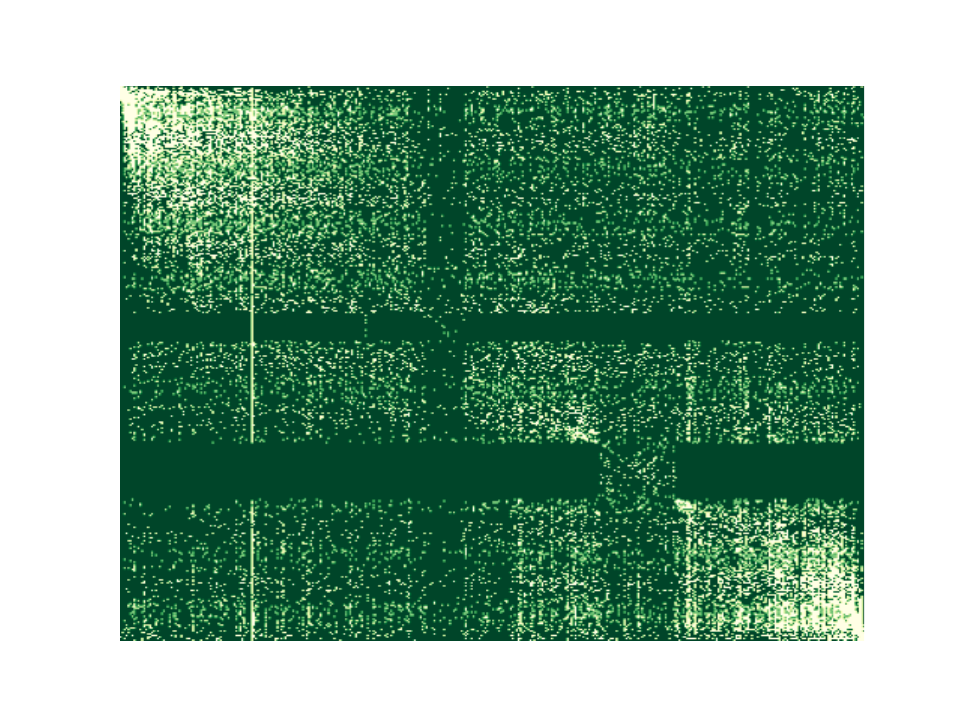}
  \end{minipage}}
\hspace{-0.1in}
\subfigure[RWPE]{
  \begin{minipage}[t]{0.28\linewidth}
  \centering
  \includegraphics[width=\textwidth,height=0.98\textwidth]{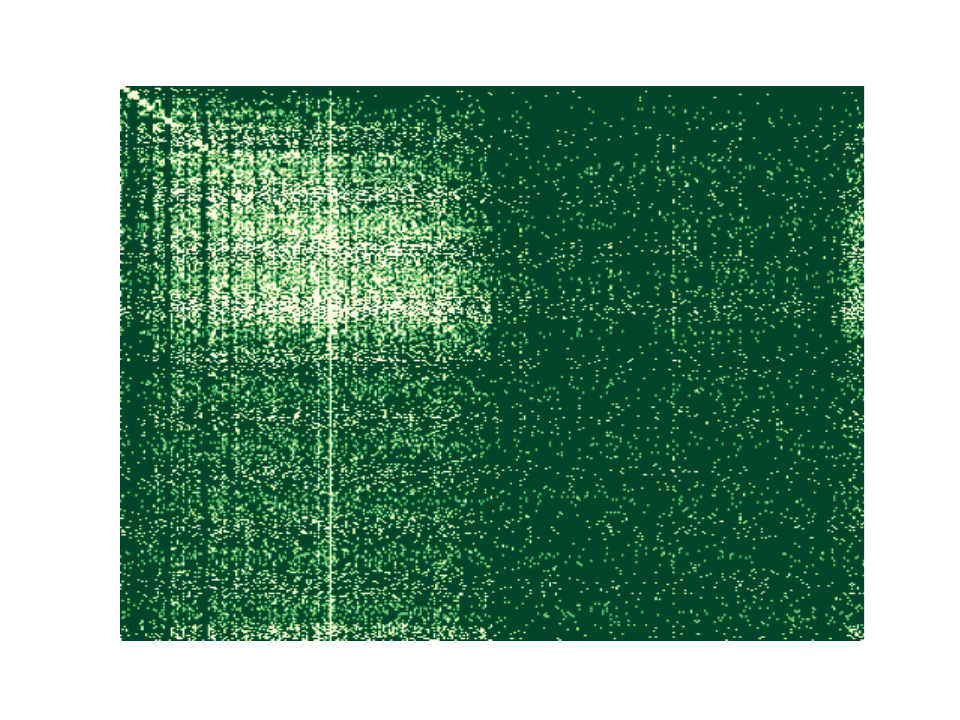}
  \end{minipage}}
\caption{\textbf{Permutation Results with Implicit Node Features.} The results are compared against the Laplacian position encoding method~(LapPE) and the random walk position encoding method~(RWPE). The brighter pixels indicate that the nodes are connected in the input graphs.}\label{fig:permutation-im-result}
\end{figure}
\begin{figure}[htb]
\centering
\vskip 0.05in
\subfigure[Original]{
      \begin{minipage}[t]{0.18\linewidth}
      \centering
      \includegraphics[width=\textwidth]{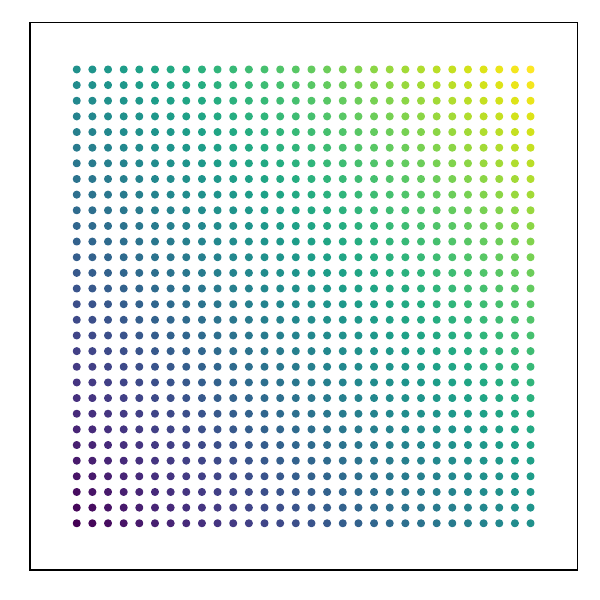}\\
      \includegraphics[width=\textwidth]{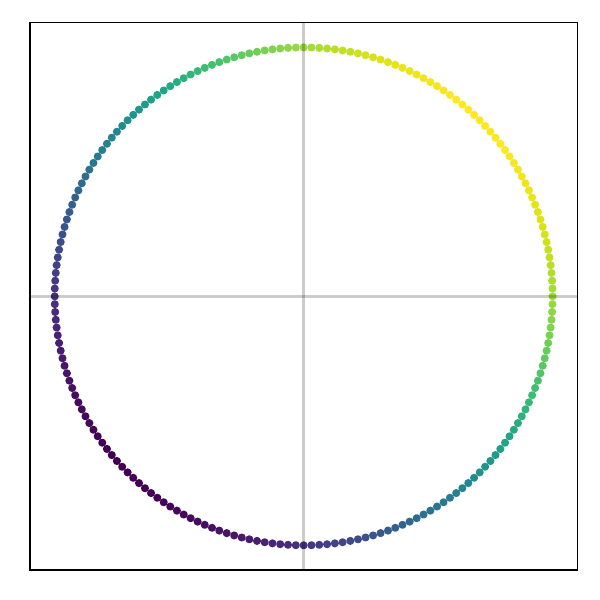}
      \end{minipage}}
 \subfigure[0.01]{
      \begin{minipage}[t]{0.18\linewidth}
      \centering
      \includegraphics[width=\textwidth]{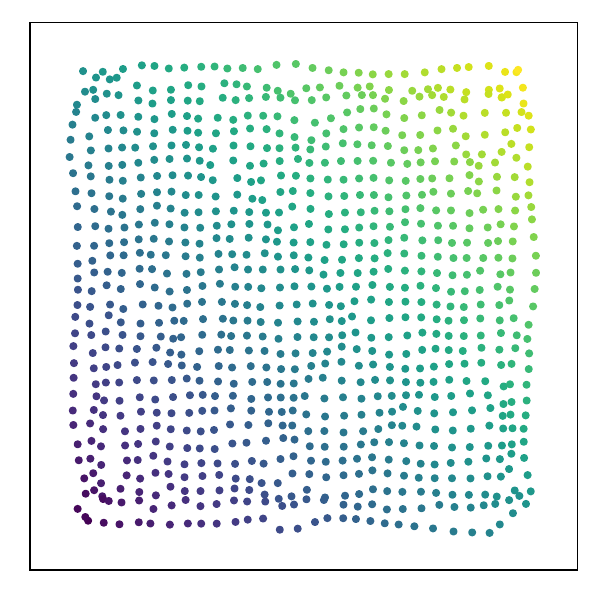}\\
      \includegraphics[width=\textwidth]{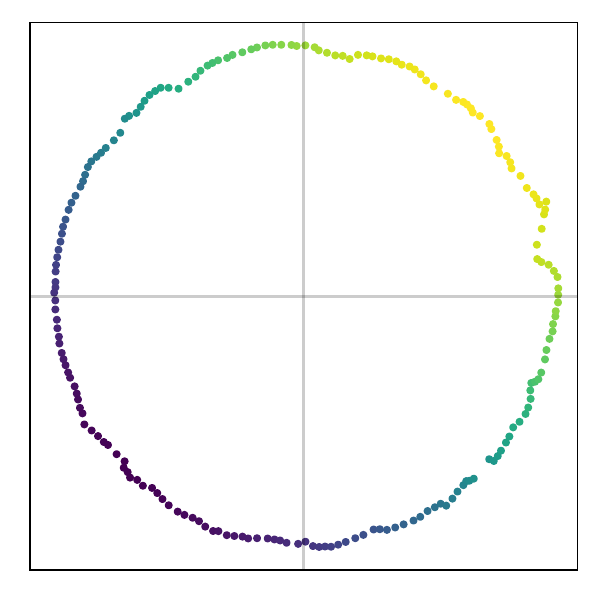}
      \end{minipage}}
 \subfigure[0.1]{
      \begin{minipage}[t]{0.18\linewidth}
      \centering
      \includegraphics[width=\textwidth]{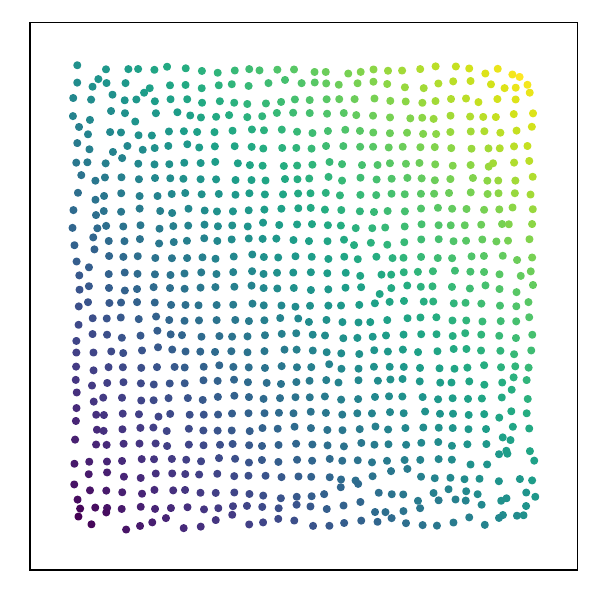}\\
      \includegraphics[width=\textwidth]{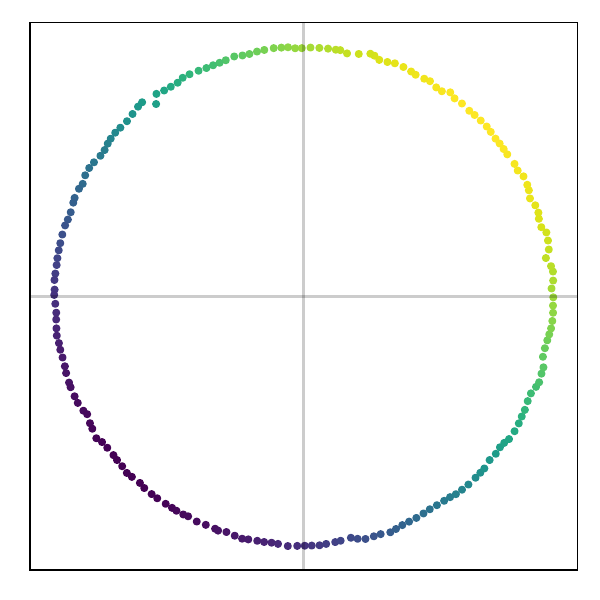}
      \end{minipage}}
\subfigure[1.0]{
      \begin{minipage}[t]{0.18\linewidth}
      \centering
      \includegraphics[width=\textwidth]{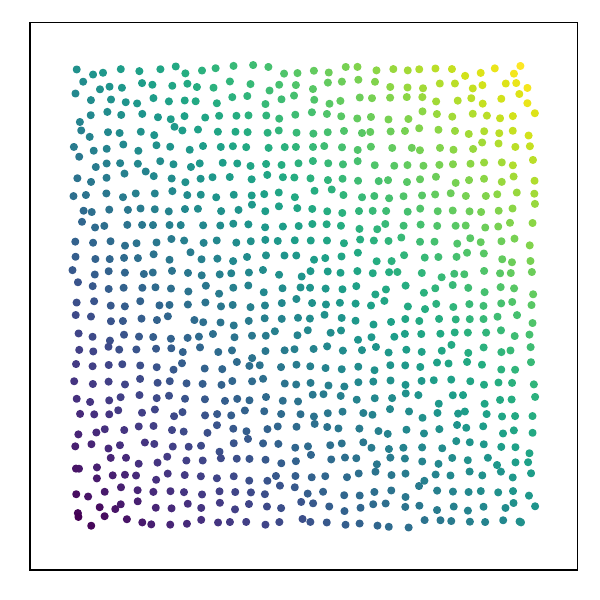}\\
      \includegraphics[width=\textwidth]{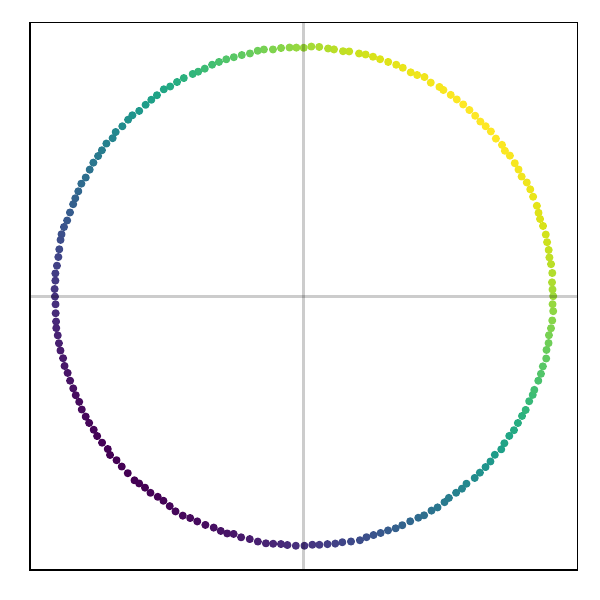}
      \end{minipage}}
 \subfigure[10.0]{
      \begin{minipage}[t]{0.18\linewidth}
      \centering
      \includegraphics[width=\textwidth]{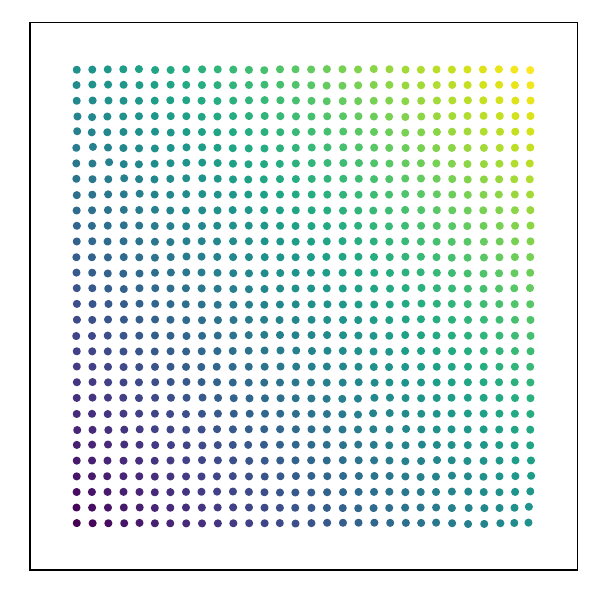}\\
      \includegraphics[width=\textwidth]{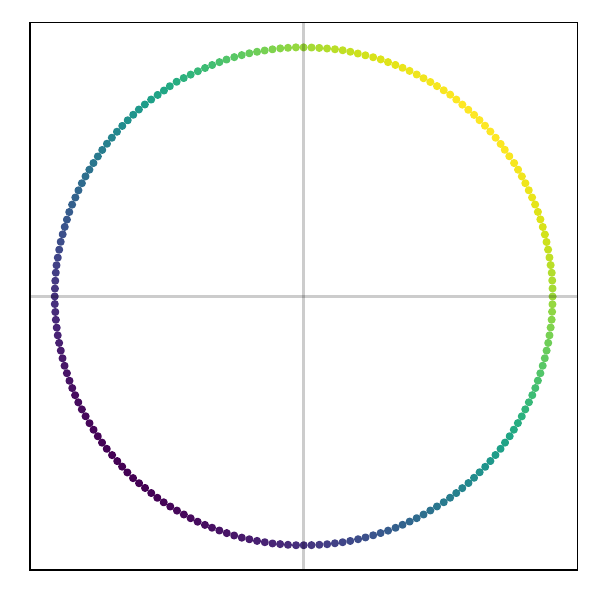}
      \end{minipage}}
 \vskip -0.05in
\caption{\textbf{Permutation Information Loss.} The results under different relaxation factors $\tau$ are compared.}
\label{fig:permutation-loss}
\end{figure}

For input graphs with implicit node features, the position regression solely relies on the distance between nodes. As a result, nodes with short paths in between should have closer positions. To evaluate our proposed position regression method, we compare it against the Laplacian position encoding method~(LapPE)~\cite{dwivedi_BenchmarkingGraphNeural_2020} and the random walk position encoding method~(RWPE)~\cite{dwivedi_GraphNeuralNetworks_2021} on Chameleon, Squirrel, and ACTOR. The relaxation factor $\tau$ is set to $10$. The corresponding permutation results on the adjacency matrix of each method are presented in Fig.~\ref{fig:permutation-im-result}, where brighter pixels indicate the connection of nodes. We can see that our method constantly permutes edges close to the main diagonal. The brighter pixels are distributed with a clear boundary to the top-right and bottom-left corners. In contrast, LapPE and RWPE fail on some or all benchmarks. It should be noted that although LapPE can permute part of the edges close to the main diagonal, the rest of the edges are scattered in the figure without a clear boundary. This demonstrates the effectiveness of our proposed method in learning structure-aware positions without explicit node features.

\begin{table*}
    \begin{minipage}{0.38\linewidth}
        \tabcaption{\textbf{Ablation Studies on the Relaxation Factor ($\tau$).}}
        \label{tab:ablation-tau}
        \centering
        \resizebox{\textwidth}{!}{
\begin{small}
\begin{tabular}{lcccc} 
\hline
          & 0.01  & 0.1   & 1              & 10              \\ 
\hline
CHAMELEON & 67.32 & 77.63 & \textbf{79.17} & 75.88           \\
PROTEINS  & 75.69 & 76.97 & 79.15          & \textbf{79.48}  \\
IMDB-B    & 76.60 & 77.50 & \textbf{78.00} & 77.00           \\
IMDB-M    & 55.8  & 55.93 & 56.13          & \textbf{56.73}  \\
\hline
\end{tabular}
\end{small}
}
    \end{minipage}%
    \hfill
    \begin{minipage}{0.3\linewidth}
        \centering
        \resizebox{\textwidth}{!}{
        \includegraphics[width=\textwidth]{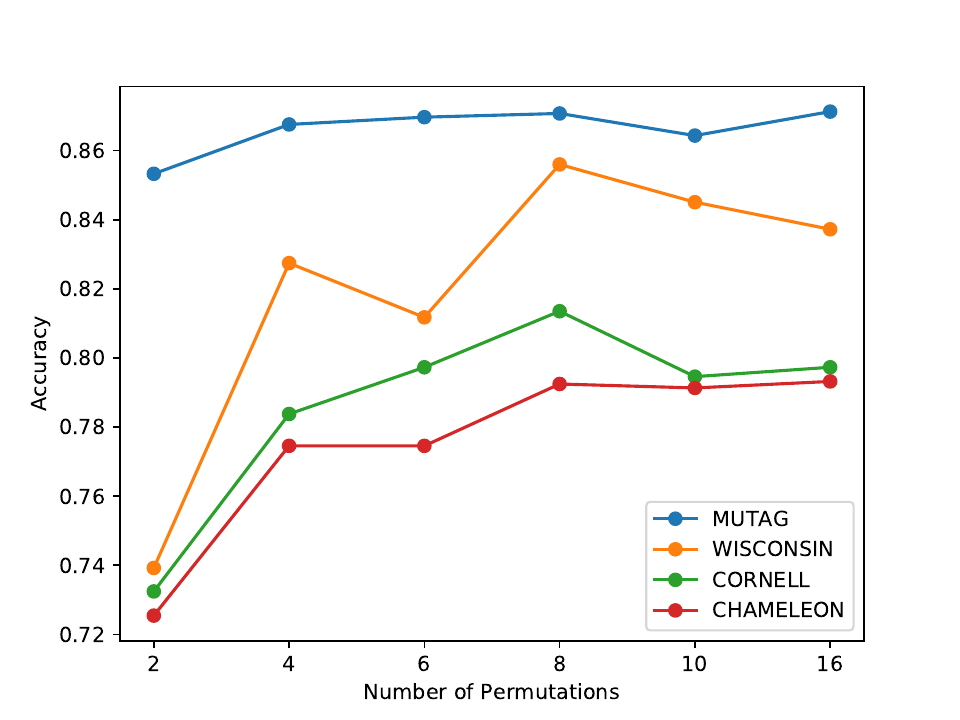}}
        \figcaption{\textbf{Ablation Studies on the Number of Permutations.}}
        \label{fig:permutation-head}
    \end{minipage}%
    \hfill
    \begin{minipage}{0.3\linewidth}
        \centering
        \resizebox{\textwidth}{!}{
        \includegraphics[width=\textwidth]{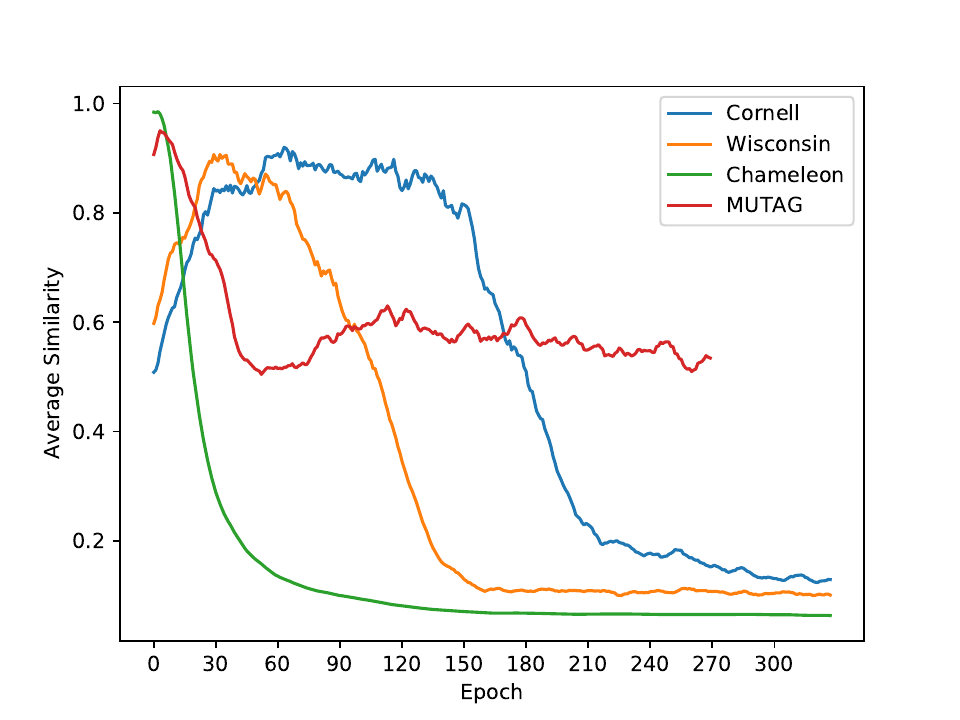}}
        \figcaption{\textbf{Feature Similarity between Different Permutations.}}
        \label{fig:permutation-sim}
    \end{minipage}
\end{table*}
\begin{figure*}[htb]
\centering
\begin{minipage}[t]{0.3\linewidth}
      \centering
      \includegraphics[width=\textwidth]{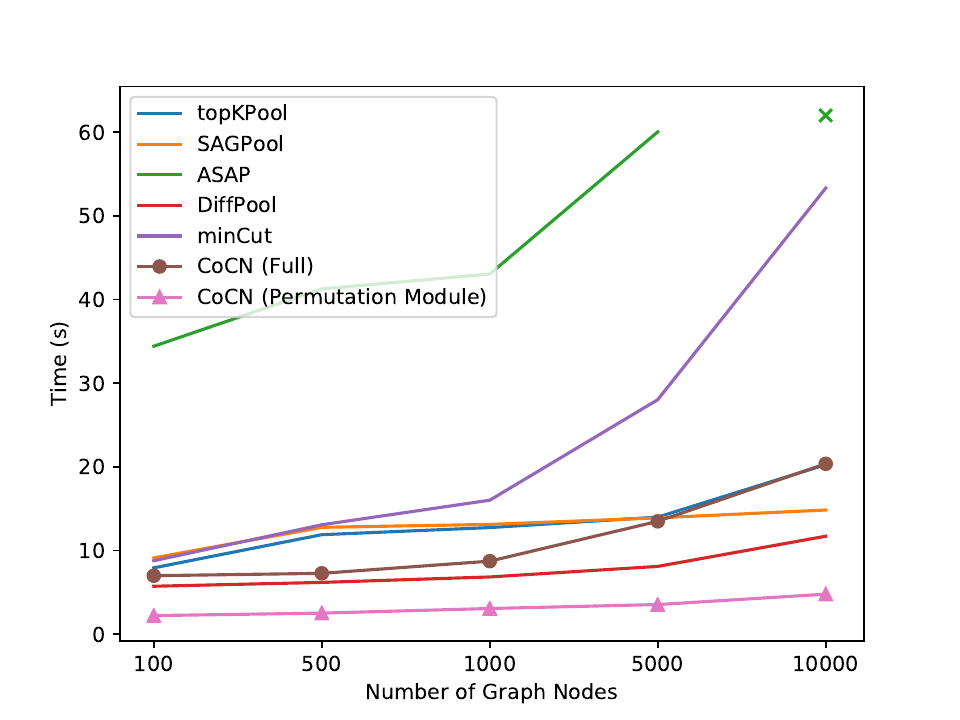}
      \caption{\textbf{Computational Time Comparison.} \textcolor[RGB]{52,163,52}{$\texttt{x}$} denotes hyperparameter settings run out of memory.}
\label{fig:time-vanilla}
\end{minipage}
\hfill
\begin{minipage}[t]{0.3\linewidth}
      \centering
      \includegraphics[width=\textwidth]{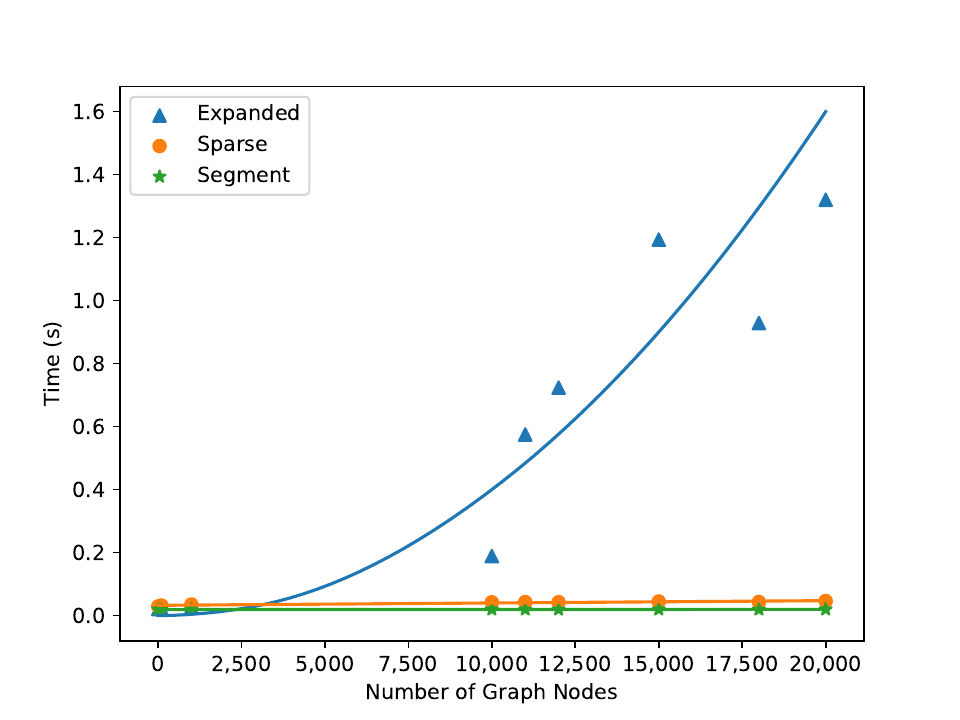}
      \caption{\textbf{Complexity Comparison on Small-scale Graphs.}}
\label{fig:time-small}
\end{minipage}
\hfill
\begin{minipage}[t]{0.3\linewidth}
      \centering
      \includegraphics[width=\textwidth]{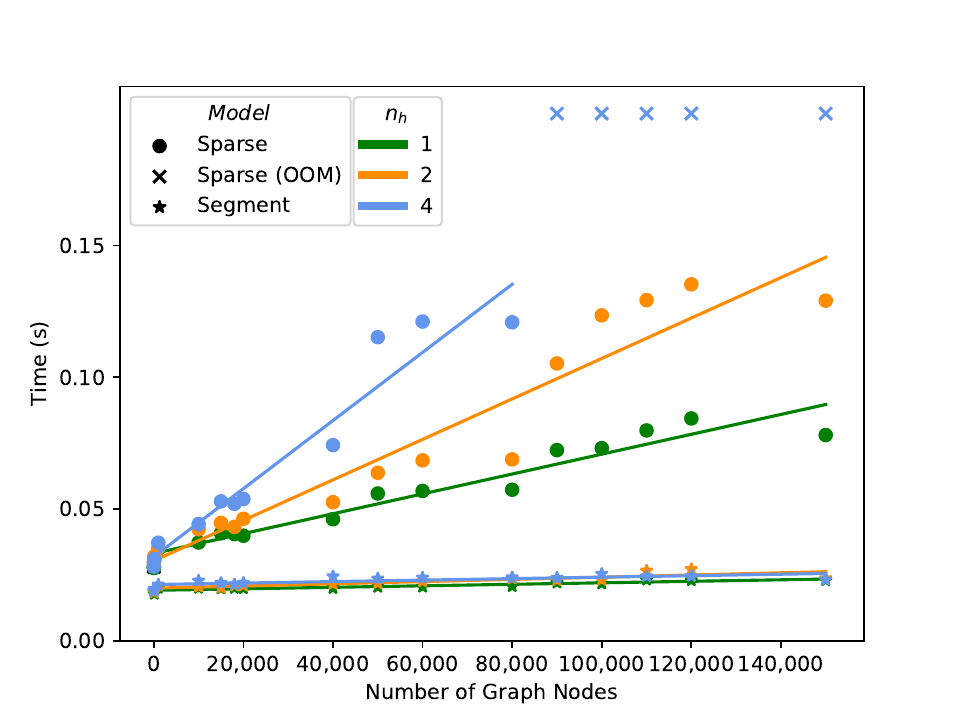}
      \caption{\textbf{Complexity Comparison on Large-scale Graphs.} $n_h$ denotes the number of permutations. \textcolor[RGB]{100,149,237}{$\texttt{x}$} denotes hyperparameter settings running out of memory.}
\label{fig:time-large}
\end{minipage}
\end{figure*}

\textbf{Information Loss.}~
We conduct graph reconstruction experiments following~\cite{bianchi_SpectralClusteringGraph_2020} to study whether the proposed approximate permutations may cause information loss. Given input features, an autoencoder is trained to recover the original features from the permuted features. The learning objective is to minimize the mean squared error~(MSE) between the input features and the recovered features. The input graphs include synthetic graphs~\cite{bianchi_SpectralClusteringGraph_2020} of ring and grid. The input features are the coordinates of nodes in 2D Euclidean space. The original synthetic graphs and the corresponding recovered graphs under different relaxation factors $\tau$ are presented in Fig.~\ref{fig:permutation-loss}. The autoencoder can recover the original node position with small distortion. As $\tau$ increases, the distortion gets reduced. When $\tau$ equals $10$, the autoencoder can reconstruct the graph with almost no visible distortion. This suggests that our permutation generation module can well approximate the permutation matrix with little information loss. 

We further conduct ablation studies on the relaxation factor $\tau$. The results are presented in Tab.~\ref{tab:ablation-tau}. CoCN achieves the best performance with large values of $\tau$, such as 1 and 10. These results empower the usage of the sparse permutation matrix and our further exploration towards more scalable CoCNs, including sparse and segment CoCNs.

\textbf{Multiple Permutations.}~
We conduct ablation studies on the number of permutations and analyze the convolution module output from different permutations. Fig.~\ref{fig:permutation-head} presents the classification accuracy under different numbers of permutations on MUTAG, Chameleon, Cornell, and Wisconsin. As the number of permutations increases, CoCN gets improvement in terms of accuracy for all the benchmarks. We further analyze the relationship between different permutations. Fig.~\ref{fig:permutation-sim} presents the average feature similarity between permutations during the training process. Given two feature vectors $\mathbf{x}$ and $\mathbf{y}$, the feature similarity is $\mathbf{x}^\top\mathbf{y}/d$ where $d$ denotes the number of feature channels. The results are normalized into $[0,1]$ for a unified presentation. From Fig.~\ref{fig:permutation-sim}, the model gradually learns to reduce the feature similarity between different permutations. This indicates that CoCN learns complementary feature representations under different permutations, and validates the necessity of multiple permutations for sufficient feature learning.

\subsubsection{Complexity}
We first conduct the running time analysis on \texttt{CoCN expanded} with other hierarchical graph learning models. For a fair comparison, the down-sampling rate $\sigma$ is set to 20\%, where the number of final output node sets equals $\sigma n$. All models perform down-sampling three times. Since  $\hat{\mathbf{A}}^t$ in Eq.~\ref{eq:approximate-ex-pos-implementation} can be computed before the training process, we set $t=1$ for the permutation generation of CoCN. The running time on 100 Erdős-Rényi graphs with different scales is presented in Fig.~\ref{fig:time-vanilla}. We can see that the computational efficiency of \texttt{CoCN expanded} is comparable against other models and the computation of the permutation only takes a small portion of the whole process.

\begin{table*}
    \begin{minipage}{0.54\linewidth}
        \tabcaption{\textbf{Ablation Study (measured by accuracy and improvements: \%) on Residual Connection and Inception Mechanism.}}
        \label{tab:ablation-module}
        \centering
        \resizebox{\textwidth}{!}{
\begin{threeparttable}
\begin{small}
\begin{tabular}{lccccccc} 
\toprule
               & Plain & \multicolumn{2}{c}{Inception} & \multicolumn{2}{c}{Res. Conn.} & \multicolumn{2}{c}{Full}  \\
               & Acc.  & Acc.  & Imp.                  & Acc.  & Imp.                   & Acc.  & Imp.              \\ 
\midrule
PROTEINS       & 73.72 & 77.16 & +4.66                 & 79.45 & \textbf{+7.77}         & 78.18 & +6.05             \\
COLLAB         & 86.03 & 87.89 & \textbf{+2.16}        & 86.19 & +0.19                  & 87.22 & +1.38             \\
IMDB-B         & 76.95 & 77.58 & +0.82                 & 77.26 & +0.40                  & 77.94 & \textbf{+1.29}    \\
questions      & 75.53 & 73.73 & -2.38                 & 77.90 & \textbf{+3.14}         & 74.05 & -1.96             \\
tolokers       & 73.12 & 77.15 & +5.51                 & 81.50 & +11.46                 & 82.03 & \textbf{+12.19}   \\
CoauthorCS     & 94.46 & 90.34 & -4.36                 & 95.25 & +0.84                  & 95.34 & \textbf{+0.93}    \\
AmazonPhoto    & 93.40 & 90.00 & -3.64                 & 93.68 & +0.30                  & 95.03 & \textbf{+1.75}    \\
\bottomrule
\end{tabular}
\end{small}
\begin{tablenotes}
    \item Imp. denotes the improvements of each CoCN variant to their corresponding plain variants.
\end{tablenotes}
\end{threeparttable}
}
    \end{minipage}%
    \hfill
    \begin{minipage}{0.45\linewidth}
        \tabcaption{\textbf{Ablation Study (measured by accuracy: \%) on Structure Features and Node Features.}}
        \label{tab:ablation-feat}
        \centering
        \resizebox{\textwidth}{!}{
\setlength{\tabcolsep}{10pt}
\begin{small}
\begin{tabular}{lccc} 
\toprule
          & Str. Feat.     & Nod. Feat. & Both            \\ 
\midrule
Chameleon & 67.82          & 77.98      & \textbf{79.17}  \\
Squirrel  & 72.79          & 71.91      & \textbf{72.95}  \\
Actor     & 33.73          & 34.36      & \textbf{35.55}  \\
MUTAG     & \textbf{88.35} & 85.12      & 87.08           \\
NCI1      & 64.08          & 78.65      & \textbf{82.28}  \\
IMDB-B    & 70.47          & 76.17      & \textbf{77.26}  \\
COLLAB    & 78.14          & 84.46      & \textbf{86.15}  \\
\bottomrule
\end{tabular}
\end{small}
}
    \end{minipage}
\end{table*}

To evaluate the computational efficiency of the scalable CoCNs, we also perform complexity comparisons among \texttt{CoCN expanded} and sparse/segment CoCNs. All models are trained with 10 epochs on random synthetic graphs with an average degree of $8$. For segment CoCN, the segmentation size $b$ and batch size $n_b$ are set to $8$ and $1000$, respectively. The comparison results are presented in Fig.~\ref{fig:time-small} and~\ref{fig:time-large}. As \texttt{CoCN expanded} cannot scale to large-scale graphs, the number of graph nodes included in the comparison is restricted to be smaller than $20,000$ and presented in Fig.~\ref{fig:time-small}. We can see that both sparse and segment CoCNs are more efficient than \texttt{CoCN expanded}. 

Delving into the comparison between sparse and segment CoCNs in Fig.~\ref{fig:time-large}, both variants can scale to large-scale graphs. However, sparse CoCN still faces out-of-memory issues under certain hyperparameter configurations. In contrast, segment CoCN achieves superior computational efficiency and maintains scalability across various graph sizes.

\subsubsection{Network Structure Analysis}
\begin{figure}[htb]
\centering
\subfigure{
      \rotatebox{90}{\scriptsize{~~~~~~~~amazon-ratings}}
      \begin{minipage}[t]{0.45\linewidth}
      \centering
      \includegraphics[width=\textwidth]{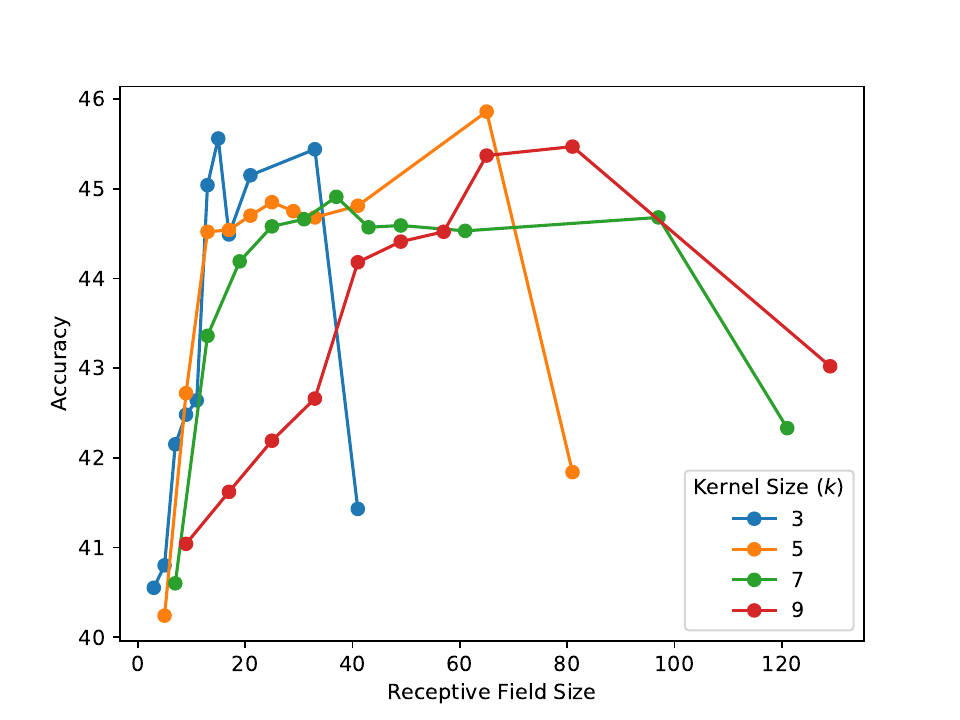}
      \end{minipage}}
 \subfigure{
      \begin{minipage}[t]{0.45\linewidth}
      \centering
      \includegraphics[width=\textwidth]{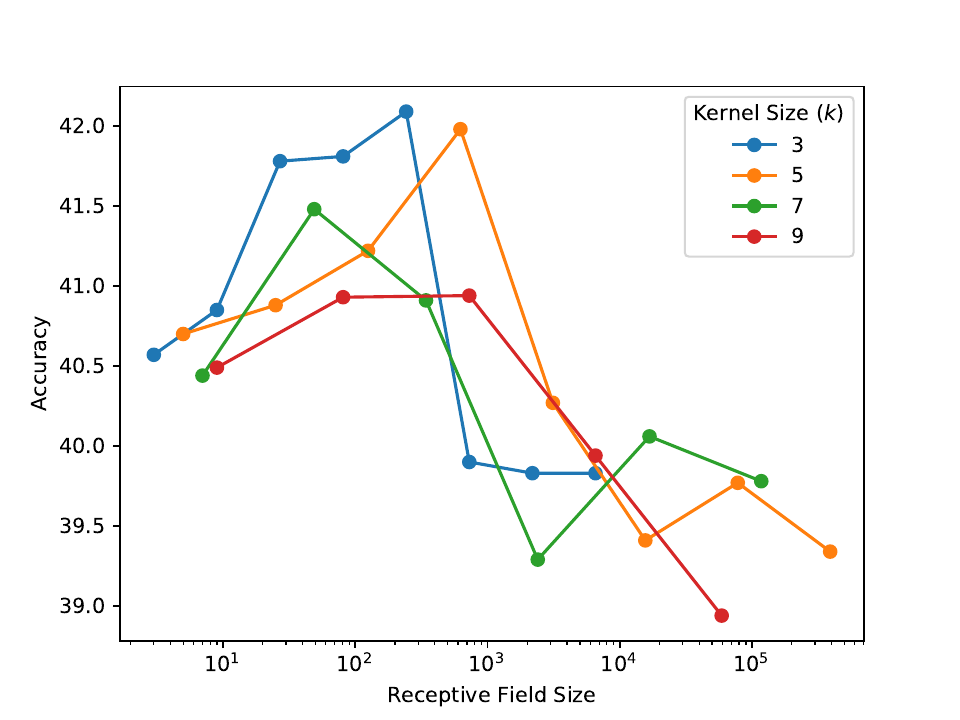}
      \end{minipage}}
\setcounter{subfigure}{0}
\subfigure[$L_1$ ($L_2$=1)]{\label{fig:kernel-layer-l1}
      \rotatebox{90}{\scriptsize{~~~~~~~~CHAMELEON}}
      \begin{minipage}[t]{0.45\linewidth}
      \centering
      \includegraphics[width=\textwidth]{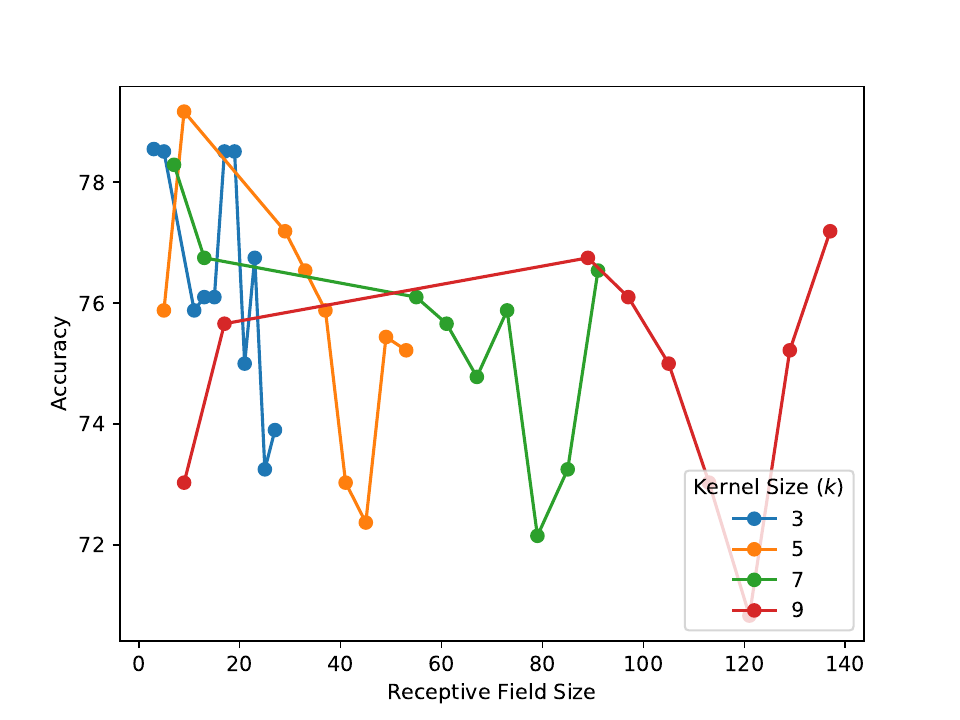}
      \end{minipage}}
\subfigure[$L_2$ ($L_1$=1)]{\label{fig:kernel-layer-l2}
      \begin{minipage}[t]{0.45\linewidth}
      \centering
      \includegraphics[width=\textwidth]{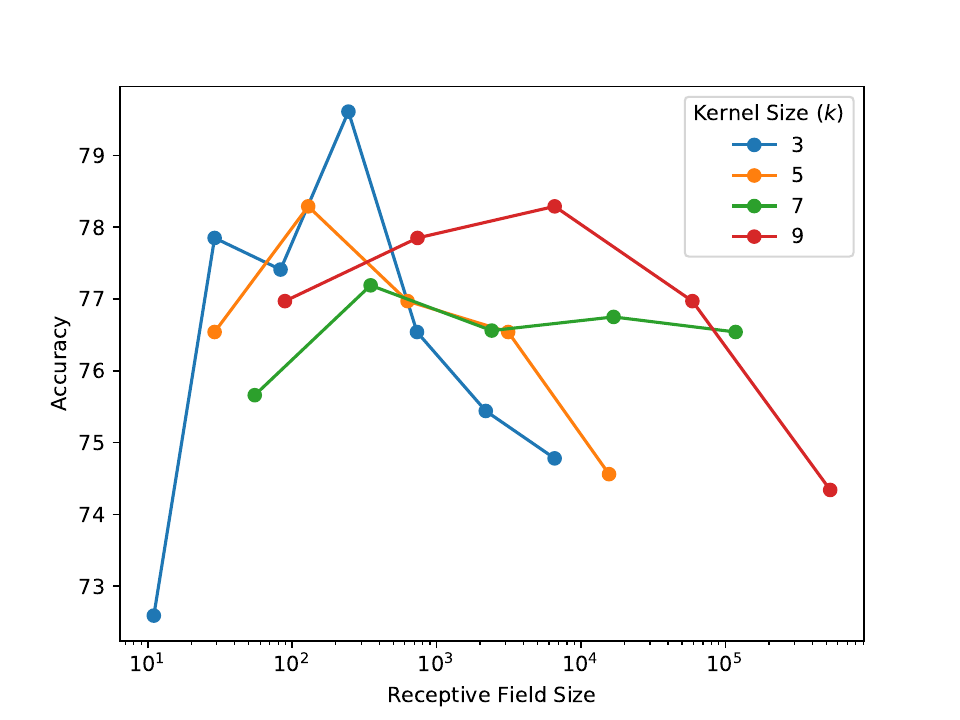}
      \end{minipage}}
\caption{\textbf{Ablation Studies on Kernel Size ($k$) and Receptive Field Size.} $L_1$ and $L_2$ denote the numbers of compressed convolution layers with unit step and compressed pooling layers, respectively}
\label{fig:kernel-layer}
\end{figure}

\textbf{Kernels and Layers.}~
We conduct ablation studies on kernel size $k$, the number of compressed convolution layers with unit step $L_1$, and the number of compressed pooling layers $L_2$ for amazon-ratings and Chameleon (Fig.~\ref{fig:kernel-layer}). For brevity, we refer to compressed convolution layers with unit step as UCo and compressed pooling layers as PCo.

As shown in Fig.~\ref{fig:kernel-layer}, CoCN achieves better performance with increased receptive field size. In Fig.~\ref{fig:kernel-layer-l1}, ablations are conducted on $L_1$ while $L_2$ is fixed to 1. The accuracy curves exhibit a slight increase, followed by a sharp decline, with peak accuracy occurring at small receptive field sizes. In contrast, CoCN attains peak performance with much larger receptive fields in the ablation study on $L_2$ (Fig.~\ref{fig:kernel-layer-l2}). This is due to the inability of UCo to expand the size of the receptive fields efficiently. It requires stacking multiple layers to achieve large receptive fields, which burdens the parameter optimization on CoCN. As a result, CoCN mainly relies on PCos to get larger receptive fields.

However, while failing to efficiently expand the receptive field, UCo can extract more fine-grain features than PCo. On amazon-ratings, stacking UCos achieves better performance than stacking PCos. This indicates that the required receptive fields for individual nodes in amazon-ratings are small, so capturing fine-grain features benefits the corresponding task. Different from amazon-ratings, nodes in CHAMELEON require larger receptive fields. Compared between Fig.~\ref{fig:kernel-layer-l1} and Fig.~\ref{fig:kernel-layer-l2} of CHAMELEON, stacking PCos expands the receptive fields efficiently and achieves better performance than stacking UCos. Moreover, in Fig.~\ref{fig:kernel-layer-l1} of CHAMELEON, the performance of CoCN starts to increase again after the decline. This indicates a trade-off between the difficulty of optimization and the expanding of receptive fields, where stacking multiple UCos can expand the receptive fields but burdens the optimization.

For different kernel sizes $k$, CoCN with smaller $k=3$ and $5$ has better performance. This is because the receptive fields of CoCN with larger kernels ($k=7$ and $9$) increase rapidly where CoCN can hardly capture fine-grain features. 

\textbf{Compositional Convolution Modules.}~
With generalized residual connection and inception mechanism, we conduct ablation studies on the compositional compressed convolution modules. As presented in Tab.~\ref{tab:ablation-module}, the residual connection can consistently improve the performance of CoCN on both graph-level and node-level tasks. For the inception module, it can benefit CoCN on most benchmarks while causing degradation against the plain variants on questions, CoauthorCS, and AmazonPhoto. 
CoCN combined with inception modules and residual connection achieves better or even the best performance on these benchmarks.
This demonstrates that explicitly separating the intrinsic features from the incremental or decremental features via residual connection is critical for CoCN on graph learning.

\subsubsection{Feature Analysis}
We perform ablation studies on node features and structure features. Results on node classification benchmarks (Chameleon, Squirrel, and Actor) and graph classification benchmarks (MUTAG, NCI1, COLLAB, and IMDB-BINARY) are presented in Tab.~\ref{tab:ablation-feat}. We can see that CoCN with both types of features has better performance on most benchmarks. Therefore, both node features and structure features generally contribute to the model prediction. It should be noted that CoCN can still obtain good performance with only structure features. These results demonstrate the structure learning ability of CoCN.

\section{Conclusion}\label{sec:conclusion}
In this paper, we presented a differentiable permutation method for generalizing Euclidean convolution to graphs. The permutation generation can be formulated as a node position assignment problem and transformed the generated node positions into an approximated permutation matrix with cyclic shift. Based on the permuted nodes, Euclidean convolution can be generalized to graph-structured data. Specifically, we designed diagonal convolution for graphs to learn both node features and the corresponding structure features. Building upon the permutation generation and diagonal convolution, we further presented Compressed Convolution Network (CoCN), a hierarchical GNN model for graph representation learning. CoCN can be integrated with successful practices from Euclidean convolution, including residual connection and inception mechanism. Experiment results show that CoCN achieves superior performance on both node-level and graph-level tasks. The effectiveness of CoCN is consistent between heterophilic and homophilic benchmarks and scalable to large-scale benchmarks. Comprehensive empirical analysis demonstrates the potential of CoCN as a general graph learning backbone. We will further extend CoCN to more downstream tasks, such as brain network study and intelligent transportation systems.

\section*{Acknowledgment}
This work was supported in part by the National Key R\&D Program of
China under Grant 2023YFC2508704, in part by National Natural Science
Foundation of China: 62236008, U21B2038 and 61931008, and
in part by the Fundamental Research Funds for the Central Universities.

\appendices

\section{Proofs}\label{app:proofs}

\subsection{Proof of Proposition \ref{prop:permutation}}
\textbf{Proposition \ref{prop:permutation}. (Permutation Convergence) } \textit{$\hat{\mathbf{P}}$ converges to standard permutation matrix as relaxation factor $\tau$ approaching positive infinity:$\underset{\tau\rightarrow +\infty}{lim}\hat{\mathbf{P}}=\mathbf{P}, \mathbf{P}\in\mathcal{P}$.}

\begin{proof} 
Assuming that we have absolute position $\mathbf{r}\in\mathbb{R}^n$ given by Eq.~\ref{eq:absolute-pos} where $\texttt{sgn}(\cdot) \in \{0,1\}$, $\mathbf{r}$ is a integer vector and $\mathbf{r}_i\in[0,n - 1]$. Let $\mathbf{m}_P=(\mathbf{m1}^\top-\mathbf{1r}^\top+n) \pmod{n}$ and $\hat{\mathbf{P}}=\texttt{exp}(-\tau \mathbf{m}_P)$. For every row of $\mathbf{m}_P$, there always exists an entry equal to $0$ where the corresponding entry in $\hat{\mathbf{P}}$ constantly equals $1$. The rest entries are integers belonging to $[1,n-1]$, and the corresponding entries in $\hat{\mathbf{P}}$ converge to $0$ as $\tau$ approaches positive infinity.
\end{proof}

\subsection{Proof of Proposition \ref{prop:invariance}}
\textbf{Proposition \ref{prop:invariance}. (Permutation Invariant Input) } \textit{
Let $f$ be a permutation equivariant function such that for any $\mathbf{P}\in\mathcal{P}$, $f(\mathbf{PX},\mathbf{PAP}^\top)=\mathbf{P}f(\mathbf{X},\mathbf{A})$. Then given $\hat{\mathbf{P}}=\texttt{PERM}(\mathbf{X}, \mathbf{A})$, $\hat{\mathbf{X}}=\hat{\mathbf{P}}\mathbf{X}$, and $\hat{\mathbf{A}}= \hat{\mathbf{P}}\mathbf{A}\hat{\mathbf{P}}^\top$, for any $\mathbf{P}\in\mathcal{P}$ on $\mathbf{X}$ and $\mathbf{A}$, $\hat{\mathbf{X}}$ and $\hat{\mathbf{A}}$ are invariant.}

\begin{proof} 
We first prove that $\texttt{PERM}(\cdot)$ is permutation equivariant. Given $\mathbf{PX}$ and $\mathbf{PAP}^\top$ for any $\mathbf{P}\in\mathcal{P}$, the output of $f$ can be written as $\mathbf{Pr}_A=f(\mathbf{PX},\mathbf{PAP}^\top)$. Substituting $\mathbf{Pr}_A$ into Eq.~\ref{eq:absolute-pos} gives:
\begin{equation}
\begin{aligned}
  &\texttt{sgn}\left(\mathbf{Pr}_A \mathbf{1}^\top-\mathbf{1} \mathbf{r}_A^\top \mathbf{P}^\top\right)\mathbf{1}\\
=&\texttt{sgn}\left(\mathbf{Pr}_A \mathbf{1}^\top \mathbf{P}^\top-\mathbf{P1r}_A^\top \mathbf{P}^\top\right)\mathbf{1}\\
=&\mathbf{P}\texttt{sgn}\left(\mathbf{r}_A \mathbf{1}^\top-\mathbf{1} \mathbf{r}_A^\top\right)\mathbf{P}^\top \mathbf{1}\\
=&\mathbf{P}\texttt{sgn}\left(\mathbf{r}_A \mathbf{1}^\top-\mathbf{1}\mathbf{r}_A^\top\right)\mathbf{1}\\
=&\mathbf{Pr}
\end{aligned}
\end{equation}Therefore, Eq.~\ref{eq:absolute-pos} is permutation equivariant. Further substituting $\mathbf{Pr}$ into Eq.~\ref{eq:permutation} gives:
\begin{equation}
\begin{aligned}
  &\texttt{exp}\left\{-\tau \left[\left(\mathbf{m1}^\top- \mathbf{1r}^\top \mathbf{P}^\top + n\right)\pmod{n}\right]\right\}\\
=&\texttt{exp}\left\{-\tau \left[\left(\mathbf{m1}^\top \mathbf{P}^\top-\mathbf{1r}^\top \mathbf{P}^\top + n\right)\pmod{n}\right]\right\}\\
=&\texttt{exp}\left\{-\tau \left[\left(\mathbf{m1}^\top- \mathbf{1r}^\top + n\right)\pmod{n}\right]\right\} \mathbf{P}^\top\\
=&\hat{\mathbf{P}}\mathbf{P}^\top
\end{aligned}
\end{equation}Hence $\texttt{PERM}(\cdot)$ is permutation equivariant. For any $\mathbf{P}\in\mathcal{P}$, we can use the generated permutation matrix $\hat{\mathbf{P}}\mathbf{P}^\top$ to permute the input features where $\hat{\mathbf{P}}\mathbf{P}^\top(\mathbf{PX})=\hat{\mathbf{P}}\mathbf{X}$ and $\hat{\mathbf{P}}\mathbf{P}^\top(\mathbf{PAP}^\top)\mathbf{P}\hat{\mathbf{P}}^\top=\hat{\mathbf{P}}\mathbf{A}\hat{\mathbf{P}}^\top$. Therefore, given $\hat{\mathbf{P}}=\texttt{PERM}(\mathbf{X}, \mathbf{A})$, $\hat{\mathbf{X}}=\hat{\mathbf{P}}\mathbf{X}$ and $\hat{\mathbf{A}}= \hat{\mathbf{P}}\mathbf{A}\hat{\mathbf{P}}^\top$, $\hat{\mathbf{X}}$ and $\hat{\mathbf{A}}$ are invariant for any $\mathbf{P}\in\mathcal{P}$ on $\mathbf{X}$ and $\mathbf{A}$.
\end{proof}

\section{Derivation}\label{app:derivation}
\subsection{Derivation on the Eq.~\ref{eq:approximate-im-pos-implementation}}\label{app:derivation-pos}
For graphs with implicit node features, we base node position regression solely on the distances between nodes, leading to proximate positions for nodes with short paths in between. This intuition can be further formulated as:
\begin{equation}
\begin{aligned}
    &\mathbf{M}^U\mathbf{11}^{\top}-\mathbf{11}^{\top}{\mathbf{M}^U}^\top = \mathbf{M}^D\odot d(\mathbf{A}),\\
    &\mathbf{M}^D = 2\mathbf{M}^U-\mathbf{11}^\top,\\
    &\mathbf{M}^U = \texttt{sgn}\left(\mathbf{r}_A \mathbf{1}^\top-\mathbf{1}\mathbf{r}_A^\top\right),
\end{aligned}
\end{equation}
where $\mathbf{M}^U$ and $\mathbf{M}^D$ denote the sign matrix, $\odot$ denotes element-wise multiplication. $d(\cdot)$ denotes distance function such that $d(\mathbf{A})\in\mathbb{R}^{n\times n}$. By substituting Eq.~\ref{eq:absolute-pos} into Eq.~\ref{eq:implicit-intuition}, the distance-based constraint on the node positions can be reformulated as:
\begin{equation}
    \mathbf{r1}^\top-\mathbf{1r}^\top=(2\mathbf{r}-n\mathbf{1})\mathbf{1}^\top\odot \frac{1}{n}d(\mathbf{A}).
\end{equation}
Then we can apply vectorization to both side of the equation and obtain
\begin{equation}
\begin{aligned}
&\quad (\mathbf{1}\otimes \mathbf{I} - \mathbf{I}\otimes\mathbf{1})\mathbf{r} \\
&= \frac{1}{n}\texttt{diag}\left(\texttt{vec}(d(\mathbf{A}))\right)
\texttt{vec}(2\mathbf{r1}^\top-n\mathbf{1}\mathbf{1}^\top)\\ 
&= \frac{1}{n}\texttt{diag}\left(\texttt{vec}(d(\mathbf{A}))\right)
\left[(\mathbf{1}\otimes \mathbf{I})2\mathbf{r}-\texttt{vec}(n\mathbf{1}\mathbf{1}^\top)\right],
 \end{aligned}
\end{equation}
where $\otimes$ denotes the Kronecker product, $\texttt{diag}(\cdot)$ denotes the diagonalization to the input vector, and $\texttt{vec}(\cdot)$ denotes vectorization to the input matrix. By combing like terms, We then employ Moore-Penrose inverse and get
\begin{equation}
\begin{aligned}
\quad\mathbf{r}
=&((\mathbf{U}+\mathbf{V})^\top(\mathbf{U}+\mathbf{V}))^{-1}\\
&(\mathbf{U}+\mathbf{V})^\top \texttt{vec}(d(\mathbf{\mathbf{A}}))\\
=&(\mathbf{U}^\top \mathbf{U}+\mathbf{U}^\top \mathbf{V}+\mathbf{V}^\top \mathbf{U}+\mathbf{V}^\top \mathbf{V})^{-1}\\
&(\mathbf{U}+\mathbf{V})^\top \texttt{vec}(d(\mathbf{A})),
 \end{aligned}
\end{equation}
where $\mathbf{U}=\mathbf{I}\otimes\mathbf{1}$, $\mathbf{V}=\texttt{diag}\left(\texttt{vec}(d(\mathbf{A}) - 1)\right)(\mathbf{1}\otimes \mathbf{I})$. We can derive the terms separately as
\begin{equation}
\begin{aligned}
\mathbf{U}^\top \mathbf{U} &=
(\mathbf{I}\otimes\mathbf{1}^\top)(\mathbf{I}\otimes\mathbf{1})=
\mathbf{I}\otimes\mathbf{1}^\top\mathbf{1}=\mathbf{I},\\
\mathbf{U}^\top \mathbf{V} &=\texttt{vec}(d(\mathbf{A}) - 1),\\
\mathbf{V}^\top \mathbf{V} &= \texttt{diag}\left(\left(\frac{2}{n}d(\mathbf{A})-1\right)^{\circ 2}\mathbf{1}\right),\\
(\mathbf{U}+\mathbf{V}&)^\top \texttt{vec}(d(\mathbf{A})) = \frac{2}{n}d(\mathbf{A})^{\circ 2}\mathbf{1},
\end{aligned}
\end{equation}
with which we finally have the approximation position for graphs with implicit node features as
\begin{equation}
\begin{aligned}
    \mathbf{r}_A \approx & \left[ \texttt{diag}\left(\left(\frac{2}{n}d(\mathbf{A})-1\right)^{\circ 2}\mathbf{1}\right) \right.\\    
    &\ + \left.\frac{4}{n}d(\mathbf{A}) - 1\right]^{-1} \frac{2}{n}d(\mathbf{A})^{\circ 2}\mathbf{1},
\end{aligned}
\end{equation}
where $(\cdot)^{\circ 2}$ denotes element-wise square and $d(\cdot)$ is implemented as the scaled shortest path distance.

\subsection{Derivation on the Eq.~\ref{eq:approximate-back}}\label{app:derivation-back}
 In the backpropagation of the absolute position $\mathbf{r}$, the original Eq.~\ref{eq:absolute-pos} can be formulated as
\begin{equation}
\begin{matrix}
    \texttt{(original)} &
    \begin{aligned}
        \frac{\partial\mathbf{r}}{\partial\mathbf{r}_A} &= 
        \frac{\partial}{\partial\mathbf{r}_A}
        \texttt{sigmoid}\left(\mathbf{C}\right)\mathbf{1},\\
        \mathbf{C} &= \texttt{ReLU}\left(\mathbf{r}_A \mathbf{1}^\top-\mathbf{1}\mathbf{r}_A^\top\right),
    \end{aligned}
\end{matrix}
\end{equation}
where $\mathbf{C}$ denotes the dense pairwise comparison result. The sigmoid function with ReLU is adopted to approximate the sign function. To avoid the dense pairwise computation in $\mathbf{C}$, first-order Taylor expansion is employed to approximate the sigmoid function. Therefore, the absolute position $\mathbf{r}$ can be reformulated as
\begin{equation}
\begin{aligned}
        \mathbf{r}_{\mathtt{back}} &= \frac{n}{2} + \frac{1}{4}\mathbf{C}\mathbf{1}\\
&= \frac{n}{2} + \frac{1}{4}\texttt{ReLU}\left(\mathbf{r}_A \mathbf{1}^\top-\mathbf{1}\mathbf{r}_A^\top\right)\mathbf{1}.
\end{aligned}
\end{equation}
For each node, only nodes with a lower ranking value contribute to the second term. Therefore, given the ranking result $\mathbf{r}$ and the approximation position $\mathbf{r}_A$ in the forward propagation, the second term of $\mathbf{r}_{\mathtt{back}}$ can be reformulated as
\begin{equation}
\begin{aligned}
\frac{1}{4}\hat{\mathbf{r}}_{i}=&\frac{1}{4}\left[\texttt{ReLU}\left(\mathbf{r}_A \mathbf{1}^\top-\mathbf{1}\mathbf{r}_A^\top\right)\mathbf{1}\right]_i\\
=&\frac{1}{4}\mathbf{r}_i\mathbf{r}_{Ai} -
        \quad\mathclap{\ \sum_{j \in \!\{c|\mathbf{r}_c <\mathbf{r}_i\!\} }}
        \quad\mathbf{r}_{Aj}.
\end{aligned}
\end{equation}
This gives rise to the formulation of backpropagation as
\begin{equation}
\begin{matrix}
    \texttt{(approx.)} &
    \begin{aligned}
        \frac{\partial\mathbf{r}}{\partial\mathbf{r}_A} &= 
        \frac{\partial}{\partial\mathbf{r}_A} 
        \ \frac{1}{4}\hat{\mathbf{r}},\\
        \hat{\mathbf{r}}_i &= \mathbf{r}_i\mathbf{r}_{Ai} -
        \quad\mathclap{\ \sum_{j \in \!\{c|\mathbf{r}_c <\mathbf{r}_i\!\} }}
        \quad\mathbf{r}_{Aj}.
    \end{aligned}
\end{matrix}
\end{equation}

\begin{figure}[htb]
\vskip 0.1in
\centering
 \subfigure[Graph Level]{\label{fig:graph-str}
      \begin{minipage}[t]{\linewidth}
      \centering
      \includegraphics[width=\textwidth]{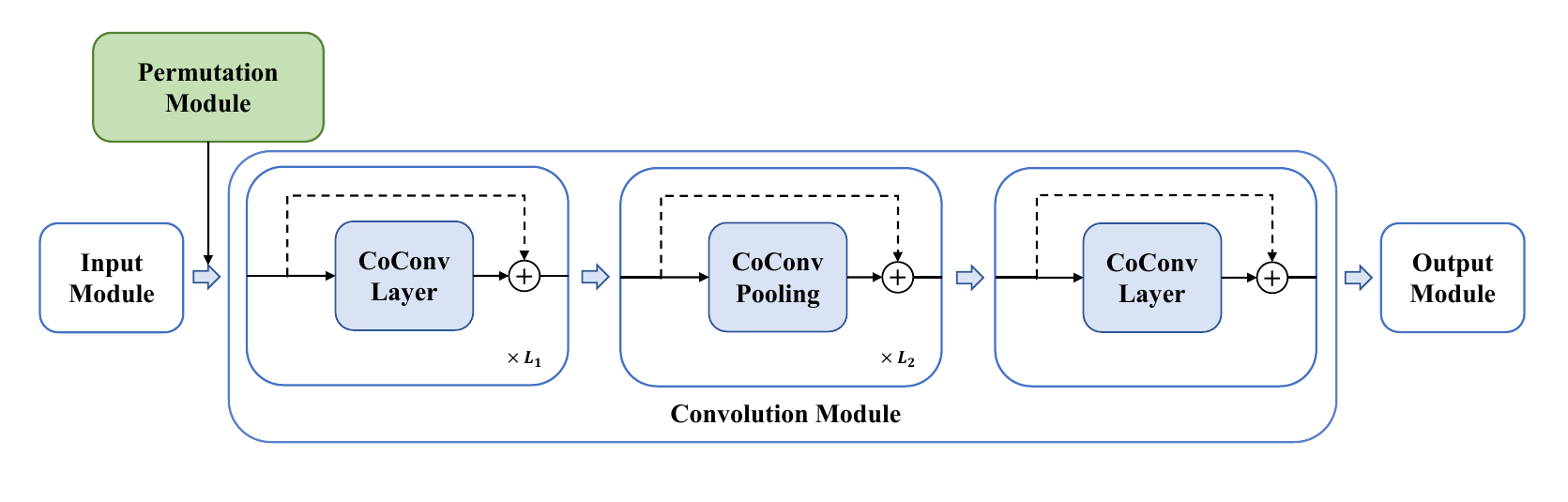}
      \end{minipage}
      }
\vskip 0.1in
\subfigure[Node Level]{\label{fig:node-str}
      \begin{minipage}[t]{\linewidth}
      \centering
      \includegraphics[width=\textwidth]{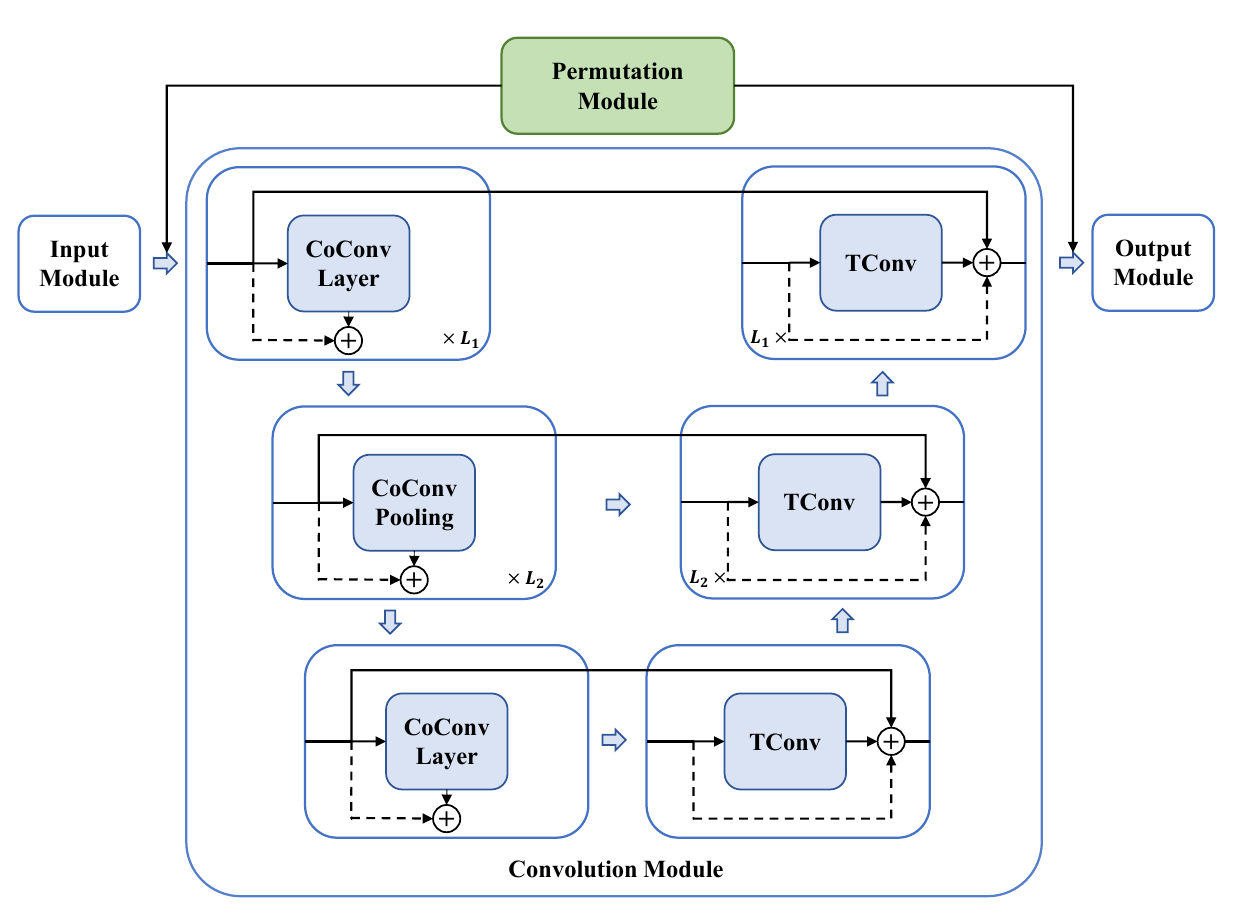}
      \end{minipage}
      }
\caption{\textbf{Compressed Convolution Network Structure.} The CoConv layer represents compressed convolution layers. Pooling represents compressed pooling layers. TConv represents transposed convolution with ReLU. The dashed lines represent shortcut with average pooling}
\label{fig:network-str}
\end{figure}

\section{Details for Experiments}\label{app:setup}
\subsection{Graph-level Tasks}
\textbf{Benchmarks.} We evaluate CoCN on nine benchmarks for graph classification, including three biochemical benchmarks~\cite{Morris+2020} (MUTAG, PROTEINS, NCI1), three social network benchmarks~\cite{Morris+2020} (COLLAB, IMDB-BINARY, IMDB-MULTI), and three brain connectomics benchmarks~\cite{said_NeuroGraphBenchmarksGraph_2023} (HCP-Task, HCP-Gender, HCP-Age). MUTAG is a collection of nitroaromatic compounds. It includes 188 graphs of chemical compounds labeled with their mutagenicity on Salmonella typhimurium. PROTEINS is a collection of 1,113 proteins with amino acids as nodes. The goal is to predict whether the protein graphs are enzymes or not. NCI1 is collected by National Cancer Institute (NCI) with 4,110 graphs of chemical compounds. The graphs are labeled by whether the corresponding compounds are positive or negative for cell lung cancer. All three biochemical datasets use the one-hot node labels as node features. IMDB-BINARY and IMDB-MULTI are two movie actor/actress collaboration network datasets labeled by the genre of the movies. COLLAB is a scientific collaboration network dataset. Each graph represents a researcher's ego network with collaborators and is labeled by the scientific study field of the researcher. HCP-Task, HCP-Gender, and HCP-Age represent three collections of brain connectomics graphs constructed from functional MRI (fMRI) data. Gender classification is a binary task distinguishing between male and female subjects. Age prediction involves categorizing individuals into three distinct groups: 22-25 years, 26-30 years, and 31-35 years. The task state prediction encompasses seven domains: Emotion Processing, Gambling, Language, Motor, Relational Processing, Social Cognition, and Working Memory.

We also conduct graph isomorphism tests on Graph8c, sr25~\cite{balcilar_BreakingLimitsMessage_2021}, and EXP~\cite{abboud_SurprisingPowerGraph_2021}. Graph8c is constituted with 11,117 possible connected non-isomorphic graphs, where each graph contains 8 nodes. We perform pairwise comparisons across all graphs within this dataset, resulting in over 61M comparative evaluations. Sr25 comprises 15 strongly regular graphs, where each graph contains 25 nodes with a degree of 12 for every node. Evaluations are performed across all pairwise combinations of these graphs, resulting in 105 distinct graph pair comparisons. EXP contains 600 pairs of 1-WL equivalent graphs. Following Balcilar et al., in \cite{balcilar_BreakingLimitsMessage_2021}, we limit the number of parameters for CoCN to 30K.

\textbf{Baselines.} For baseline models, we consider \textbf{(1)} GNNs with different convolution designs including PATCHY-SAN~\cite{niepert_LearningConvolutionalNeural_2016}, ChebNet~\cite{defferrard_ConvolutionalNeuralNetworks_2016}, GCN~\cite{kipf_SemiSupervisedClassificationGraph_2017}, ResGCN~\cite{li_DeepGCNsCanGCNs_2019}, GraphSAGE~\cite{hamilton_InductiveRepresentationLearning_2017}, GAT~\cite{velickovic_GraphAttentionNetworks_2018}, GIN~\cite{xu*_HowPowerfulAre_2019}, SGC~\cite{wu_SimplifyingGraphConvolutional_2019}, k-GNN~\cite{morris_WeisfeilerLemanGo_2019}, PPGN~\cite{maron_ProvablyPowerfulGraph_2019}, 
GNNML~\cite{balcilar_BreakingLimitsMessage_2021}, PG-GNN~\cite{huang_GoingDeeperPermutationSensitive_2022}, PathNet~\cite{michel_PathNeuralNetworks_2023} and PathNN~\cite{sun_HomophilyStructureawarePath_2022}; \textbf{(2)} Node drop based hierarchical graph pooling models including TopKPool~\cite{gao_GraphUNets_2019}, SAGPool~\cite{lee_SelfAttentionGraphPooling_2019} and ASAP~\cite{ranjan_ASAPAdaptiveStructure_2020}; and \textbf{(3)} Node clustering based hierarchical graph pooling models including DiffPool~\cite{ying_HierarchicalGraphRepresentation_2018}, 
SEP~\cite{wu_StructuralEntropyGuided_2022} and GMT~\cite{baek_AccurateLearningGraph_2022}. For brain connectomics benchmarks, we also follow the original experimental setting~\cite{said_NeuroGraphBenchmarksGraph_2023} and include CNN and Random Forest (RF).

\textbf{Experimental Setup.} For experiments on all datasets, the learning rate is set $\in \{1\times 10^{-4}, 1\times 10^{-3}\}$, the hidden size is set to 64, and the dropout rate is set $\in \{0.3, 0.5\}$. For Adam \cite{kingma_AdamMethodStochastic_2015} optimizer, weight decay is set $\in\{1\times 10^{-4}, 5\times 10^{-4}, 1\times 10^{-3}, 1\times 10^{-2}, 1\times 10^{-1}\}$. We also use early stopping regularization, where we stop the training if there is no further reduction in the validation loss during 100 epochs. The maximum epoch number is set to 200. The batch size is set to 4 on MUTAG and PROTEINS, 8 on NCI1, IMDB-BINARY and IMDB-MULTI, and 32 on COLLAB, HCP-Task, HCP-Gender, and HCP-Age. We follow the network structure in Fig.~\ref{fig:graph-str} to construct CoCN. For all datasets, the smoothness parameter $t$ in Eq.~\ref{eq:approximate-ex-pos-implementation} is set $\in\{6,8\}$, the number of permutations is set to $8$, the relaxation factor $\tau$ is set $\in \{0.1,1,10\}$, $L_1$ is set $\in\{1, 3, 4\}$, $L_2$ is set to $1$, and the kernel size is set $\in \{5,7,9\}$.

\begin{table}[t]
\caption{\textbf{Detailed Experimental Setup For Node-level Tasks.}}
\label{tab:node-exp-setting}
\begin{center}
\begin{small}
\begin{sc}
\begin{tabular}{lcccc} 
\toprule
                & $t$ & $L_1$ & $L_2$ & Dropout  \\ 
\midrule
CHAMELEON       & \{0, 1\}   & \multicolumn{2}{c}{\{0, 1, 2, 3, 4\}}     & \{0.1, 0.5\}      \\
SQUIRREL        & 1   & \multicolumn{2}{c}{\{0, 1, 2, 3, 4\}}     & \{0.1, 0.5\}      \\
CORNELL         & 0   & 1     & 3     & 0.5      \\
TEXAS           & 0   & 1     & 3     & 0.5      \\
WISCONSIN       & 0   & 1     & 3     & 0.5      \\
ACTOR           & 0   & 2     & 4     & 0.3      \\
genius          & 6   & 4     & 1     & 0.1      \\
questions       & 6   & 4     & 1     & 0.1      \\
amazon-ratings  & 6   & 4     & 0     & 0.5      \\
tolokers        & 1   & 1     & 3     & 0.1      \\
minesweeper     & 4   & 5     & 1     & 0.4      \\
CoauthorCS      & 6   & 4     & 1     & 0.5      \\
CoauthorPhysics & 6   & 4     & 1     & 0.7      \\
AmazonPhoto     & 6   & 5     & 1     & 0.5      \\
AmazonComputers & 6   & 6     & 1     & 0.5      \\
PCQM-Contact    & 6   & 4     & 1     & 0.0      \\
\bottomrule
\end{tabular}
\end{sc}
\end{small}
\end{center}
\end{table}

\subsection{Node-level Tasks}
\textbf{Benchmarks.} For node-level tasks, we include six small-scale node classification benchmarks (Chameleon, Squirrel \cite{rozemberczki_MultiScaleAttributedNode_2021}, Cornell, Texas, Wisconsin \cite{pei_GeomGCNGeometricGraph_2020} and Actor \cite{tang_SocialInfluenceAnalysis_2009}), five large-scale heterophilic benchmarks (amazon-ratings, minesweeper, tolokers, questions~\cite{platonov_CriticalLookEvaluation_2023}, genius~\cite{lim_LargeScaleLearning_2021}), and four large-scale homophilic benchmarks~\cite{shchur_PitfallsGraphNeural_2019} (CoauthorCS, CoauthorPhysics, AmazonPhoto, AmazonComputers). 

Chameleon and Squirrel are collected from Wikipedia. Each graph corresponds to a topic in Wikipedia. The nodes represent Web pages with keywords of the pages as node features. The edges are links between Web pages. Based on the average monthly traffic, the nodes are classified into five classes. Cornell, Texas, and Wisconsin are Web page datasets collected from computer departments of different universities. Actor is an actor-actor relation dataset with nodes representing actors. For each node, the features are the keywords of the corresponding actor on Wikipedia. Two nodes are connected if they appear on the same page. 

CoauthorCS and CoauthorPhysics are collected from the Microsoft Academic Graph. Each graph represents a co-authorship graph where nodes denote researchers and edges denote co-authorship relationships. The task is to categorize each author's dominant research domain as either computer science or physics. AmazonComputers and AmazonPhotos constitute Amazon co-purchase graphs. Nodes denote products and edges denote co-purchase relation between pairs of items. Graph labels categorize the products. 

Questions is collected from the Yandex Q question-answering website, constituted user activity graphs from September 2021 to August 2022. In these graphs, nodes represent users interested in 'medicine', and edges indicate one user answering another user's posted question. The task at hand is to predict which users remained active on the website. Amazon-ratings utilizes the Amazon product co-purchasing network metadata sourced from the SNAP Datasets~\cite{snapnets}. Nodes represent products, and edges encode frequent co-purchase relations between pairs of items. The task involves predicting the average reviewer rating for each product. Tolokers is constituted of crowdsourcing participation data sourced from the Toloka platform. Nodes denote contributors~(tolokers) who have participated in at least one out of 13 selected projects. Edges link pairs of tolokers who have completed the same tasks. The associated task is to predict which tolokers have been banned from projects. Minesweeper is a synthetic 100x100 grid network. Nodes denote grid cells. The associated prediction task is to classify which nodes are mine or not. Genius is an online social network with nodes representing users. The task is to predict whether the accounts are marked or not.

We also evaluate CoCN on a link prediction benchmark PCQM-Contact~\cite{dwivedi_LongRangeGraph_2022}. PCQM-Contact is collected for molecular property prediction evaluation with 529,434 molecular graphs. The task is to predict whether pairs of nodes contact each other in the 3D space.

\textbf{Baselines.}~
We consider GNNs with different convolution designs which can make node-level predictions, including ChebNet~\cite{defferrard_ConvolutionalNeuralNetworks_2016}, GCN~\cite{kipf_SemiSupervisedClassificationGraph_2017}, GraphSAGE\cite{hamilton_InductiveRepresentationLearning_2017}, GatedGCN~\cite{bresson_ResidualGatedGraph_2018}, APPNP~\cite{gasteiger_PredictThenPropagate_2018}, GAT~\cite{velickovic_GraphAttentionNetworks_2018}, SGC~\cite{wu_SimplifyingGraphConvolutional_2019}, MixHop~\cite{abu-el-haija_MixHopHigherOrderGraph_2019}, Geom-GCN~\cite{pei_GeomGCNGeometricGraph_2020}, GCNII~\cite{chen_SimpleDeepGraph_2020}, H$_2$GCN~\cite{zhu_HomophilyGraphNeural_2020}, GINE~\cite{hu*_StrategiesPretrainingGraph_2019, xu_HowNeuralNetworks_2021}, FAGCN~\cite{bo_LowfrequencyInformationGraph_2021}, LINKX~\cite{lim_LargeScaleLearning_2021}, GRAND~\cite{chamberlain_GRANDGraphNeural_2021}, GloGNN~\cite{li_FindingGlobalHomophily_2022}, FSGNN~\cite{maurya_SimplifyingApproachNode_2022}, GBK-GNN~\cite{du_GBKGNNGatedBiKernel_2022}, JacobiConv~\cite{wang_HowPowerfulAre_2022}, GGCN~\cite{yan_TwoSidesSame_2022}, GPRGCN~\cite{chien_AdaptiveUniversalGeneralized_2022}, ACMII-GCN~\cite{luan_RevisitingHeterophilyGraph_2022}, Graph Diffuser~\cite{glickman_DiffusingGraphAttention_2023}, A-DGN~\cite{gravina_AntiSymmetricDGNStable_2023}, AERO-GNN~\cite{lee_DeepAttentionGraph_2023}, Drew-GCN~\cite{gutteridge_DRewDynamicallyRewired_2023}, Cache-GNN~\cite{ma_AugmentingRecurrentGraph_2023} with random walk positional encoding~\cite{dwivedi_GraphNeuralNetworks_2021} and ResNet with GNN methods\cite{platonov_CriticalLookEvaluation_2023}.

\textbf{Experimental Setup.} For experiments on all datasets, the learning rate is set $\in \{1\times 10^{-4}, 1\times 10^{-3}\}$, the hidden size is set $\in\{64,128\}$, and the weight decay is set $\in\{1\times 10^{-4}, 5\times 10^{-4}, 1\times 10^{-3}, 1\times 10^{-2}, 1\times 10^{-1}\}$. The maximum epoch number is set to 1,000. We stop the training if there is no further reduction in the validation loss during 150 epochs.  We follow the network structure in Fig.~\ref{fig:node-str} to construct CoCN. The number of permutations is set $\in\{8, 10\}$, the relaxation factor $\tau$ is set  $\in \{0.1,1,10\}$ and the kernel size is set to $5$. The rest of the hyper-parameter setting and detailed network structure for all datasets are summarized in Tab.~\ref{tab:node-exp-setting}.

\subsection{Model Analysis}
For all the experiments in Subsection \ref{ssec:ablation}, the number of permutations is set to $8$, the learning rate is set to $1\times 10^{-3}$ for large-scale node classification benchmarks and $1\times 10^{-4}$ for the others, and the hidden size is set to $64$. For the permutation information loss experiment, we construct the autoencoder with the permutation generation module. The position smoothness parameter $t$ is set to $6$ and the dropout rate is set to $0$, the maximum epoch is set to $10,000$ and the early stopping patience is set to $2,000$ epochs. For the rest of the experiments, the experimental setup is the same as the classification experiments except for the ablation items.

\begin{table}[t]
\caption{\textbf{Full comparison Node Classification Results (measured by accuracy: \%) on Original / Filtered Chameleon and Squirrel.}}\label{tab:app-filter}
\begin{center}
\begin{threeparttable}
\begin{small}
\setlength{\tabcolsep}{3pt}
\begin{tabular}{lcccccc} 
\toprule
            & \multicolumn{3}{c}{Chameleon} & \multicolumn{3}{c}{Squirrel}  \\ 
\cline{2-7}
            & Ori.  & Filt. & Ranks         & Ori.  & Filt. & Ranks         \\ 
\midrule
ResNet~\cite{platonov_CriticalLookEvaluation_2023}      & 49.52 & 36.73 & 12/13         & 33.88 & 36.55 & 12/8          \\
ResNet+SGC~\cite{wu_SimplifyingGraphConvolutional_2019}  & 49.93 & 41.01 & 11/4          & 34.36 & 38.36 & 11/7          \\
ResNet+adj~\cite{platonov_CriticalLookEvaluation_2023}  & 71.07 & 38.67 & 4/11          & 65.46 & 38.37 & 4/6           \\
GCN~\cite{kipf_SemiSupervisedClassificationGraph_2017}         & 50.18 & 40.89 & 9.5/5         & 39.06 & 39.47 & 8/4           \\
GraphSAGE~\cite{hamilton_InductiveRepresentationLearning_2017}   & 50.18 & 37.77 & 9.5/12        & 35.83 & 36.09 & 10/9           \\
GAT~\cite{velickovic_GraphAttentionNetworks_2018}         & 45.02 & 39.21 & 15/9          & 32.21 & 35.62 & 14/11         \\
H$_2$GCN~\cite{zhu_HomophilyGraphNeural_2020}     & 46.27 & 26.75 & 14/14         & 29.45 & 35.10 & 15/14         \\
GPRGNN~\cite{chien_AdaptiveUniversalGeneralized_2022}      & 47.26 & 39.93 & 13/7          & 33.39 & 38.95 & 13/5          \\
FSGNN~\cite{maurya_SimplifyingApproachNode_2022}       & 77.85 & 40.61 & 3/6           & 68.93 & 35.92 & 3/10           \\
GloGNN~\cite{li_FindingGlobalHomophily_2022}      & 70.04 & 25.90 & 5/15          & 61.21 & 35.11 & 5/13          \\
FAGCN~\cite{bo_LowfrequencyInformationGraph_2021}       & 64.23 & 41.90 & 7/3           & 47.63 & 41.08 & 6/2           \\
GBK-GNN~\cite{du_GBKGNNGatedBiKernel_2022}     & 51.36 & 39.61 & 8/8           & 37.06 & 35.51 & 9/13          \\
JacobiConv~\cite{wang_HowPowerfulAre_2022}  & 68.33 & 39.00 & 6/10           & 46.17 & 29.71 & 7/15          \\ 
\midrule
\textbf{CoCN va.~(Ours)} & 79.17 & 41.95 & 2/2           & 72.95 & 39.69 & 2/3           \\
\textbf{CoCN exp.~(Ours)} & \textbf{79.35} & \textbf{43.15} & \textbf{1}/\textbf{1}           & \textbf{72.98} & \textbf{41.57} & \textbf{1}/\textbf{1}           \\
\bottomrule
\end{tabular}
\end{small}
\begin{tablenotes}
    \item CoCN va. denotes CoCN vanilla and CoCN exp. denotes CoCN expanded.
\end{tablenotes}
\end{threeparttable}
\end{center}
\end{table}


\ifCLASSOPTIONcaptionsoff
  \newpage
\fi
\bibliography{biblio}

\begin{thebibliography}{100}
\providecommand{\url}[1]{#1}
\csname url@samestyle\endcsname
\providecommand{\newblock}{\relax}
\providecommand{\bibinfo}[2]{#2}
\providecommand{\BIBentrySTDinterwordspacing}{\spaceskip=0pt\relax}
\providecommand{\BIBentryALTinterwordstretchfactor}{4}
\providecommand{\BIBentryALTinterwordspacing}{\spaceskip=\fontdimen2\font plus
\BIBentryALTinterwordstretchfactor\fontdimen3\font minus \fontdimen4\font\relax}
\providecommand{\BIBforeignlanguage}[2]{{%
\expandafter\ifx\csname l@#1\endcsname\relax
\typeout{** WARNING: IEEEtran.bst: No hyphenation pattern has been}%
\typeout{** loaded for the language `#1'. Using the pattern for}%
\typeout{** the default language instead.}%
\else
\language=\csname l@#1\endcsname
\fi
#2}}
\providecommand{\BIBdecl}{\relax}
\BIBdecl

\bibitem{kipf_SemiSupervisedClassificationGraph_2017}
T.~N. Kipf and M.~Welling, ``Semi-{{Supervised Classification}} with {{Graph Convolutional Networks}},'' in \emph{International Conference on Learning Representations}, {Toulon, France}, 2017.

\bibitem{velickovic_GraphAttentionNetworks_2018}
P.~Veli{\v c}kovi{\'c}, G.~Cucurull, A.~Casanova, A.~Romero, P.~Li{\`o}, and Y.~Bengio, ``Graph {{Attention Networks}},'' in \emph{International {{Conference}} on {{Learning Representations}}}, {Vancouver, Canada}, 2018.

\bibitem{maurya_SimplifyingApproachNode_2022}
S.~K. Maurya, X.~Liu, and T.~Murata, ``Simplifying approach to node classification in {{Graph Neural Networks}},'' \emph{Journal of Computational Science}, vol.~62, p. 101695, 2022.

\bibitem{zhang_LinkPredictionBased_2018}
M.~Zhang and Y.~Chen, ``Link {{Prediction Based}} on {{Graph Neural Networks}},'' in \emph{International {{Conference}} on {{Neural Information Processing Systems}}}.\hskip 1em plus 0.5em minus 0.4em\relax Red Hook, NY, USA: Curran Associates Inc., 2018, pp. 5171--5181.

\bibitem{ying_TransformersReallyPerform_2021}
C.~Ying, T.~Cai, S.~Luo, S.~Zheng, G.~Ke, D.~He, Y.~Shen, and T.-Y. Liu, ``Do {{Transformers Really Perform Bad}} for {{Graph Representation}}?'' in \emph{Advances in {{Neural Information Processing Systems}}}, vol.~34.\hskip 1em plus 0.5em minus 0.4em\relax Curran Associates, Inc., 2021, pp. 28\,877--28\,888.

\bibitem{li_DeepGCNsMakingGCNs_2021}
G.~Li, M.~M{\"u}ller, G.~Qian, I.~C. Delgadillo, A.~Abualshour, A.~Thabet, and B.~Ghanem, ``{{DeepGCNs}}: {{Making GCNs Go}} as {{Deep}} as {{CNNs}},'' \emph{IEEE Transactions on Pattern Analysis and Machine Intelligence}, pp. 1--1, 2021.

\bibitem{zhang_DynamicGraphMessage_2022}
L.~Zhang, M.~Chen, A.~Arnab, X.~Xue, and P.~H.~S. Torr, ``Dynamic {{Graph Message Passing Networks}} for {{Visual Recognition}},'' \emph{IEEE Transactions on Pattern Analysis and Machine Intelligence}, pp. 1--17, Sep. 2022.

\bibitem{bessadok_GraphNeuralNetworks_2023}
A.~Bessadok, M.~A. Mahjoub, and I.~Rekik, ``Graph {{Neural Networks}} in {{Network Neuroscience}},'' \emph{IEEE Transactions on Pattern Analysis and Machine Intelligence}, vol.~45, no.~5, pp. 5833--5848, May 2023.

\bibitem{velickovic_MessagePassingAll_2022}
P.~Veli{\v c}kovi{\'c}, ``Message passing all the way up,'' \emph{arXiv}, 2022.

\bibitem{zhou_GraphNeuralNetworks_2020}
J.~Zhou, G.~Cui, S.~Hu, Z.~Zhang, C.~Yang, Z.~Liu, L.~Wang, C.~Li, and M.~Sun, ``Graph neural networks: {{A}} review of methods and applications,'' \emph{AI Open}, vol.~1, pp. 57--81, Jan. 2020.

\bibitem{alon_BottleneckGraphNeural_2021}
U.~Alon and E.~Yahav, ``On the {{Bottleneck}} of {{Graph Neural Networks}} and its {{Practical Implications}},'' in \emph{International Conference for Learning Representations}, Virtual Only, 2021.

\bibitem{topping_UnderstandingOversquashingBottlenecks_2021}
J.~Topping, F.~Di~Giovanni, B.~P. Chamberlain, X.~Dong, and M.~M. Bronstein, ``Understanding over-squashing and bottlenecks on graphs via curvature,'' in \emph{International {{Conference}} for {{Learning Representations}}}, Nov. 2021.

\bibitem{digiovanni_OversquashingMessagePassing_2023a}
F.~Di~Giovanni, L.~Giusti, F.~Barbero, G.~Luise, P.~Li{\`o}, and M.~Bronstein, ``On over-squashing in message passing neural networks: The impact of width, depth, and topology,'' in \emph{{{International Conference}} on {{Machine Learning}}}, vol. 202.\hskip 1em plus 0.5em minus 0.4em\relax Honolulu, Hawaii, USA: JMLR.org, 2023, pp. 7865--7885.

\bibitem{abu-el-haija_MixHopHigherOrderGraph_2019}
S.~{Abu-El-Haija}, B.~Perozzi, A.~Kapoor, N.~Alipourfard, K.~Lerman, H.~Harutyunyan, G.~V. Steeg, and A.~Galstyan, ``{{MixHop}}: {{Higher-Order Graph Convolutional Architectures}} via {{Sparsified Neighborhood Mixing}},'' in \emph{International {{Conference}} on {{Machine Learning}}}.\hskip 1em plus 0.5em minus 0.4em\relax {Long Beach, USA}: {PMLR}, 2019, pp. 21--29.

\bibitem{klicpera_DiffusionImprovesGraph_2019}
J.~Klicpera, S.~{Wei{\ss} enberger}, and S.~G{\"u}nnemann, ``Diffusion {{Improves Graph Learning}},'' in \emph{International {{Conference}} on {{Advances}} in {{Neural Information Processing Systems}}}, vol.~32.\hskip 1em plus 0.5em minus 0.4em\relax {Vancouver, Canada}: {Curran Associates, Inc.}, 2019, pp. 13\,333--13\,345.

\bibitem{rong_DropEdgeDeepGraph_2020}
Y.~Rong, W.~Huang, T.~Xu, and J.~Huang, ``{{DropEdge}}: {{Towards Deep Graph Convolutional Networks}} on {{Node Classification}},'' in \emph{International {{Conference}} for {{Learning Representations}}}, Addis Ababa, Ethiopia, Mar. 2020.

\bibitem{black_UnderstandingOversquashingGNNs_2023}
M.~Black, Z.~Wan, A.~Nayyeri, and Y.~Wang, ``Understanding {{Oversquashing}} in {{GNNs}} through the {{Lens}} of {{Effective Resistance}},'' in \emph{Proceedings of the 40th {{International Conference}} on {{Machine Learning}}}.\hskip 1em plus 0.5em minus 0.4em\relax PMLR, Jul. 2023, pp. 2528--2547.

\bibitem{zhang_GraphConvolutionalNetworks_2019}
S.~Zhang, H.~Tong, J.~Xu, and R.~Maciejewski, ``Graph convolutional networks: A comprehensive review,'' \emph{Computational Social Networks}, vol.~6, no.~1, p.~11, Nov. 2019.

\bibitem{ying_HierarchicalGraphRepresentation_2018}
Z.~Ying, J.~You, C.~Morris, X.~Ren, W.~Hamilton, and J.~Leskovec, ``Hierarchical {{Graph Representation Learning}} with {{Differentiable Pooling}},'' in \emph{Advances in {{Neural Information Processing Systems}}}, vol.~31.\hskip 1em plus 0.5em minus 0.4em\relax {Montr\'eal, Canada}: {Curran Associates, Inc.}, 2018, pp. 4805--4815.

\bibitem{gao_GraphUNets_2019}
H.~Gao and S.~Ji, ``Graph {{U-Nets}},'' in \emph{International {{Conference}} on {{Machine Learning}}}.\hskip 1em plus 0.5em minus 0.4em\relax {Long Beach, California, USA}: {PMLR}, 2019, pp. 2083--2092.

\bibitem{he_DeepResidualLearning_2016}
K.~He, X.~Zhang, S.~Ren, and J.~Sun, ``Deep {{Residual Learning}} for {{Image Recognition}},'' in \emph{{{IEEE Conference}} on {{Computer Vision}} and {{Pattern Recognition}}}, {Las Vegas, NV, USA}, 2016, pp. 770--778.

\bibitem{krizhevsky_ImageNetClassificationDeep_2017}
A.~Krizhevsky, I.~Sutskever, and G.~E. Hinton, ``{{ImageNet}} classification with deep convolutional neural networks,'' \emph{Communications of the ACM}, vol.~60, no.~6, pp. 84--90, 2017.

\bibitem{ji_3DConvolutionalNeural_2013}
S.~Ji, W.~Xu, M.~Yang, and K.~Yu, ``{{3D Convolutional Neural Networks}} for {{Human Action Recognition}},'' \emph{IEEE Transactions on Pattern Analysis and Machine Intelligence}, vol.~35, no.~1, pp. 221--231, 2013.

\bibitem{tran_LearningSpatiotemporalFeatures_2015}
D.~Tran, L.~Bourdev, R.~Fergus, L.~Torresani, and M.~Paluri, ``Learning {{Spatiotemporal Features}} with {{3D Convolutional Networks}},'' in \emph{{{IEEE International Conference}} on {{Computer Vision}}}.\hskip 1em plus 0.5em minus 0.4em\relax {Santiago, Chile}: {IEEE Computer Society}, 2015, pp. 4489--4497.

\bibitem{niepert_LearningConvolutionalNeural_2016}
M.~Niepert, M.~Ahmed, and K.~Kutzkov, ``Learning convolutional neural networks for graphs,'' in \emph{International {{Conference}} on {{Machine Learning}}}, vol.~48.\hskip 1em plus 0.5em minus 0.4em\relax {New York, NY, USA}: {PMLR}, 2016, pp. 2014--2023.

\bibitem{eliasof_PathGCNLearningGeneral_2022}
M.~Eliasof, E.~Haber, and E.~Treister, ``{{pathGCN}}: {{Learning General Graph Spatial Operators}} from {{Paths}},'' in \emph{International {{Conference}} on {{Machine Learning}}}.\hskip 1em plus 0.5em minus 0.4em\relax {Baltimore, Maryland, USA}: {PMLR}, 2022, pp. 5878--5891.

\bibitem{murphy_JanossyPoolingLearning_2019}
R.~L. Murphy, B.~Srinivasan, V.~Rao, and B.~Ribeiro, ``Janossy {{Pooling}}: {{Learning Deep Permutation-Invariant Functions}} for {{Variable-Size Inputs}},'' in \emph{International {{Conference}} on {{Learning Representations}}}, {New Orleans, LA, USA}, 2019.

\bibitem{murphy_RelationalPoolingGraph_2019}
------, ``Relational {{Pooling}} for {{Graph Representations}},'' in \emph{Internet {{Conference}} on {{Machine Learning}}}.\hskip 1em plus 0.5em minus 0.4em\relax {Long Beach, California, USA}: {PMLR}, 2019, pp. 4663--4673.

\bibitem{huang_GoingDeeperPermutationSensitive_2022}
Z.~Huang, Y.~Wang, C.~Li, and H.~He, ``Going {{Deeper}} into {{Permutation-Sensitive Graph Neural Networks}},'' in \emph{International {{Conference}} on {{Machine Learning}}}.\hskip 1em plus 0.5em minus 0.4em\relax {Baltimore, Maryland, USA}: {PMLR}, 2022, pp. 9377--9409.

\bibitem{sun_AllinARow_2023}
J.~Sun, S.~Wang, X.~Han, Z.~Xue, and Q.~Huang, ``All in a row: {{Compressed}} convolution networks for graphs,'' in \emph{International {{Conference}} on {{Machine Learning}}}, vol. 202.\hskip 1em plus 0.5em minus 0.4em\relax Honolulu, USA: PMLR, 2023, pp. 33\,061--33\,076.

\bibitem{bruna_SpectralNetworksLocally_2013}
J.~Bruna, W.~Zaremba, A.~Szlam, and Y.~LeCun, ``Spectral {{Networks}} and {{Locally Connected Networks}} on {{Graphs}},'' in \emph{International {{Conference}} for {{Learning Representations}}}, {Banff, Canada}, 2013.

\bibitem{defferrard_ConvolutionalNeuralNetworks_2016}
M.~Defferrard, X.~Bresson, and P.~Vandergheynst, ``Convolutional neural networks on graphs with fast localized spectral filtering,'' in \emph{International {{Conference}} on {{Neural Information Processing Systems}}}.\hskip 1em plus 0.5em minus 0.4em\relax {Red Hook, NY, USA}: {Curran Associates Inc.}, 2016, pp. 3844--3852.

\bibitem{wu_SimplifyingGraphConvolutional_2019}
F.~Wu, A.~Souza, T.~Zhang, C.~Fifty, T.~Yu, and K.~Weinberger, ``Simplifying {{Graph Convolutional Networks}},'' in \emph{International {{Conference}} on {{Machine Learning}}}.\hskip 1em plus 0.5em minus 0.4em\relax {Long Beach, USA}: {PMLR}, 2019, pp. 6861--6871.

\bibitem{dick_PathIntegralsQuantum_2020}
R.~Dick, ``Path {{Integrals}} in {{Quantum Mechanics}},'' in \emph{Advanced {{Quantum Mechanics}}: {{Materials}} and {{Photons}}}, ser. Graduate {{Texts}} in {{Physics}}, R.~Dick, Ed.\hskip 1em plus 0.5em minus 0.4em\relax Cham: Springer International Publishing, 2020, pp. 311--329.

\bibitem{ma_PathIntegralBased_2020}
Z.~Ma, J.~Xuan, Y.~G. Wang, M.~Li, and P.~Li{\`o}, ``Path {{Integral Based Convolution}} and {{Pooling}} for {{Graph Neural Networks}},'' in \emph{Advances in {{Neural Information Processing Systems}}}, vol.~33.\hskip 1em plus 0.5em minus 0.4em\relax {virtual}: {Curran Associates, Inc.}, 2020, pp. 16\,421--16\,433.

\bibitem{wang_HowPowerfulAre_2022}
X.~Wang and M.~Zhang, ``How {{Powerful}} are {{Spectral Graph Neural Networks}},'' in \emph{International {{Conference}} on {{Machine Learning}}}.\hskip 1em plus 0.5em minus 0.4em\relax {Baltimore, Maryland, USA}: {PMLR}, 2022, pp. 23\,341--23\,362.

\bibitem{guo_GraphNeuralNetworks_2023}
Y.~Guo and Z.~Wei, ``Graph {{Neural Networks}} with {{Learnable}} and {{Optimal Polynomial Bases}},'' in \emph{{{International Conference}} on {{Machine Learning}}}.\hskip 1em plus 0.5em minus 0.4em\relax PMLR, 2023, pp. 12\,077--12\,097.

\bibitem{michel_PathNeuralNetworks_2023}
G.~Michel, G.~Nikolentzos, J.~F. Lutzeyer, and M.~Vazirgiannis, ``Path {{Neural Networks}}: {{Expressive}} and {{Accurate Graph Neural Networks}},'' in \emph{{{International Conference}} on {{Machine Learning}}}.\hskip 1em plus 0.5em minus 0.4em\relax PMLR, 2023, pp. 24\,737--24\,755.

\bibitem{sun_HomophilyStructureawarePath_2022}
Y.~Sun, H.~Deng, Y.~Yang, C.~Wang, J.~Xu, R.~Huang, L.~Cao, Y.~Wang, and L.~Chen, ``Beyond {{Homophily}}: {{Structure-aware Path Aggregation Graph Neural Network}},'' in \emph{{{International Joint Conference}} on {{Artificial Intelligence}}}, vol.~3, 2022, pp. 2233--2240.

\bibitem{nguyen_RevisitingOversmoothingOversquashing_2023}
K.~Nguyen, H.~Nong, V.~Nguyen, N.~Ho, S.~Osher, and T.~Nguyen, ``Revisiting over-smoothing and over-squashing using ollivier-ricci curvature,'' in \emph{{{International Conference}} on {{Machine Learning}}}, vol. 202.\hskip 1em plus 0.5em minus 0.4em\relax Honolulu, Hawaii, USA: JMLR.org, 2023, pp. 25\,956--25\,979.

\bibitem{suresh_BreakingLimitGraph_2021}
S.~Suresh, V.~Budde, J.~Neville, P.~Li, and J.~Ma, ``Breaking the {{Limit}} of {{Graph Neural Networks}} by {{Improving}} the {{Assortativity}} of {{Graphs}} with {{Local Mixing Patterns}},'' in \emph{{{ACM SIGKDD Conference}} on {{Knowledge Discovery}} \& {{Data Mining}}}.\hskip 1em plus 0.5em minus 0.4em\relax New York, NY, USA: Association for Computing Machinery, 2021, pp. 1541--1551.

\bibitem{chen_MeasuringRelievingOverSmoothing_2020a}
D.~Chen, Y.~Lin, W.~Li, P.~Li, J.~Zhou, and X.~Sun, ``Measuring and {{Relieving}} the {{Over-Smoothing Problem}} for {{Graph Neural Networks}} from the {{Topological View}},'' in \emph{AAAI Conference on Artificial Intelligence}, vol.~34, no.~04, 2020, pp. 3438--3445.

\bibitem{arnaiz-rodriguez_DiffWireInductiveGraph_2022}
A.~{Arnaiz-Rodr{\'{\i}}guez}, A.~Begga, F.~Escolano, and N.~M. Oliver, ``{{DiffWire}}: {{Inductive Graph Rewiring}} via the {{Lov{\'a}sz Bound}},'' in \emph{Proceedings of the {{First Learning}} on {{Graphs Conference}}}.\hskip 1em plus 0.5em minus 0.4em\relax PMLR, Dec. 2022, pp. 15:1--15:27.

\bibitem{dong_UnderstandingReducingGraph_2023}
M.~Dong and Y.~Kluger, ``Towards {{Understanding}} and {{Reducing Graph Structural Noise}} for {{GNNs}},'' in \emph{Proceedings of the 40th {{International Conference}} on {{Machine Learning}}}.\hskip 1em plus 0.5em minus 0.4em\relax PMLR, Jul. 2023, pp. 8202--8226.

\bibitem{gutteridge_DRewDynamicallyRewired_2023}
B.~Gutteridge, X.~Dong, M.~M. Bronstein, and F.~D. Giovanni, ``{{DRew}}: {{Dynamically Rewired Message Passing}} with {{Delay}},'' in \emph{International {{Conference}} on {{Machine Learning}}}.\hskip 1em plus 0.5em minus 0.4em\relax PMLR, Jul. 2023, pp. 12\,252--12\,267.

\bibitem{boguna_NetworkGeometry_2021}
M.~Bogu{\~n}{\'a}, I.~Bonamassa, M.~De~Domenico, S.~Havlin, D.~Krioukov, and M.~{\'A}. Serrano, ``Network geometry,'' \emph{Nature Reviews Physics}, vol.~3, no.~2, pp. 114--135, 2021.

\bibitem{Hu2006EfficientHF}
Y.~Hu, ``Efficient, high-quality force-directed graph drawing,'' \emph{Mathematica journal}, vol.~10, pp. 37--71, 2006.

\bibitem{chen_HARPHierarchicalRepresentation_2018}
H.~Chen, B.~Perozzi, Y.~Hu, and S.~Skiena, ``{{HARP}}: {{Hierarchical Representation Learning}} for {{Networks}},'' in \emph{{{AAAI Conference}} on {{Artificial Intelligence}}}, vol.~32.\hskip 1em plus 0.5em minus 0.4em\relax {New Orleans, Louisiana, USA}: {AAAI Press}, 2018.

\bibitem{hu_HierarchicalGraphConvolutional_2019}
F.~Hu, Y.~Zhu, S.~Wu, L.~Wang, and T.~Tan, ``Hierarchical graph convolutional networks for semi-supervised node classification,'' in \emph{{{International Joint Conference}} on {{Artificial Intelligence}}}.\hskip 1em plus 0.5em minus 0.4em\relax {Macao, China}: {AAAI Press}, 2019, pp. 4532--4539.

\bibitem{deng_GraphZoomMultilevelSpectral_2020}
C.~Deng, Z.~Zhao, Y.~Wang, Z.~Zhang, and Z.~Feng, ``{{GraphZoom}}: {{A Multi-level Spectral Approach}} for {{Accurate}} and {{Scalable Graph Embedding}},'' in \emph{International {{Conference}} on {{Learning Representations}}}, {Addis Ababa, Ethiopia}, 2020.

\bibitem{wang_SecondOrderPoolingGraph_2020}
Z.~Wang and S.~Ji, ``Second-{{Order Pooling}} for {{Graph Neural Networks}},'' \emph{IEEE Transactions on Pattern Analysis and Machine Intelligence}, pp. 1--1, 2020.

\bibitem{baek_AccurateLearningGraph_2022}
J.~Baek, M.~Kang, and S.~J. Hwang, ``Accurate {{Learning}} of {{Graph Representations}} with {{Graph Multiset Pooling}},'' in \emph{International {{Conference}} on {{Learning Representations}}}, {virtual}, 2022.

\bibitem{cangea_SparseHierarchicalGraph_2018}
C.~Cangea, P.~Veli{\v c}kovi{\'c}, N.~Jovanovi{\'c}, T.~Kipf, and P.~Li{\`o}, ``Towards {{Sparse Hierarchical Graph Classifiers}},'' in \emph{Neural {{Information Processing Systems Workshop}} on {{Relational Representation Learning}}}, {Montr\'eal, Canada}, 2018.

\bibitem{lee_SelfAttentionGraphPooling_2019}
J.~Lee, I.~Lee, and J.~Kang, ``Self-{{Attention Graph Pooling}},'' in \emph{International {{Conference}} on {{Machine Learning}}}.\hskip 1em plus 0.5em minus 0.4em\relax {Long Beach, California, USA}: {PMLR}, 2019, pp. 3734--3743.

\bibitem{gao_TopologyAwareGraphPooling_2021}
H.~Gao, Y.~Liu, and S.~Ji, ``Topology-{{Aware Graph Pooling Networks}},'' \emph{IEEE Transactions on Pattern Analysis and Machine Intelligence}, vol.~43, no.~12, pp. 4512--4518, 2021.

\bibitem{wu_StructuralEntropyGuided_2022}
J.~Wu, X.~Chen, S.~Li, and K.~Xu, ``Structural entropy guided graph hierarchical pooling,'' in \emph{International Conference on Machine Learning}, vol. 162.\hskip 1em plus 0.5em minus 0.4em\relax {Baltimore, Maryland, USA}: {PMLR}, 2022, pp. 24\,017--24\,030.

\bibitem{Carter2009VisualGT}
N.~C. Carter, ``Visual group theory.''\hskip 1em plus 0.5em minus 0.4em\relax Mathematical Association of America, 2009, pp. 65--66.

\bibitem{szegedy_GoingDeeperConvolutions_2015}
C.~Szegedy, W.~Liu, Y.~Jia, P.~Sermanet, S.~Reed, D.~Anguelov, D.~Erhan, V.~Vanhoucke, and A.~Rabinovich, ``Going {{Deeper}} with {{Convolutions}},'' in \emph{{{IEEE Conference}} on {{Computer Vision}} and {{Pattern Recognition}}}, Jun. 2015, pp. 1--9.

\bibitem{noh_LearningDeconvolutionNetwork_2015}
H.~Noh, S.~Hong, and B.~Han, ``Learning {{Deconvolution Network}} for {{Semantic Segmentation}},'' in \emph{{{IEEE International Conference}} on {{Computer Vision}}}.\hskip 1em plus 0.5em minus 0.4em\relax {Santiago, Chile}: {IEEE Computer Society}, 2015, pp. 1520--1528.

\bibitem{lim_LargeScaleLearning_2022}
D.~Lim, F.~M. Hohne, X.~Li, S.~L. Huang, V.~Gupta, O.~P. Bhalerao, and S.-N. Lim, ``Large {{Scale Learning}} on {{Non-Homophilous Graphs}}: {{New Benchmarks}} and {{Strong Simple Methods}},'' in \emph{Advances in {{Neural Information Processing Systems}}}, {virtual}, 2022, pp. 20\,887--20\,902.

\bibitem{paszke_PyTorchImperativeStyle_2019}
A.~Paszke, S.~Gross, F.~Massa, A.~Lerer, J.~Bradbury, G.~Chanan, T.~Killeen, Z.~Lin, N.~Gimelshein, L.~Antiga, A.~Desmaison, A.~K{\"o}pf, E.~Yang, Z.~DeVito, M.~Raison, A.~Tejani, S.~Chilamkurthy, B.~Steiner, L.~Fang, J.~Bai, and S.~Chintala, ``{{PyTorch}}: An imperative style, high-performance deep learning library,'' in \emph{International {{Conference}} on {{Neural Information Processing Systems}}}.\hskip 1em plus 0.5em minus 0.4em\relax {Red Hook, NY, USA}: {Curran Associates Inc.}, 2019, pp. 8026--8037.

\bibitem{xu*_HowPowerfulAre_2019}
K.~Xu, W.~Hu, J.~Leskovec, and S.~Jegelka, ``How {{Powerful}} are {{Graph Neural Networks}}?'' in \emph{International {{Conference}} on {{Learning Representations}}}, {New Orleans, LA, USA}, 2019.

\bibitem{hamilton_InductiveRepresentationLearning_2017}
W.~L. Hamilton, R.~Ying, and J.~Leskovec, ``Inductive representation learning on large graphs,'' in \emph{International {{Conference}} on {{Neural Information Processing Systems}}}.\hskip 1em plus 0.5em minus 0.4em\relax {Red Hook, USA}: {Curran Associates Inc.}, 2017, pp. 1025--1035.

\bibitem{ranjan_ASAPAdaptiveStructure_2020}
E.~Ranjan, S.~Sanyal, and P.~Talukdar, ``{{ASAP}}: {{Adaptive Structure Aware Pooling}} for {{Learning Hierarchical Graph Representations}},'' in \emph{{{AAAI Conference}} on {{Artificial Intelligence}}}, vol.~34, {New York, NY, USA}, 2020, pp. 5470--5477.

\bibitem{fey_FastGraphRepresentation_2019}
M.~Fey and J.~E. Lenssen, ``Fast {{Graph Representation Learning}} with {{PyTorch Geometric}},'' in \emph{International {{Conference}} on {{Learning Representations Workshop}} on {{Graphs}} and {{Manifolds}}}, 2019.

\bibitem{Morris+2020}
C.~Morris, N.~M. Kriege, F.~Bause, K.~Kersting, P.~Mutzel, and M.~Neumann, ``{{TUDataset}}: {{A}} collection of benchmark datasets for learning with graphs,'' in \emph{International {{Conference}} on {{Machine Learning}} Workshop on Graph Representation Learning and Beyond}, 2020.

\bibitem{said_NeuroGraphBenchmarksGraph_2023}
A.~Said, R.~Bayrak, T.~Derr, M.~Shabbir, D.~Moyer, C.~Chang, and X.~Koutsoukos, ``{{NeuroGraph}}: {{Benchmarks}} for {{Graph Machine Learning}} in {{Brain Connectomics}},'' in \emph{Advances in {{Neural Information Processing Systems}}}, vol.~36, Dec. 2023, pp. 6509--6531.

\bibitem{balcilar_BreakingLimitsMessage_2021}
M.~Balcilar, P.~Heroux, B.~Gauzere, P.~Vasseur, S.~Adam, and P.~Honeine, ``Breaking the {{Limits}} of {{Message Passing Graph Neural Networks}},'' in \emph{{{International Conference}} on {{Machine Learning}}}.\hskip 1em plus 0.5em minus 0.4em\relax PMLR, Jul. 2021, pp. 599--608.

\bibitem{abboud_SurprisingPowerGraph_2021}
R.~Abboud, {\.I}.~{\.I}. Ceylan, M.~Grohe, and T.~Lukasiewicz, ``The {{Surprising Power}} of {{Graph Neural Networks}} with {{Random Node Initialization}},'' in \emph{{{International Joint Conference}} on {{Artificial Intelligence}}}, vol.~3, Aug. 2021, pp. 2112--2118.

\bibitem{morris_WeisfeilerLemanGo_2019}
C.~Morris, M.~Ritzert, M.~Fey, W.~L. Hamilton, J.~E. Lenssen, G.~Rattan, and M.~Grohe, ``Weisfeiler and {{Leman Go Neural}}: {{Higher-Order Graph Neural Networks}},'' \emph{AAAI Conference on Artificial Intelligence}, vol.~33, no.~01, pp. 4602--4609, 2019.

\bibitem{li_DeepGCNsCanGCNs_2019}
G.~Li, M.~Muller, A.~Thabet, and B.~Ghanem, ``{{DeepGCNs}}: {{Can GCNs Go As Deep As CNNs}}?'' in \emph{{{IEEE}}/{{CVF International Conference}} on {{Computer Vision}}}.\hskip 1em plus 0.5em minus 0.4em\relax Seoul, South Korea: IEEE, Oct. 2019, pp. 9266--9275.

\bibitem{maron_ProvablyPowerfulGraph_2019}
H.~Maron, H.~{Ben-Hamu}, H.~Serviansky, and Y.~Lipman, ``Provably {{Powerful Graph Networks}},'' in \emph{Advances in {{Neural Information Processing Systems}}}, vol.~32.\hskip 1em plus 0.5em minus 0.4em\relax Curran Associates, Inc., 2019.

\bibitem{pei_GeomGCNGeometricGraph_2020}
H.~Pei, B.~Wei, K.~C.-C. Chang, Y.~Lei, and B.~Yang, ``Geom-{{GCN}}: {{Geometric Graph Convolutional Networks}},'' in \emph{International {{Conference}} on {{Learning Representations}}}, {Addis Ababa, Ethiopia}, 2020.

\bibitem{bo_LowfrequencyInformationGraph_2021}
D.~Bo, X.~Wang, C.~Shi, and H.~Shen, ``Beyond {{Low-frequency Information}} in {{Graph Convolutional Networks}},'' in \emph{{{AAAI Conference}} on {{Artificial Intelligence}}}, vol.~35, {virtual}, 2021, pp. 3950--3957.

\bibitem{gasteiger_PredictThenPropagate_2018}
J.~Gasteiger, A.~Bojchevski, and S.~G{\"u}nnemann, ``Predict then {{Propagate}}: {{Graph Neural Networks}} meet {{Personalized PageRank}},'' in \emph{International {{Conference}} on {{Learning Representations}}}, {Addis Ababa, Ethiopia}, 2018.

\bibitem{lim_LargeScaleLearning_2021}
D.~Lim, F.~Hohne, X.~Li, S.~L. Huang, V.~Gupta, O.~Bhalerao, and S.~N. Lim, ``Large {{Scale Learning}} on {{Non-Homophilous Graphs}}: {{New Benchmarks}} and {{Strong Simple Methods}},'' in \emph{Advances in {{Neural Information Processing Systems}}}, vol.~34.\hskip 1em plus 0.5em minus 0.4em\relax Curran Associates, Inc., 2021, pp. 20\,887--20\,902.

\bibitem{yan_TwoSidesSame_2022}
Y.~Yan, M.~Hashemi, K.~Swersky, Y.~Yang, and D.~Koutra, ``Two {{Sides}} of the {{Same Coin}}: {{Heterophily}} and {{Oversmoothing}} in {{Graph Convolutional Neural Networks}},'' in \emph{{{IEEE International Conference}} on {{Data Mining}}}.\hskip 1em plus 0.5em minus 0.4em\relax {Orlando, FL, USA}: {IEEE}, 2022, pp. 1287--1292.

\bibitem{luan_RevisitingHeterophilyGraph_2022}
S.~Luan, C.~Hua, Q.~Lu, J.~Zhu, M.~Zhao, S.~Zhang, X.-W. Chang, and D.~Precup, ``Revisiting {{Heterophily For Graph Neural Networks}},'' in \emph{Advances in {{Neural Information Processing Systems}}}, {virtual}, 2022.

\bibitem{chien_AdaptiveUniversalGeneralized_2022}
E.~Chien, J.~Peng, P.~Li, and O.~Milenkovic, ``Adaptive {{Universal Generalized PageRank Graph Neural Network}},'' in \emph{International {{Conference}} on {{Learning Representations}}}, {virtual}, 2022.

\bibitem{chen_SimpleDeepGraph_2020}
M.~Chen, Z.~Wei, Z.~Huang, B.~Ding, and Y.~Li, ``Simple and {{Deep Graph Convolutional Networks}},'' in \emph{International {{Conference}} on {{Machine Learning}}}.\hskip 1em plus 0.5em minus 0.4em\relax {virtual}: {PMLR}, 2020, pp. 1725--1735.

\bibitem{lee_DeepAttentionGraph_2023}
S.~Y. Lee, F.~Bu, J.~Yoo, and K.~Shin, ``Towards {{Deep Attention}} in {{Graph Neural Networks}}: {{Problems}} and {{Remedies}},'' in \emph{{{International Conference}} on {{Machine Learning}}}.\hskip 1em plus 0.5em minus 0.4em\relax PMLR, 2023, pp. 18\,774--18\,795.

\bibitem{rozemberczki_MultiScaleAttributedNode_2021}
B.~Rozemberczki, C.~Allen, and R.~Sarkar, ``Multi-{{Scale}} attributed node embedding,'' \emph{Journal of Complex Networks}, vol.~9, no.~2, 2021.

\bibitem{tang_SocialInfluenceAnalysis_2009}
J.~Tang, J.~Sun, C.~Wang, and Z.~Yang, ``Social influence analysis in large-scale networks,'' in \emph{{{ACM SIGKDD}} International Conference on {{Knowledge}} Discovery and Data Mining}.\hskip 1em plus 0.5em minus 0.4em\relax {New York, NY, USA}: {Association for Computing Machinery}, 2009, pp. 807--816.

\bibitem{platonov_CriticalLookEvaluation_2023}
O.~Platonov, D.~Kuznedelev, M.~Diskin, A.~Babenko, and L.~Prokhorenkova, ``A critical look at the evaluation of {{GNNs}} under heterophily: {{Are}} we really making progress?'' in \emph{International {{Conference}} on {{Learning Representations}}}, {Kigali, Rwanda}, 2023.

\bibitem{shchur_PitfallsGraphNeural_2019}
O.~Shchur, M.~Mumme, A.~Bojchevski, and S.~G{\"u}nnemann, ``Pitfalls of {{Graph Neural Network Evaluation}},'' \emph{arXiv}, 2019.

\bibitem{dwivedi_LongRangeGraph_2022}
V.~P. Dwivedi, L.~Ramp{\'a}{\v s}ek, M.~Galkin, A.~Parviz, G.~Wolf, A.~T. Luu, and D.~Beaini, ``Long {{Range Graph Benchmark}},'' in \emph{Advances in {{Neural Information Processing Systems}}}, vol.~35, 2022, pp. 22\,326--22\,340.

\bibitem{zhu_HomophilyGraphNeural_2020}
J.~Zhu, Y.~Yan, L.~Zhao, M.~Heimann, L.~Akoglu, and D.~Koutra, ``Beyond homophily in graph neural networks: Current limitations and effective designs,'' in \emph{{{International Conference}} on {{Neural Information Processing Systems}}}.\hskip 1em plus 0.5em minus 0.4em\relax {Red Hook, NY, USA}: {Curran Associates Inc.}, 2020, pp. 7793--7804.

\bibitem{li_FindingGlobalHomophily_2022}
X.~Li, R.~Zhu, Y.~Cheng, C.~Shan, S.~Luo, D.~Li, and W.~Qian, ``Finding {{Global Homophily}} in {{Graph Neural Networks When Meeting Heterophily}},'' in \emph{International {{Conference}} on {{Machine Learning}}}, vol. 162.\hskip 1em plus 0.5em minus 0.4em\relax {Baltimore, Maryland, USA}: {PMLR}, 2022, pp. 13\,242--13\,256.

\bibitem{chamberlain_GRANDGraphNeural_2021}
B.~Chamberlain, J.~Rowbottom, M.~I. Gorinova, M.~Bronstein, S.~Webb, and E.~Rossi, ``{{GRAND}}: {{Graph Neural Diffusion}},'' in \emph{{{International Conference}} on {{Machine Learning}}}.\hskip 1em plus 0.5em minus 0.4em\relax PMLR, 2021, pp. 1407--1418.

\bibitem{gravina_AntiSymmetricDGNStable_2023}
A.~Gravina, D.~Bacciu, and C.~Gallicchio, ``Anti-{{Symmetric DGN}}: A stable architecture for {{Deep Graph Networks}},'' in \emph{International {{Conference}} on {{Learning Representations}}}, 2023.

\bibitem{ma_AugmentingRecurrentGraph_2023}
G.~Ma, V.~A. Vo, T.~L. Willke, and N.~K. Ahmed, ``Augmenting recurrent graph neural networks with a cache,'' in \emph{ACM SIGKDD Conference on Knowledge Discovery and Data Mining}.\hskip 1em plus 0.5em minus 0.4em\relax New York, NY, USA: Association for Computing Machinery, 2023, p. 1608–1619.

\bibitem{glickman_DiffusingGraphAttention_2023}
D.~Glickman and E.~Yahav, ``Diffusing {{Graph Attention}},'' \emph{arXiv}, 2023.

\bibitem{bresson_ResidualGatedGraph_2018}
X.~Bresson and T.~Laurent, ``Residual {{Gated Graph ConvNets}},'' \emph{arXiv}, 2018.

\bibitem{hu*_StrategiesPretrainingGraph_2019}
W.~Hu*, B.~Liu*, J.~Gomes, M.~Zitnik, P.~Liang, V.~Pande, and J.~Leskovec, ``Strategies for {{Pre-training Graph Neural Networks}},'' in \emph{International {{Conference}} on {{Learning Representations}}}, 2019.

\bibitem{dwivedi_BenchmarkingGraphNeural_2020}
V.~P. Dwivedi, C.~K. Joshi, T.~Laurent, Y.~Bengio, and X.~Bresson, ``Benchmarking {{Graph Neural Networks}},'' \emph{arXiv}, 2020.

\bibitem{dwivedi_GraphNeuralNetworks_2021}
V.~P. Dwivedi, A.~T. Luu, T.~Laurent, Y.~Bengio, and X.~Bresson, ``Graph {{Neural Networks}} with {{Learnable Structural}} and {{Positional Representations}},'' in \emph{International Conference on Representation Learning}, 2022.

\bibitem{bianchi_SpectralClusteringGraph_2020}
F.~M. Bianchi, D.~Grattarola, and C.~Alippi, ``Spectral clustering with graph neural networks for graph pooling,'' in \emph{International {{Conference}} on {{Machine Learning}}}.\hskip 1em plus 0.5em minus 0.4em\relax {virtual}: {PMLR}, 2020, pp. 874--883.

\bibitem{kingma_AdamMethodStochastic_2015}
D.~P. Kingma and J.~Ba, ``Adam: {{A Method}} for {{Stochastic Optimization}},'' in \emph{International {{Conference}} for {{Learning Representations}}}, {San Diego, USA}, 2015.

\bibitem{snapnets}
J.~Leskovec and A.~Krevl, ``{SNAP Datasets}: {Stanford} large network dataset collection,'' \url{http://snap.stanford.edu/data}, 2014.

\bibitem{xu_HowNeuralNetworks_2021}
K.~Xu, M.~Zhang, J.~Li, S.~S. Du, K.-i. Kawarabayashi, and S.~Jegelka, ``How {{Neural Networks Extrapolate}}: {{From Feedforward}} to {{Graph Neural Networks}},'' \emph{arXiv}, Mar. 2021.

\bibitem{du_GBKGNNGatedBiKernel_2022}
L.~Du, X.~Shi, Q.~Fu, X.~Ma, H.~Liu, S.~Han, and D.~Zhang, ``{{GBK-GNN}}: {{Gated Bi-Kernel Graph Neural Networks}} for {{Modeling Both Homophily}} and {{Heterophily}},'' in \emph{{{ACM Web Conference}}}.\hskip 1em plus 0.5em minus 0.4em\relax {New York, NY, USA}: {Association for Computing Machinery}, 2022, pp. 1550--1558.

\end{thebibliography}

\begin{IEEEbiography}[{\includegraphics[width=1in,height=1.25in,keepaspectratio]{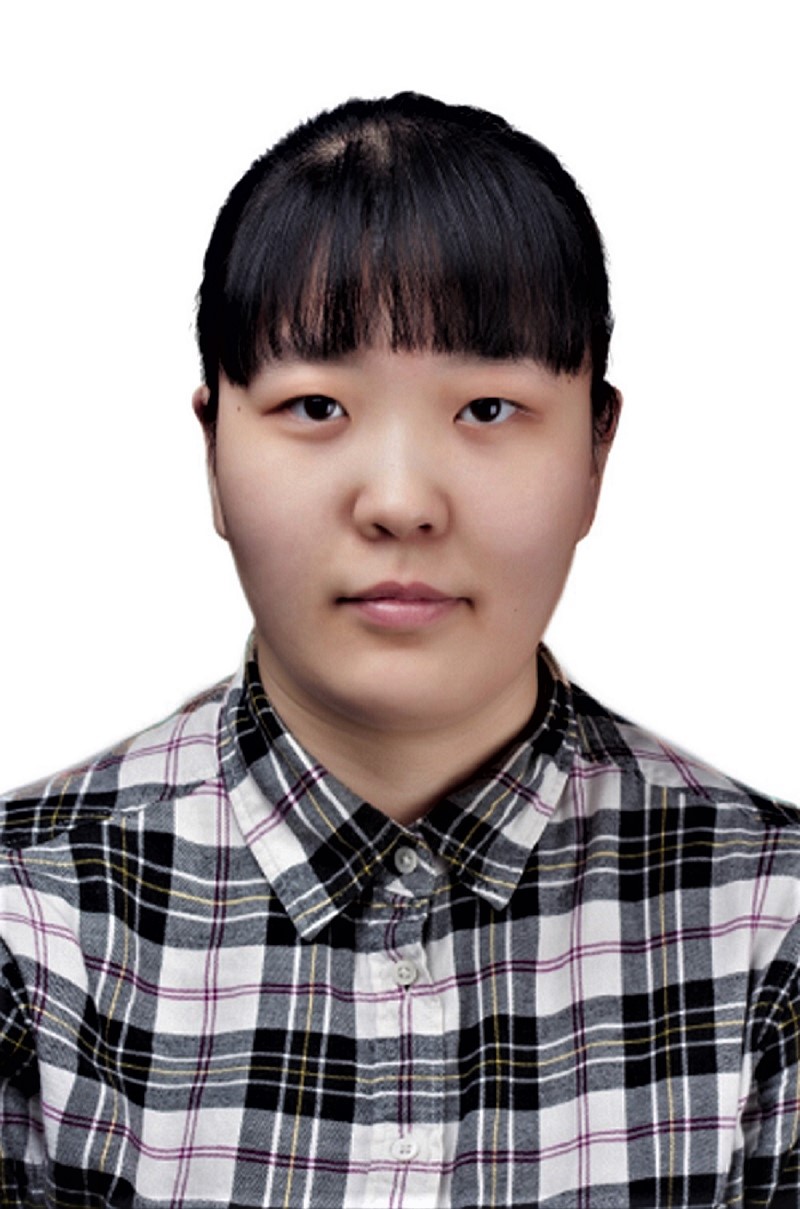}}]{Junshu Sun} 
Ph.D. student at the Institute of Computing Technology, Chinese Academy of Sciences. She received the B.S. degree in Biomedical Engineering from the University of Electronic Science and Technology in 2021. Her current research interests include graph representation learning
and geometric deep learning.
\end{IEEEbiography}

\begin{IEEEbiography}[{\includegraphics[width=1in,height=1.25in,clip,keepaspectratio]{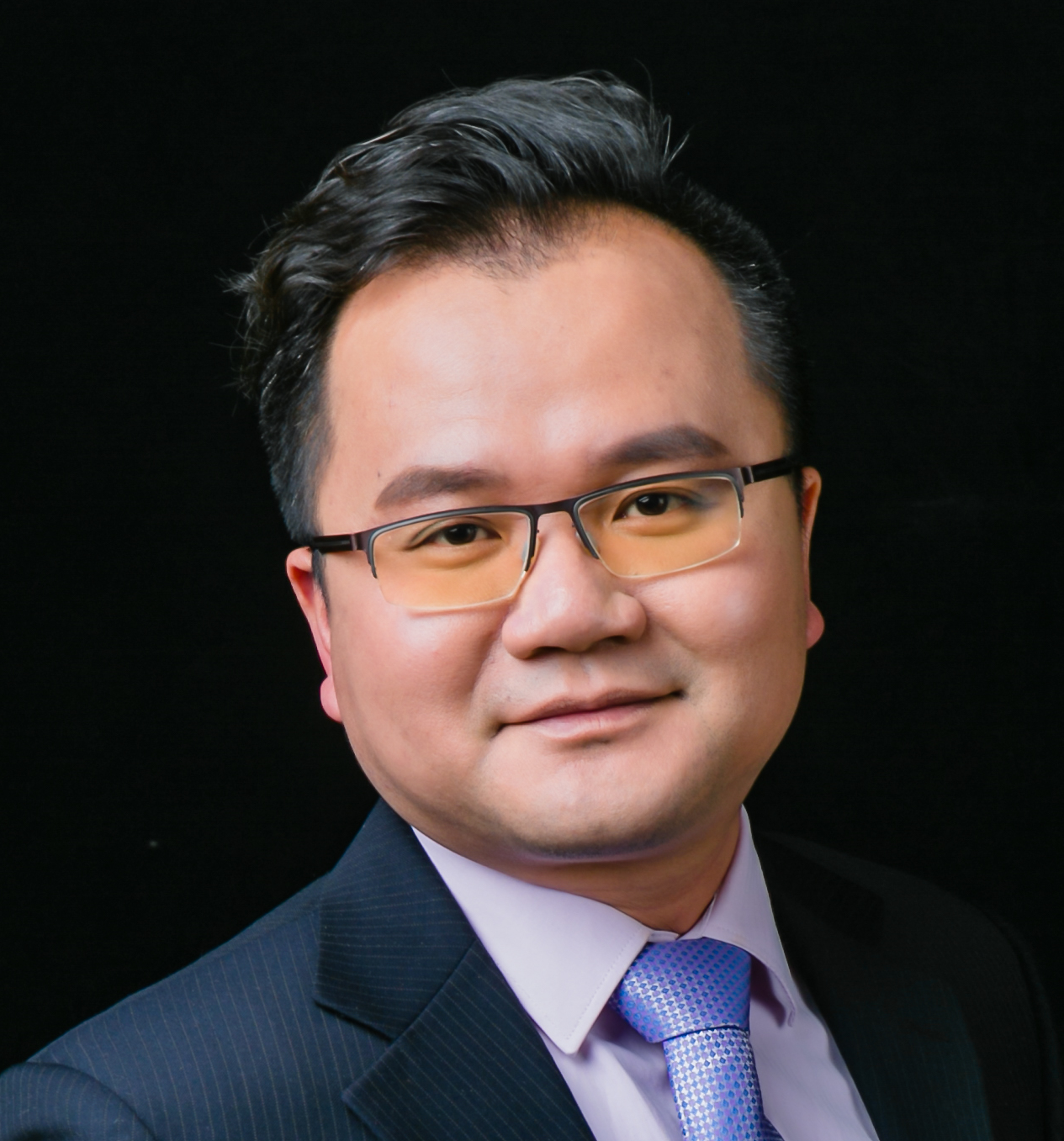}}]{Shuhui Wang}
	Full professor with the Key Laboratory of Intelligent Information Processing (CAS), Institute of Computing Technology, Chinese Academy of Sciences. He received the B.S. degree in electronics engineering from Tsinghua University, in 2006, and the Ph.D. degree from the Institute of Computing Technology, Chinese Academy of Sciences, in 2012. He is also with Pengcheng Laboratory, Shenzhen. His research interests include image/video understanding/retrieval, cross-media analysis, and visual-textual knowledge extraction.
\end{IEEEbiography}

\begin{IEEEbiography}[{\includegraphics[width=1in,height=1.25in,clip,keepaspectratio]{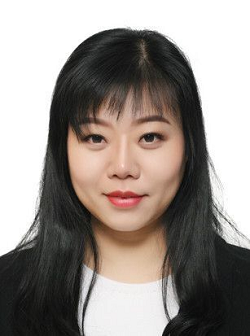}}]{Chenxue Yang}
	received the Ph.D. degree in computer science and technology from the University of Electronic Science and Technology of China, Chengdu, China, in 2016. She was a Post-Doctoral Fellow at the National Key Laboratory of Pattern Recognition, Institute of Automation, Chinese Academy of Sciences, Beijing, China, from 2016 to 2018, and a Senior Engineer at Huawei, from 2018 to 2019. She is now working with the Agricultural Information Institute, Chinese Academy of Agricultural Sciences, Beijing. Her main research interests include visual/multimodal machine learning, smart agriculture, and agricultural data mining. 
\end{IEEEbiography}

\begin{IEEEbiography}[{\includegraphics[width=1in,height=1.25in,clip,keepaspectratio]{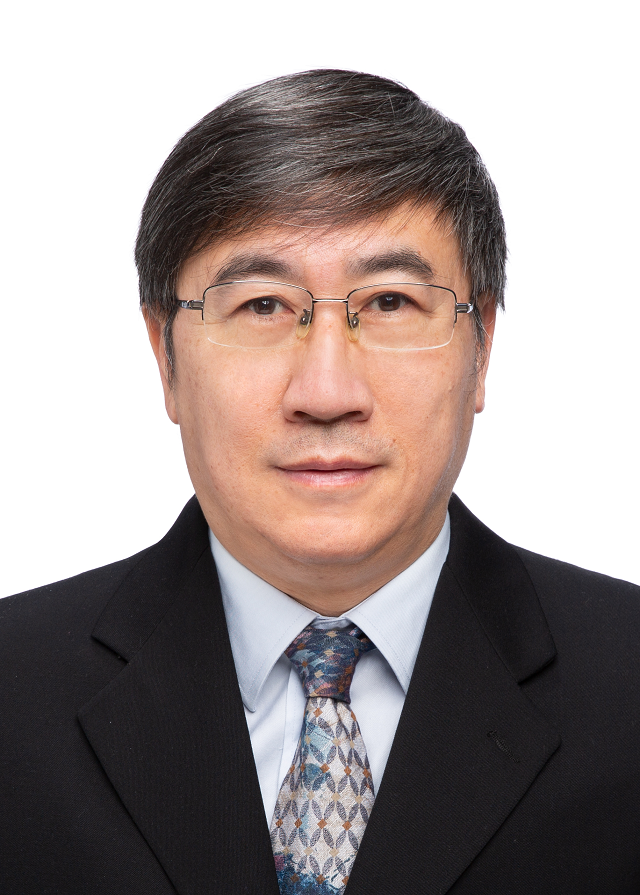}}]{Qingming Huang}
	 Chair professor with the School of Computer Science and Technology, University of Chinese Academy of Sciences. He received the B.S. degree in computer science and the Ph.D. degree in computer engineering from the Harbin Institute of Technology, China, in 1988 and 1994, respectively. He has published more than 500 academic papers in international journals, such as IEEE Transactions on Pattern Analysis and Machine Intelligence, IEEE Transactions on Image Processing, IEEE Transactions on Multimedia, IEEE Transactions on Circuits and Systems for Video Technology, and top-level international conferences, including the ACM Multimedia, ICCV, CVPR, ECCV, VLDB, and IJCAI. He was the associate editor for IEEE Transactions on Circuits and Systems for Video Technology and the associate editor for Acta Automatica Sinica. His research interests include multimedia computing, image/video processing, pattern recognition, and computer vision.
\end{IEEEbiography}

\bibliographystyle{IEEEtran}


%

\end{document}